\documentclass[preprint,article,accept,moreauthors,pdftex]{Definitions/mdpi} 

\firstpage{1} 
\pubvolume{1}
\issuenum{1}
\articlenumber{1}
\pubyear{2021}
\copyrightyear{2021}
\history{Received: date; Accepted: date; Published: date}

\DeclareMathOperator*{\argmin}{arg\,min}

\usepackage{xspace}

\newcommand{\ie}{i.e.\ }

\newcommand{\eg}{e.g.\ }

\newcommand{\cf}{cf.\ }

\newcommand{\m}{\,m\xspace}

\newcommand{\W}{\,W\xspace}

\newcommand{\Wh}{\,Wh\xspace}

\newcommand{\MHz}{\,MHz\xspace}

\newcommand{\GHz}{\,GHz\xspace}

\newcommand{\fps}{\,FPS\xspace}

\newcommand{\fpsPerWatt}{\,FPS/W\xspace}

\newcommand{\MDEs}{\,MDE/s\xspace}

\newcommand{\px}{\,pixels\xspace}

\newcommand{\percent}{\,\%\xspace}

\newcommand{\TULIPP}{\textsc{Tulipp}\xspace}

\newcommand{\KITTI}{KITTI\xspace}

\newcommand{\RESSTAC}{ReS$^2$tAC\xspace}

\usepackage[acronyms]{glossaries} 

\usepackage[nameinlink, capitalize]{cleveref}

\usepackage{KAcolors}  

\usetikzlibrary{positioning,matrix,graphs,arrows,patterns,shapes,external,calc,fit,backgrounds,3d}
\usepackage{pgfplots}
\usepgfplotslibrary{ternary}

\tikzset{
  /tikz/line width=0.6pt,
}

\pgfplotsset{
  compat=1.11,%
  plot coordinates/math parser=false,%
  tick label style={font=\footnotesize},%
  label style={font=\small},%
	scale only axis,%
	axis lines=center,%
	axis on top,%
  every axis legend/.append style={cells={anchor=west},draw=none,font=\small},%
  every axis plot/.append style={semithick},%
	KIT scatter plot A/.style={%
    draw=none,only marks,mark=*,mark options={draw=KITblue,fill=KITblue}},%
	KIT scatter plot B/.style={%
    draw=none,only marks,mark=square*,mark options={draw=KITred,fill=KITred}},%
	KIT scatter plot C/.style={%
    draw=none,only marks,mark=diamond*,mark options={draw=KITorange,fill=KITorange}},%
	KIT scatter plot explicit/.style={%
    scatter,%
    scatter/classes={%
      a={mark=*,KITblue},%
      b={mark=square*,KITred},%
      c={mark=diamond*,KITorange}},%
    only marks,%
    scatter src=explicit symbolic,%
    z buffer=sort},%
  KIT ybar plot A/.style={%
    ybar,fill=KITblue,draw=none},%
  KIT ybar plot B/.style={%
    ybar,fill=KITred,draw=none},%
  KIT ybar plot C/.style={%
    ybar,fill=KITorange,draw=none},%
  KIT xbar plot A/.style={%
    ybar,fill=KITblue,draw=none},%
  KIT xbar plot B/.style={%
    ybar,fill=KITred,draw=none},%
  KIT xbar plot C/.style={%
    ybar,fill=KITorange,draw=none},%
  KIT line plot A/.style={%
    KITblue,semithick},%
  KIT line plot B/.style={%
    KITred,semithick},%
  KIT line plot C/.style={%
    KITorange,semithick},%
  KIT line plot D/.style={%
    KITlilac,semithick},%
  KIT line plot E/.style={%
    KITbrown,semithick},%
  KIT line plot F/.style={%
    KITblue,semithick,dashed},%
  KIT line plot G/.style={%
    KITred,semithick,dashed},%
  KIT line plot H/.style={%
    KITorange,semithick,dashed},%
  KIT line plot I/.style={%
    KITlilac,semithick,dashed},%
  KIT line plot J/.style={%
    KITbrown,semithick,dashed},%
	KIT smooth plot A/.style={%
    KITblue,semithick,smooth},%
  KIT smooth plot B/.style={%
    KITred,semithick,smooth},%
  KIT smooth plot C/.style={%
    KITorange,semithick,smooth},%
  KIT smooth plot D/.style={%
    KITlilac,semithick,smooth},%
  KIT smooth plot E/.style={%
    KITbrown,semithick,smooth},%
  KIT smooth plot F/.style={%
    KITblue,semithick,dashed,smooth},%
  KIT smooth plot G/.style={%
    KITred,semithick,dashed,smooth},%
  KIT smooth plot H/.style={%
    KITorange,semithick,dashed,smooth},%
  KIT smooth plot I/.style={%
    KITlilac,semithick,dashed,smooth},%
  KIT smooth plot J/.style={%
    KITbrown,semithick,dashed,smooth},%
}  

\usepackage{subfig}

\usepackage{import} 
\usepackage{overpic}

\usepackage{amssymb}

\usepackage{todonotes}

\makeatletter
\renewcommand{\todo}[2][]{\tikzexternaldisable\@todo[#1]{#2}\tikzexternalenable}
\makeatother

\usepackage[english]{babel}
\usepackage{blindtext}

\Title{ReS$^2$tAC - UAV-borne Real-time SGM Stereo Optimized for Embedded ARM and CUDA Devices}

\Author{Boitumelo Ruf $^{1,2,}$*, Jonas Mohrs $^{1}$, Martin Weinmann $^{2}$, Stefan Hinz $^{2}$ and Jürgen Beyerer $^{1,3}$}

\AuthorNames{Boitumelo Ruf, Jonas Mohrs, Martin Weinmann, Stefan Hinz and Jürgen Beyerer}

\address{%
$^{1}$ \quad Fraunhofer Institute of Optronics, System Technologies and Image Exploitation (IOSB), Fraunhofer Center for Machine Learning, Karlsruhe, Germany \\
$^{2}$ \quad Institute of Photogrammetry and Remote Sensing, Karlsruhe Institute of Technology (KIT), Karlsruhe, Germany \\
$^{3}$ \quad Vision and Fusion Laboratory, Karlsruhe Institute of Technology (KIT), Karlsruhe, Germany}

\corres{Correspondence: boitumelo.ruf@iosb.fraunhofer.de}

\newacronym{ADAS}{ADAS}{advanced driver assistance systems}
\newacronym{ASG}{ASG}{{Average Shading Gradient}}

\newacronym{BPU}{BPU}{{Branch Prediction Unit}}

\newacronym[shortplural={CNNs}, firstplural={convolutional neural networks (CNNs)}, longplural={convolutional neural networks}]{CNN}{CNN}{convolutional neural network}
\newacronym{COTS}{COTS}{commercial off-the-shelf}
\newacronym[shortplural={CRFs}, longplural={{Conditional Random Fields}}]{CRF}{CRF}{{Conditional Random Field}}
\newacronym{CT}{CT}{{census transform}}

\newacronym{DLT}{DLT}{{Direct Linear Transformation}}
\newacronym{DoG}{DoG}{{Difference of Gaussian}}

\newacronym{EPnP}{EPnP}{{Efficient Perspective-n-Point}}

\newacronym[shortplural={FPGAs}, longplural={field-programmable gate arrays}]{FPGA}{FPGA}{field-programmable gate array}
\newacronym{FPS}{FPS}{frames per second}
\newacronym{FPSW}{FPS/W}{FPS per watt}

\newacronym{GPGPU}{GPGPU}{general purpose computation on a {GPU}}
\newacronym[shortplural={GPUs}, longplural={graphic processing units}]{GPU}{GPU}{graphic processing unit}
\newacronym{GPS}{GPS}{{Global Positioning System}}
\newacronym{GTA}{GTA V}{{Grand Theft Auto V}}

\newacronym{HLS}{HLS}{high-level synthesis}

\newacronym{ICP}{ICP}{{Iterative-Closest-Point}}
\newacronym{IMU}{IMU}{{Inertial Measurement Unit}}
\newacronym{INS}{INS}{{Inertial Navigation System}}

\newacronym{LIDAR}{LiDAR}{{Light Detection and Ranging}}
\newacronym{L1-rel}{$\text{L1-rel}$}{{relative $\text{L1}$-Norm}}
\newacronym{L1-abs}{$\text{L1-abs}$}{{absolute $\text{L1}$-Norm}}

\newacronym{MDEs}{MDE/s}{million disparity estimations per second}
\newacronym{MGM}{MGM}{More-Global Matching}
\newacronym[shortplural={MRFs}, longplural={{Markov Random Fields}}]{MRF}{MRF}{{Markov Random Field}}
\newacronym{MVS}{MVS}{Multi-View Stereo}

\newacronym{NCC}{NCC}{normalized cross-correlation}

\newacronym{PCL}{PCL}{{Point Cloud Library}}

\newacronym{RANSAC}{RANSAC}{{Random Sampling Consensus}}
\newacronym{RMSE}{RMSE}{{Root Mean Square Error}}
\newacronym[shortplural={ROIs}, longplural={regions of interest}]{ROI}{RoI}{region of interest}

\newacronym{SAD}{SAD}{sum of absolute differences}
\newacronym{SFM}{SfM}{{Structure-from-Motion}}
\newacronym{SGBM}{SGBM}{{Semi-Global Block Matching}}
\newacronym{SGM}{SGM}{{Semi-Global Matching}}
\newacronym{SIMD}{SIMD}{{Single-Instruction-Multiple-Data}}
\newacronym{SISD}{SISD}{{Single-Instruction-Single-Data}}
\newacronym{SMDE}{SMDE}{{Self-supervised Monocular Depth Estimation}}
\newacronym[shortplural={SoCs}]{SoC}{SoC}{{System-on-a-Chip}}
\newacronym{SSE}{SSE}{{Streaming SIMD Extensions}}
\newacronym{SSIM}{SSIM}{Structural Similarity}
\newacronym[shortplural={STNs}, longplural={Spatial Transformer Networks}]{STN}{STN}{{Spatial Transformer Network}}

\newacronym[shortplural={UAVs}, longplural={unmanned aerial vehicles}]{UAV}{UAV}{unmanned aerial vehicle}

\newacronym{WTA}{WTA}{winner-takes-it-all}

\abstract{
With the emergence of low-cost robotic systems, such as \acrlong*{UAV}, the importance of embedded high-performance image processing increased more and more. %
For a long time, FPGAs were the only processing hardware, that were capable of high-performance computing, while at the same time preserving a low power consumption, essential for embedded systems. %
However, the recently increasing availability of embedded GPU-based systems, such as the NVIDIA Jetson series, comprised of an ARM CPU and a NVIDIA Tegra GPU, allows for massively parallel embedded computing on graphics hardware. %
With this in mind, we propose an approach for real-time embedded stereo processing on ARM and CUDA enabled devices, which is based on the popular and widely used \acrlong*{SGM} algorithm. %
In this, we propose an optimization of the algorithm for embedded CUDA GPUs, by utilizing massively parallel computing, as well as utilizing the NEON intrinsics to optimize the algorithm for vectorized \acrshort*{SIMD} processing on embedded ARM CPUs. %
We have evaluated our approach with different configurations on two public stereo benchmark datasets to demonstrate that they can reach an error rate as low as $3.3\percent$. %
Furthermore, our experiments show that the fastest configuration of our approach reaches up to $46$\fps on VGA image resolution. %
Finally, in a use-case specific qualitative evaluation, we have evaluated the power consumption of our approach and deployed it on the DJI Manifold 2-G attached to a DJI Matrice 210v2 RTK \gls*{UAV}, demonstrating its suitability for real-time stereo processing onboard a \gls*{UAV}. %
} 

\keyword{embedded stereo vision, real-time stereo processing, disparity estimation, semi-global matching, GPGPU, SIMD, UAV}  

\begin{document}


\section{Introduction}
\label{sec:intro}

\sloppy

\glsreset{UAV}
\glsreset{FPGA}
\glsreset{GPU}
\glsreset{SGM}
\glsreset{SIMD}

In recent years, the use and importance of \glspl*{UAV} in different markets, such as aerial video and photography, precision farming, security monitoring and disaster relief, as well as 3D reconstruction and mapping has greatly increased \citep{Nex2014uav, Restas2015drone, Perz2019uav}. %
And with the ongoing technological advancements, in terms of size, power and durability, the number of areas in which \glspl*{UAV} are used become more and more. %
Prototypes for delivering goods or even transporting people are already available. %
With the increasing use of \glspl*{UAV}, it is becoming evermore important that their usability and safety is ensured \citep{Sebbane2018intelligent, Shakhatreh2019uav}. %
In doing so, modern \glspl*{UAV} are equipped with a range of sensors, including stereo vision sensors, typically used for perceiving the surrounding of the \gls*{UAV} in order to perform the tasks of obstacle detection and avoidance or 3D mapping. %
Compared to active sensors like \gls*{LIDAR} scanners, camera systems in combination with state-of-the-art algorithms are typically more practical in performing these tasks, especially in terms of costs, weight and power-consumption. %
Moreover, such stereo vision sensors are often already integrated in \gls*{COTS} \glspl*{UAV}. %

On the other hand, while a \gls*{LIDAR} sensor directly provides data on the 3D geometry of the scene, using a stereo camera for the same task requires to process the stereo image data and perform a disparity/depth estimation \citep{Nex2011lidar}. %
This, in turn, requires a high-performance embedded processing on-board the \gls*{UAV}. %
For a long time, so-called \glspl*{FPGA} were the only processing hardware, that were capable of high-performance computing, while at the same time preserving a low power consumption, essential for embedded systems. %
In recent years, however, the availability of embedded \glspl*{GPU}, such as the NVIDIA Tegra, allows for massively parallel embedded computing on graphics hardware, which is typically more flexible than \glspl*{FPGA} and less cumbersome to program. %
Furthermore, with the increasing use of deep learning for a wide range of applications, the importance and availability of embedded  \glspl*{GPU} have grown even more. %
With the Jetson boards, comprised of an embedded ARM CPU and an embedded Tegra GPU, NVIDIA provides a suitable alternative to \glspl*{FPGA} for embedded high-performance computing. %
Especially since these systems are recently also being integrated in low-cost \gls*{COTS} \glspl*{UAV}, such as the DJI Matrice \gls*{UAV} in combination with the DJI Manifold or the UVify IFO-S \gls*{UAV}. %

With this in mind, we propose an approach for real-time embedded stereo processing on ARM and CUDA enabled devices, which is based on the well-known and widely used \gls*{SGM} algorithm first proposed by \citet{Hirschmueller2005, Hirschmueller2008}. %
Our main contributions are:  %
\begin{itemize} %
\item the optimization of the algorithm for embedded CUDA \glspl*{GPU}, such as the NVIDIA Tegra, by utilizing massively parallel computing, %
\item the use of the NEON intrinsics to optimize the algorithm for vectorized \acrshort*{SIMD} processing on embedded ARM CPUs, and %
\item the deployment of our approach on the DJI Manifold 2-G attached to a DJI Matrice 210v2 RTK \gls*{UAV} and a use-case specific evaluation with respect to accuracy, processing speed and power consumption. %
\end{itemize} %
Even though we deployed and tested our approach for real-time processing on board a \gls*{UAV}, it is also suitable for other embedded systems, such as those deployed on ground-based robots or those used in \gls*{ADAS}. %

\subsection{Paper outline}

This paper is structured as follows: In \Cref{sec:related_work}, we briefly summarize the related work on embedded stereo processing utilizing embedded \gls*{FPGA}, GPU or CPU hardware and point out how our approach differs from those found in literature. %
In \Cref{sec:methodology}, we first illustrate the general processing pipeline of our approach, in which we also review the general process of deriving the scene depth from a stereo image pair and provide a short review on the \gls*{SGM} algorithm, before illustrating in detail our optimizations for massively parallel stereo processing on CUDA-enabled GPUs and vectorized SIMD processing with NEON intrinsics ARM CPUs. %
We evaluate our approach on two stereo benchmark datasets with respect to accuracy, processing speed and power consumption, as well as in a use-case specific scenario. %
We present the results of our experiments in \Cref{sec:experiments} and discuss our findings in \Cref{sec:discussion}, before providing a summary, concluding remarks, and a short outlook on future improvements in \Cref{sec:conclusion}. %
\subsection{Related Work}%
\label{sec:related_work}
\sloppy

In the following sections, we summarize the related work on embedded stereo processing. %
In \Cref{sec:related_work_fpga}, we will first look at studies that have deployed stereo algorithms on \gls*{FPGA} hardware. 
This is followed by an overview on the emergence of embedded \gls*{GPU} hardware for real-time stereo processing in \Cref{sec:related_work_gpu}. %
Lastly, in \Cref{sec:related_work_cpu}, we revise related work on deploying real-time stereo processing on CPU hardware, both for high-end desktop and embedded environments. %
We also point out how our approach differs from the related work utilizing embedded \gls*{GPU} and CPU hardware for real-time stereo processing. %

\subsubsection{Embedded stereo processing on FPGAs}
\label{sec:related_work_fpga}

The use of \glspl*{FPGA} is key to achieve high-performance image processing with minimal power consumption, especially when relying on computationally expensive algorithms. %
Thus, most implementations of stereo algorithms for embedded systems, in particular of the \gls*{SGM} algorithm \citep{Hirschmueller2005, Hirschmueller2008}, are based on \gls*{FPGA} technology. %
First optimizations of the \gls*{SGM} algorithm, such as those presented in \citep{Gehrig2009real, Banz2010realtime}, were deployed on a PCIe-\gls*{FPGA} card inside a conventional PC or on a separate development kit, achieving real-time frame rates of 27\fps and 30\fps on low-resolution imagery, \ie images with a size of 320$\times$240\px and 640$\times$480\px respectively. %
Due to ongoing technological advancements, the implementation of \citet{Wang2015real}, deployed on an Altera Stratix-IV \gls*{FPGA}-Board, already achieved a frame rate of 67\fps on images with a size of 1024$\times$768\px in 2015. %
However, typical characteristics of embedded systems, besides the dedicated and specialized processing of a specific task, are a small form factor and the integration in larger systems or cooperative environments. %

In their work, \citet{Schmid2013stereo} have deployed the implementation of \citet{Gehrig2010} on a small quadrotor for stereo vision based navigation achieving 14.6\fps on a Spartan 6 \gls*{FPGA}. %
Further \gls*{SoC} developments with respect to size and performance allowed to deploy computationally expensive algorithms on increasingly smaller systems with higher performance. %
\citet{Honegger2014real} implemented the \gls*{SGM} algorithm on a small SO-DIMM sized \gls*{SoC} equipped with a Xilinx Artix7 \gls*{FPGA} and reaching 60\fps with a frame size of 753$\times$480\px . %
By reducing the frame size to 320$\times$240\px the implementation of \citet{Barry2015fpga} reached 120\fps and was used to navigate a small and fast flying fixed-wing \gls*{UAV} around obstacles. %

A number of recent studies \citep{Hofmann2016scalable, Rahnama2018r3sgm, Zhao2020fp} have shown, that further optimizations, such as reducing the number of processing paths in the \gls*{SGM} optimization or increasing parallelization by splitting the input images in independent stripes, as well as the technological advancements, allow to reach frame rates of over 100\fps, while at the same time increasing the accuracy of the stereo algorithm by using a higher image resolution and reducing the form factor, leading to a reduced power consumption of the \gls*{SoC}. %
Yet, the use of \glspl*{FPGA} for real-time embedded image processing involves a cumbersome and time-consuming development, optimization and deployment process. %
In order to reduce development costs of such systems, substantial effort is done to enhance the process of \gls*{HLS} and, in turn, alleviate the development of algorithms for \glspl*{FPGA} with more high-level languages such as C/C++ \citep{Ruf2018realtime, Kalb2019, Zhao2020fp}. %

\subsubsection{On the emergence of embedded processing on GPUs}
\label{sec:related_work_gpu}

The development cycles for implementing and optimizing image processing algorithms for massively parallel processing on GPUs, on the other hand, are much shorter and thus less expensive. %
In addition, GPUs provide a much higher processing power, ideal for algorithms with high computational effort, such as stereo image processing. %
Early works \citep{Rosenberg2006real, Ernst2008mutual} have utilized the rendering pipeline of OpenGL to deploy the \gls*{SGM} stereo algorithm on graphics hardware and reached frame rates of up to 8\fps and 4\fps on VGA image resolution respectively. %
With the introduction of the CUDA-API in 2007, the development costs for \gls*{GPGPU} have dropped even more. %
And so, a lot of implementations of the \gls*{SGM} algorithm for real-time stereo processing on hardware without embedded constraints are optimized and deployed on graphics hardware \citep{Banz2011real, Michael2013realtime, Hernandez2016embedded}. %
\citet{Hernandez2016embedded} show that, with increasing computational power, the use of \gls*{GPGPU} on modern, high-performing graphics hardware, such as the NVIDIA Titan X, allows to reach frame rates of up to 237\fps on VGA image resolution with the conventional use of eight optimization paths inside the \gls*{SGM} optimization. %
Even higher frame rates of up to 475\fps and 886\fps are possible, if the number of optimization paths are reduced to four or two respectively. %

While GPUs provide great computational performance, a major drawback is given by their high power consumption. %
The deployment of the \gls*{SGM} algorithm on a high-end GPU only achieves 1.90\fpsPerWatt on VGA image resolution \citep{Hernandez2016embedded}, while a comparable configuration of the algorithm deployed on a state-of-the-art embedded \gls*{FPGA} achieves 15\fpsPerWatt on a larger image resolution \citep{Zhao2020fp}. %
However, due to the increasing importance of deep learning algorithms for robotic applications, more and more embedded CPU-GPU-based \glspl*{SoC} are released and integrated into robotic systems. %
Recently, such \glspl*{SoC} with embedded GPUs have even been integrated onto \gls*{COTS} \glspl*{UAV}, and thus been made available for the mainstream user. %
In their work, \citet{Hernandez2016embedded} have also deployed their implementation on the NVIDIA Jetson TX1, encapsulating the embedded Tegra X1 GPU, reaching 42\fps (4.19\fpsPerWatt) on VGA resolution and a four path \gls*{SGM} optimization. %
\citet{Chang2020zigzag} further optimized the computation of the \gls*{NCC} matching cost for the use on GPUs and deployed their \gls*{SGM}-based stereo algorithm on the NVIDIA Jetson TX2 reaching 28\fps on images with a size of 1242$\times$375\px. %

In our work, we have also optimized the \gls*{SGM} algorithm for real-time processing on CUDA devices and deployed it on the NVIDIA Jetson TX2 and the more powerful NVIDIA Jetson Xavier AGX. %
In this, we evaluate the performance of different configurations and optimization strategies with respect to performance and power consumption. %
The computationally most expensive part of the \gls*{SGM} algorithm is the aggregation along the different scanlines. %
At the same time, due to the nature of dynamic programming, this is also the part which can by parallelized most effectively, since the computation of each scanline can be done fully independently, without the need of synchronization. %
In terms of \gls*{GPGPU}, this is typically done by instantiating one thread on the graphics hardware for each scanline, resulting in a massively parallel processing \gls*{SGM} path aggregation. %
We have adopted this approach as described in \Cref{sec:methodology-gpu}. %
\begin{table}[t]
\centering
\begin{tabular}{l l l l l l l} \toprule
Reference & HW Device & Embedded \gls*{SoC} & Resolution & Disp. Range & FPS & Power \\ \midrule
\citet{Gehrig2009real} & \gls*{FPGA} & & $320\times240$ & 64 & 27 & <\,3\W \\
\citet{Banz2010realtime} & \gls*{FPGA} & & $640\times480$ & 128 & 30 & n/a \\
\citet{Honegger2014real} & \gls*{FPGA} & \checkmark & $753\times480$ & 32 & 60 & <\,5\W \\
\citet{Wang2015real}$\dagger$ & \gls*{FPGA} & & $1024\times768$ & 96 & 67 & n/a \\
\citet{Barry2015fpga} & \gls*{FPGA} & \checkmark & $320\times240$ & 32 & 120 & <\,5\W$\ast$ \\
\citet{Hofmann2016scalable}$\dagger$ & \gls*{FPGA} & \checkmark & $640\times480$ & 64 & 140 & n/a \\
\citet{Ruf2018realtime} & \gls*{FPGA} & \checkmark & $640\times360$ & 64 & 29 & n/a \\
\citet{Rahnama2018r3sgm}$\dagger$ & \gls*{FPGA} & \checkmark & $640\times480$ & 128 & 109 & <\,3\W \\
\citet{Zhao2020fp}$\dagger$ & \gls*{FPGA} & \checkmark & $1242\times374$ & 128 & 161 & 6.6\W \\ \midrule
\citet{Banz2011real}$\dagger$ & GPU & & $1024\times768$ & 128 & 25 & n/a \\
\citet{Michael2013realtime} & GPU & & $640\times480$ & 64 & 11.7 & n/a \\
\citet{Hernandez2016embedded}$\dagger$ & GPU & \checkmark & $640\times480$ & 128 & 42 & <\,10\W \\
\RESSTAC-CUDA (ours) & GPU & \checkmark & $640\times480$ & 128 & 24 & \textasciitilde\,20\W$\ast$ \\ \midrule
\citet{Gehrig2010}$\dagger$ & CPU & & $640\times320$ & 16 & 14 & n/a \\
\citet{Arndt2013parallel}$\dagger$ & CPU & \checkmark & $640\times480$ & 64 & 0.5 & n/a \\
\citet{Spangenberg2014large}$\dagger$ & CPU & & $640\times480$ & 128 & 16 & n/a \\
\RESSTAC-NEON (ours) & CPU & \checkmark & $640\times480$ & 128 & 7.2 & \textasciitilde\,18\W$\ast$ \\
\bottomrule
\end{tabular}
\caption{Overview of numerous studies that have developed and deployed an \gls*{SGM}-based stereo algorithm on different hardware architectures. %
The highest frame rates, with respect to the corresponding power consumption, are achieved by \gls*{FPGA}-based implementation. %
$\ast$: Power consumption stated for the whole system, \eg including image capture. %
$\dagger$: Studies state measurements for different configurations, however, the configuration listed provides a good trade-off between image resolution and frame rate.} %
\label{tab:related_work}
\end{table}

\subsubsection{Are embedded CPUs suitable for stereo processing?}
\label{sec:related_work_cpu}

In contrast to \glspl*{FPGA} and GPUs, which are designed to be less flexible and yet very powerful in processing specific tasks on a large amount of data, CPUs are designed to do more general and versatile processing, needed to allow computers to instantly react to new sensor input. %
Even though they have much higher clock frequencies, CPUs are often not capable to keep up with the performance achieved by their more specialized counterparts, due to their small number of cores and, in turn, limited ability of parallelization. %

One of the first deployments of the \gls*{SGM} algorithm on a conventional CPU was done by \citet{Gehrig2010}. %
They have implemented a number of different parallelization techniques, among others splitting the eight path optimization scheme into two independent scans. %
With this, they achieve 14\fps on images with a size of 640$\times$320\px, but they only considered a range of 16 disparities. %
They have deployed their algorithm on an Intel Core i7 with four cores and a clock frequency of 3.3\GHz. %
A few years later, \citet{Spangenberg2014large} have achieved 16\fps on VGA resolution and 128 disparities, also running their implementation on a conventional Intel Core i7 with four cores. %
Apart from a number of algorithmic optimizations, \eg disparity space compression and striped computation, they have parallelized the processing by utilizing \gls*{SIMD} vectorization with the \acrshort*{SSE} instruction set from Intel, combined with multi-threading. %

The work of \citet{Arndt2013parallel} is one of the first to deploy an implementation of the \gls*{SGM} algorithm on an embedded CPU, namely the Freescale P4080, reaching a frame rate of only 0.5\fps on VGA image resolution. %
When considering that this was done around the same time as the works of \citet{Gehrig2010} and \citet{Spangenberg2014large}, it illustrates the gap between conventional and embedded CPUs in terms of performance. %
However, the embedded CPU technology has also evolved and gained performance. %
Most of the modern high-end embedded \glspl*{SoC} typically consist of an ARM CPU and a \gls*{FPGA} or a GPU, for example the Xilinx Ultrascale series or the NVIDIA Jetson series. %
\citet{Rahnama2018real} implemented the ELAS stereo algorithm \citep{Geiger2011} on a Xilinx ZC706 \gls*{SoC} made up of an ARM CPU and a \gls*{FPGA}. %
In this, they have deployed the computationally most expensive stages of the algorithm onto the \gls*{FPGA} (if possible) and used the ARM CPU to process the stages with unpredictable memory access patterns. %
\citet{Saidi2020} accelerated a simple stereo algorithm on an ARM CPU achieving frame rates of up to 59\fps on an image resolution of 320$\times$240\px. %
In this, they have parallelized the algorithm by using multi-threading as well as \gls*{SIMD} vectorization on ARM with the NEON instruction set \citep{Arm2013neon}. %

Just as the dynamic programming in the \gls*{SGM} path aggregation is well suited for massively parallel computing on graphics hardware, its computation can also be effectively vectorized by \gls*{SIMD} processing as \citet{Spangenberg2014large} have shown. %
With this in mind, we want to investigate the ability of embedded CPUs in performing high-accuracy stereo processing in real-time based on the \gls*{SGM} algorithm. %
To the best of our knowledge, our work is the first to implement and optimize the \acrlong*{SGM} stereo algorithm on an embedded ARM CPU, leveraging multi-threading parallelization and \gls*{SIMD} vectorization with NEON intrinsics. %
We have summarized a relevant excerpt of the related work on real-time \gls*{SGM} stereo processing on \gls*{FPGA}, GPU and CPU hardware in \Cref{tab:related_work}. %
\section{Materials and Methods}%
\label{sec:methodology}

\sloppy

\glsreset{CT}
\glsreset{NCC}

In the following sections, we first give a general overview on the processing pipeline of our approach (\Cref{sec:methodology-overview}), in which we also review the general process of deriving the scene depth from a stereo image pair and provide a short review on the \gls*{SGM} algorithm. %
This is followed by detailed descriptions on our optimizations for massively parallel stereo processing on CUDA-enabled GPUs (\Cref{sec:methodology-gpu}) and vectorized SIMD processing with NEON intrinsics ARM CPUs (\Cref{sec:methodology-cpu}). %

\subsection{Processing pipeline for real-time dense disparity estimation}
\label{sec:methodology-overview}

\begin{figure}[t] %
	\resizebox{0.98\columnwidth}{!}{\subimport{figures/}{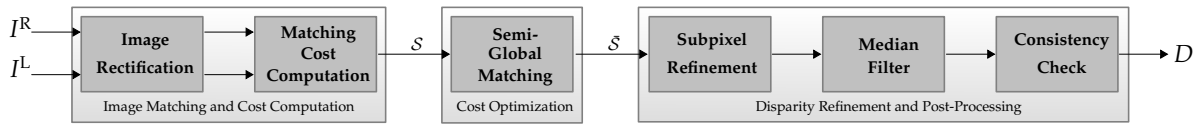}}%
	\caption{Processing pipeline for real-time dense disparity estimation consisting of three subsequent steps, which in turn are made of smaller building blocks.}
	\label{fig:stereo_pipeline_overview}
\end{figure}

In their work, \citet{Scharstein2002} have studied and categorized a number of stereo algorithms for dense disparity or depth estimation based on their processing steps. %
From their observations they have derived a general processing pipeline which provides a basic blueprint for most modern stereo algorithms, as well as for our approach proposed in this work, as illustrated in \Cref{fig:stereo_pipeline_overview}. %
In this, the estimation of dense disparity maps can be subdivided into three subsequent steps, which are made of smaller building blocks. %
In the following, we will look at each individual processing step, give a short overview of its task, and subsequently provide a detailed description on how we have instantiated it in our proposed approach. %

\subsubsection{Dense image matching and cost computation}
\label{sec:methodology-cost}

Finding two corresponding image points, that depict the same scene point in both images of a stereo camera setup (\cf \Cref{sec:appendix_stereo}), means to match the pixels of the reference image, typically the image of the left camera, against each pixel in the matching image within a certain disparity range $d \in \Gamma = [d_\mathrm{min},d_\mathrm{max}]$. %
In this, the similarity between these the two pixels is modeled by a similarity measure from which a cost function $\Phi$ can be deduced, which typically is at its minimum if both pixels coincide.
This so-called matching cost between a pixel $\mathrm{p}$ in the left image and a corresponding pixel $\mathrm{q}$, located according to a disparity shift $d$ in the right image, is stored in the three-dimensional cost volume $\mathcal{S}$: %
\begin{equation} %
	\mathcal{S}(\mathrm{p}, d) = \Phi\left( I^\mathrm{L}(p_x, p_y), I^\mathrm{R}(p_x - d, p_y) \right). %
	\label{eq:image_matching}
\end{equation} %
Thus, the objective in finding two corresponding image points is to minimize the matching cost computed by $\Phi$. %

When relying on distinctive image features like SIFT \citep{Lowe2004distinctive}, SURF \citep{Bay2006surf} or ORB \citep{Rublee2011orb} features, a unique matching between two image points can be found. %
However, such image features can only be calculated in descriptive image regions, resulting in a very sparse correspondence field. %
Thus, in the case of dense image matching \Cref{eq:image_matching} is evaluated for each pixel in the reference image, computing a pixel-wise matching cost $s$ according to $\Phi$, which indicates the similarity between the pixel pair. %
This cost is then stored inside a three-dimensional cost volume $\mathcal{S}$ from which the disparity map is later extracted. %
In our work, we have implemented the two most commonly used similarity measures for real-time dense image matching, namely the Hamming distance of the non-parametric \gls*{CT} \citep{Zabih1994} and the \gls*{NCC}. %
Since the Hamming distance of the \gls*{CT} is minimal when both image patches are most similar, it can directly be used as the matching cost $s_\mathrm{CT}$. %
The \gls*{NCC}, however, is equal to 1 when both image patches are equal. %
Thus, the \gls*{NCC} is inverted and truncated before being evaluated as matching cost: $s_\mathrm{NCC} = 1 - \max{\left(0, \Phi_\mathrm{NCC}\right)}$.

Since the disparity is only evaluated along the same pixel row (\cf \Cref{eq:image_matching}), it is assumed that the input images $I^\mathrm{L}$ and $I^\mathrm{R}$ are rectified prior to the process of image matching. %
Similar to the way described by \citet{Ruf2018realtime}, we use a standard calibration routine implemented in the OpenCV library \citep{Zhang2000flexible} to calculate two rectification maps, which allow to efficiently resample the input images such that the epipolar lines lie horizontally on the image rows. %

\subsubsection{Cost optimization and disparity computation}
\label{sec:methodology-optimization}
\glsreset{SGM}

Given the previously computed three-dimensional cost volume $\mathcal{S}$, the next step consists in extracting a plausible disparity map $D$. %
Since each voxel of $\mathcal{S}$ holds the matching cost of a particular pixel $\mathrm{p}$ of the reference image, associated with a certain disparity $d$ within the studied disparity range, a straightforward approach to compute a disparity map from the cost volume is to take the \textit{\gls*{WTA}} solution for all pixels inside the reference image: 
\begin{equation} %
	D(\mathrm{p}) = \argmin_{d \in \Gamma} \mathcal{S}(\mathrm{p}, d). %
	\label{eq:wta}
\end{equation} %
However, due to the limited descriptiveness of the cost functions and resulting ambiguities in the cost volume, this would lead to a noisy and unsuitable disparity map. %
Thus, it is important to perform a cost optimization and in turn regularize the cost volume. %

\citet{Scharstein2002} have categorized the stereo methods according to their cost optimization strategies into \textit{local} and \textit{global} methods. %
While the first group of algorithms only optimize the cost volume in a locally confined window and make implicit smoothness assumptions, global methods explicitly state their regularization scheme and perform an optimization within the whole image domain. %
Global methods will thus produce more accurate disparity maps compared to those estimated by local methods. %
Yet, at the same time, the use of global methods for embedded stereo processing is not feasible due to their complexity. %

In their taxonomy, \citet{Scharstein2002} additionally propose a third group of algorithms, which alleviate dynamic programming to compute a disparity map. %
Algorithms in this group usually state an explicit smoothness assumption and approximate a global optimization scheme, which is why this group can be considered as a subgroup of the global methods. %
The most prominent algorithm, especially for real-time embedded processing, is the so-called \gls*{SGM} algorithm \citep{Hirschmueller2005, Hirschmueller2008}. %
In this, the optimization scheme is formulated as a \gls*{MRF} and the minimization of the energy function with its explicit smoothness assumption is approximated by aggregating the matching costs along a number of concentric paths for each pixel $\mathrm{p}$ within the image domain: %
\begin{equation} %
\label{eq:sgm_path} %
\begin{aligned} %
	L_\mathrm{r}(\mathrm{p}, d) = \mathcal{S}(\mathrm{p}, d) &+ \min\limits_{d'\in \Gamma}\left(L_\mathrm{r}(\mathrm{p-r}, d')+\mathcal{V}(d,d')\right),\ \text{with} \\ %
	\mathcal{V}(d,d') &= %
	\begin{cases} %
		0 ,\ \text{if}\ d = d' \\ %
		P_1 ,\ \text{if}\ \left|d-d'\right| = 1 \\ %
		P_2 ,\ \text{if}\ \left|d-d'\right| > 1. %
	\end{cases} %
\end{aligned}  %
\end{equation} %
For each pixel $\mathrm{p}$ and disparity $d$ inside the cost volume $\mathcal{S}$, the matching costs are recursively aggregated into $L_\mathrm{r}$, while moving along the path with the direction $\mathrm{r}$. %
Within the smoothness term $\mathcal{V}(d,d')$ the matching costs of the previously considered pixel, with respect to all evaluated disparities $d'$, are penalized according to the disparity difference between $d$ and $d'$. %
Finally, for each pixel, all path costs are summed up and stored inside an aggregated cost volume $\bar{\mathcal{S}}$:
\begin{equation} %
\label{eq:sgm_path_aggr} %
\bar{\mathcal{S}}(\mathrm{p}, d) = \sum\limits_\mathrm{r} L_\mathrm{r}(\mathrm{p}, d) ,
\end{equation} %
before extracting the \gls*{WTA} disparity map according to \Cref{eq:wta} from the same.

The use of dynamic programming, \ie breaking down the minimization problem of the energy function into the aggregation of independent one-dimensional paths, makes the \gls*{SGM} approach well-suited for massively parallel computing and vector processing, and in turn for embedded processing. %
At the same time, many studies have shown that the results of the \gls*{SGM} algorithm are very accurate, making it one of the most widely used algorithms for real-time and accurate stereo processing. %
In our work, we have parallelized and optimized the \gls*{SGM} algorithm to run in real-time with \gls*{SIMD} vector processing on embedded ARM CPUs as well as CUDA enabled GPUs. %

\subsubsection{Post-processing}
\label{sec:methodology-post-processing}

There are a number of post-processing steps, \ie filtering, regularization and further optimizations, which can be applied to the initial disparity map in the final stage of the processing pipeline. %
In our pipeline, we have implemented a subpixel disparity refinement, a left-right consistency check for occlusion detection, as well as a final median filter. %

\paragraph{Subpixel disparity refinement:} %

The initial disparity map computed by the \gls*{SGM} optimization is made up of discrete disparity values, which is sufficient for a diversity of robotic applications such as perception of the surrounding and obstacle detection. %
However, when the aim is to accurately reconstruct the scene, it is important to also account for slanted surfaces and thus incorporate a subpixel refinement of the disparity. %
A simple and yet effective way to implement such a refinement is to use the minimum matching cost for each pixel, \ie the matching cost corresponding to the \gls*{WTA} disparity $\hat{d}$, as well as the matching costs of the two neighboring disparities in front and behind of $\hat{d}$, and fit a parabola through these three matching costs. %
The location of the minimum of this parabola with respect to $\hat{d}$ is then considered as the subpixel refinement and added to $\hat{d}$. 
This optimization has only a minor computational overhead and is thus well-suited for real-time processing. %
However, it requires to work with floating point arithmetic which might be of restriction to some embedded hardware. %

\paragraph{Occlusion detection by left-right consistency check:} %

A typical approach to detect and filter pixels in occluded areas is the left-right consistency check (\cf \Cref{sec:appendix_consistencycheck}). %
This, however, requires the computation of a second disparity map $D^{\mathrm{R}}$, which corresponds to the right image of the stereo pair. %
A straightforward approach to compute $D^{\mathrm{R}}$ would be to swap and horizontally flip the input images and repeat the image matching, cost optimization and disparity computation as described above. %
This, however, would mean to execute the first and computationally most expensive steps of the processing pipeline twice for each image pair. %
Yet, the computation of $D^{\mathrm{R}}$ can be efficiently approximated by reusing the aggregated cost volume $\bar{\mathcal{S}}$ from the cost optimization step:

\begin{equation} %
\label{eq:right_disparity} %
D^{\mathrm{R}}(\mathrm{p}) = \argmin_{d \in \Gamma} \bar{\mathcal{S}}((p_x+d, p_y), d).
\end{equation} %

In this work, we employ the approximated computation of $D^{\mathrm{R}}$ for real-time processing and evaluate how it performs compared to computing the right disparity map from scratch. %

\paragraph{Median filter:} %

For a final outlier removal, we employ a 3$\times$3 median filtering for all remaining pixels with valid disparities. %
A median filtering requires a sorting of all disparities within the local window, which is especially cumbersome when optimizing for vector processing on a CPU. %
Thus, we are utilizing sorting networks to efficiently perform a bubble sort when running the algorithm on the ARM CPU (\cf \Cref{sec:methodology_cpu_median}). %
We chose a confined neighborhood size of 3$\times$3\px because of two reasons: %
First, to not introduce too much smoothing or object fattening in the resulting disparity map. %
And secondly, to keep the computational complexity in the process of sorting the disparity values low. %

\subsection{Real-time processing by massively parallel computing on CUDA-enabled GPUs}
\label{sec:methodology-gpu}

\glsreset{GPGPU}

As described in \Cref{sec:appendix_gpu_programming}, the CUDA API allows for massively parallel \gls*{GPGPU} on NVIDIA GPUs.
Thus, we have implemented each step of the stereo algorithm in separate CUDA kernels in order to optimize the stereo algorithm for embedded NVIDIA GPUs. %
Since each kernel execution is aimed to achieve a high utilization of the GPU, we refrained from a parallel execution of the CUDA kernel methods with CUDA streams. %
In the following, we provide a detailed description on how we have optimized and instantiated each step of the algorithm for an efficient \gls*{GPGPU}. %

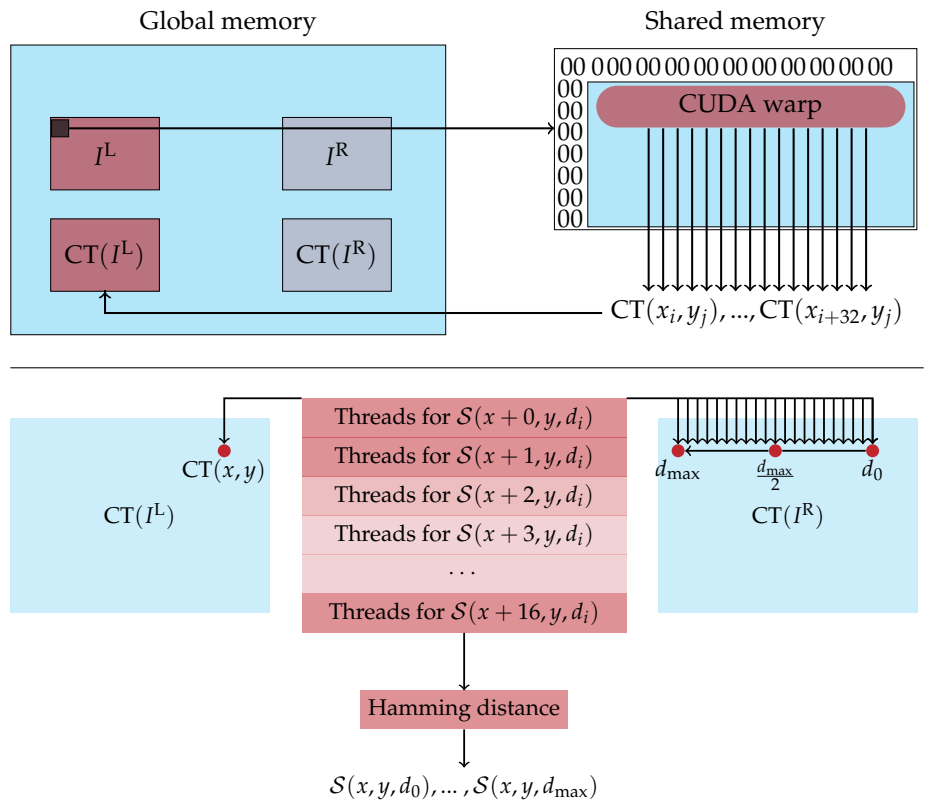
\begin{figure}[t] %
	\centering %
	\subfloat{\resizebox{0.75\textwidth}{!}{\begin{tikzpicture}

\node[] (text1) at (4,10.6)  {Global memory};
\node[] (text2) at (11,10.6) {Shared memory};


\node[rectangle, fill=KITcyanblue, fill opacity=0.3, draw, draw opacity=1, minimum width=6cm, minimum height=4cm] (global) at (4,8.3) {};
\node[rectangle, draw, draw opacity=1, minimum width=5cm, minimum height=2.5cm] (shared1) at (11,9) {};
\node[rectangle, fill=KITcyanblue, fill opacity=0.3, draw, draw opacity=1, minimum width=4.5cm, minimum height=2cm] (shared2) at (11.21,8.79) {};

\node[rectangle, fill=KITred,   fill opacity=0.6, draw, draw opacity=1, minimum width=1.5cm, minimum height=1cm, text opacity=1]     (img1) at (2.3,8.8)   {$I^\mathrm{L}$};
\node[rectangle, fill=KITred,   fill opacity=0.2, draw, draw opacity=1, minimum width=1.5cm, minimum height=1cm, text opacity=1]     (img2) at (5.5,8.8)   {$I^\mathrm{R}$};
\node[rectangle, fill=KITred,   fill opacity=0.6, draw, draw opacity=1, minimum width=1.5cm, minimum height=1cm, text opacity=1]     (img3) at (2.3,7.4)   {$\text{CT}(I^\mathrm{L})$};
\node[rectangle, fill=KITred,   fill opacity=0.2, draw, draw opacity=1, minimum width=1.5cm, minimum height=1cm, text opacity=1]     (img4) at (5.5,7.4)   {$\text{CT}(I^\mathrm{R})$};
\node[rectangle, fill=KITblack, fill opacity=0.8, draw, draw opacity=1, minimum width=0.2cm, minimum height=0.2cm]   (wnd) at (1.68,9.15) {};

\draw[->, thick] (wnd.east) -- (8.5,9.15);
\node[rounded rectangle, fill=KITred, fill opacity=0.6, text opacity=1, minimum width = 4.5cm] (warp) at (11.21,9.45) {CUDA warp};

\node[] (result) at (11.3,6.6) {$\text{CT}(x_i,y_j),...,\text{CT}(x_{i+32},y_j)$};

\draw[->,thick] (9.8,9.15)   -- (9.8, 6.9);
\draw[->,thick] (10,9.15)    -- (10, 6.9);
\draw[->,thick] (10.2,9.15)    -- (10.2, 6.9);
\draw[->,thick] (10.4,9.15)    -- (10.4, 6.9);
\draw[->,thick] (10.6,9.15)    -- (10.6, 6.9);
\draw[->,thick] (10.8,9.15)    -- (10.8, 6.9);
\draw[->,thick] (11,9.15)    -- (11, 6.9);
\draw[->,thick] (11.2,9.15)    -- (11.2, 6.9);
\draw[->,thick] (11.4,9.15)   -- (11.4, 6.9);
\draw[->,thick] (11.6,9.15)    -- (11.6, 6.9);
\draw[->,thick] (11.8,9.15)    -- (11.8, 6.9);
\draw[->,thick] (12,9.15)    -- (12.0, 6.9);
\draw[->,thick] (12.2,9.15)    -- (12.2, 6.9);
\draw[->,thick] (12.4,9.15)    -- (12.4, 6.9);
\draw[->,thick] (12.6,9.15)    -- (12.6, 6.9);
\draw[->,thick] (12.8,9.15)    -- (12.8, 6.9);

\draw[->,thick] (result.west) -- (2.3,6.6) |- (img3.south);

\node[] (z1) at (8.72,7.9) {$00$};
\node[] (z2) at (8.72,8.2) {$00$};
\node[] (z3) at (8.72,8.5) {$00$};
\node[] (z4) at (8.72,8.8) {$00$};
\node[] (z5) at (8.72,9.1) {$00$};
\node[] (z6) at (8.72,9.4) {$00$};
\node[] (z7) at (8.72,9.7) {$00$};
\node[] (z8) at (10.2,10.0) {$00$};
\node[] (z9) at (8.76,10.0) {$00$};
\node[] (z9) at (9.1,10.0) {$0$};
\node[] (z11) at (9.4,10.0) {$00$};
\node[] (z12) at (9.8,10.0) {$00$};
\node[] (z13) at (10.6,10.0) {$00$};
\node[] (z14) at (11.0,10.0) {$00$};
\node[] (z15) at (11.4,10.0) {$00$};
\node[] (z16) at (11.8,10.0) {$00$};
\node[] (z17) at (12.2,10.0) {$00$};
\node[] (z18) at (12.6,10.0) {$00$};
\node[] (z19) at (13.0,10.0) {$00$};

\end{tikzpicture}}} \\ %
	\rule{0.75\textwidth}{0.4pt} \\ %
	\subfloat{\resizebox{0.75\textwidth}{!}{\begin{tikzpicture}
	
	\node[rectangle, opacity=0.5, fill=KITred, text opacity=1, minimum width=5cm] (t1) at (5,6)   {Threads for $\mathcal{S}(x+0,y, d_i)$};
	\node[rectangle, opacity=0.5, fill=KITred, text opacity=1, minimum width=5cm] (t2) at (5,5.4) {Threads for $\mathcal{S}(x+1,y, d_i)$};
	\node[rectangle, opacity=0.3, fill=KITred, text opacity=1, minimum width=5cm] (t3) at (5,4.8) {Threads for $\mathcal{S}(x+2,y, d_i)$};
	\node[rectangle, opacity=0.2, fill=KITred, text opacity=1, minimum width=5cm] (t4) at (5,4.2) {Threads for $\mathcal{S}(x+3,y, d_i)$};
	\node[rectangle, opacity=0.2, fill=KITred, text opacity=1, minimum width=5cm, minimum height=0.6cm] (t5) at (5,3.6) {$\cdots$};
	\node[rectangle, opacity=0.5, fill=KITred, text opacity=1, minimum width=5cm] (t6) at (5,3)   {Threads for $\mathcal{S}(x+16,y, d_i)$};
	
	\node[rectangle, minimum width=4cm, minimum height=3cm, opacity=0.2, fill=KITcyanblue, text opacity=1] (ref) at (0, 4.5) {$\text{CT}(I^\mathrm{L})$};
	\node[rectangle, minimum width=4cm, minimum height=3cm, opacity=0.2, fill=KITcyanblue, text opacity=1] (ref) at (10, 4.5) {$\text{CT}(I^\mathrm{R})$};

\draw[circle,fill=KITred,minimum size=0.2cm,inner sep=0pt]
	(1.3,5.5) node[fill=KITred] (refPixel){};
\draw[circle,fill=KITred,minimum size=0.2cm,inner sep=0pt]
	(11.3,5.5) node[fill=KITred]  (matPixel1) {};
\draw[circle,fill=KITred,minimum size=0.2cm,inner sep=0pt]
	(9.8,5.5) node[fill=KITred]  (matPixel2) {};
\draw[circle,fill=KITred,minimum size=0.2cm,inner sep=0pt]
	(8.3,5.5) node[fill=KITred]   (matPixel3) {};
	
\draw[thick] (matPixel1.west) -- (matPixel2.east);
\draw[->,thick] (matPixel2.west) -- (matPixel3.east);

\node[anchor=north, yshift=-0.2] (b0) at (refPixel)  {$\text{CT}(x,y)$};
\node[anchor=north, yshift=-0.2] (b1) at (matPixel1) {$d_0$};
\node[anchor=north, yshift=-0.3] (b2) at (matPixel2) {$\frac{d_\mathrm{max}}{2}$};
\node[anchor=north, yshift=-0.2] (b3) at (matPixel3) {$d_\mathrm{max}$};

\node[rectangle, fill=KITred, opacity=0.5, text opacity=1] (hamming) at (5,1.5) {Hamming distance};
\node[] (cv) at (5,0.3) {$\mathcal{S}(x,y,d_{0}), ...\ ,\mathcal{S}(x,y,d_\mathrm{max})$};

\draw[->,thick] (t1.north west) -| (1.3,6.3)  -- (refPixel.north);
\draw[->,thick] (t1.north east) -| (11.3,6.3) -- (matPixel1.north);

\foreach \x in {0,...,20}{
	\draw[->,thick] (8.3 + 0.15 * \x, 6.3) -- (8.3 + 0.15 * \x, 5.6);
}

\draw[->,thick] (t6.south) -- (hamming.north);
\draw[->,thick] (hamming.south) -- (cv.north);

\end{tikzpicture}}} %
	\caption {%
	    \glsreset{CT}
		\textbf{Top:} To calculate the \acrlong*{CT} by a CUDA warp for a specific image region, the image data of this region is first copied from the global memory to the shared memory, the latter having higher access speeds. %
		In the second step, each thread of the CUDA warp calculates the \gls*{CT} for one pixel inside this region. %
		\textbf{Bottom:} The threads of a CUDA warp calculate the Hamming distance for 16 disparities simultaneously. %
	 } %
	\label{fig:cuda_impl_census} %
\end{figure} %

\subsubsection{Matching cost computation}
\label{sec:methodology-gpu-cost}
\glsreset{CT}
\glsreset{NCC}

As already mentioned in \Cref{sec:methodology-cost}, we have implemented two different matching cost functions, namely the Hamming distance of the \gls*{CT} as well as an inverted and truncated version of the \gls*{NCC}. %
A detailed description on the implementation and the optimization of these two cost functions for execution on a GPU with CUDA is given in the following. %

\paragraph{Calculating the census transformation and its Hamming distance:}

For the parallel calculation of the \gls*{CT} on the GPU, we assign different image regions to each instantiation of the corresponding CUDA kernel. %
Before the actual \gls*{CT} is calculated, each kernel instantiation copies the pixel data of the considered image region into shared memory in order to achieve a higher access speed. %
The computation of the \gls*{CT} is then performed in parallel by each thread of the thread block for a specific pixel in the assigned image region. %
To account for pixels at the image border, where a part of the neighborhood lies outside of the image, a zero-valued margin with the size of the neighborhood radius is assigned to the image for the calculation of the \gls*{CT}. %
Furthermore, we separated the calculation of the Hamming distance from the calculation of the \gls*{CT}, since the two kernel methods of these two steps are instantiated with different parameters. %
For the parallelization on the GPU, we have implemented a \gls*{CT} with a neighborhood size of $5\times5$\px and $9\times7$\px, the latter being the largest neighborhood that fits into a 64\,bit integer. %
The choice of the former neighborhood size is justified by the limitations of the optimized implementation for the CPU (\cf \Cref{sec:methodology-cpu-ct}). %
An overview of our implementation is illustrated by \Cref{fig:cuda_impl_census} (top). %

In the calculation of the Hamming distance, the reference and matching image are divided into stripes and each stripe is assigned to different CUDA warps. %
Each thread of the CUDA warps then calculates the Hamming distance from the corresponding census descriptors at a certain pixel position and disparity. %
We have chosen the dimensions of the CUDA warps in such a way that for 16 different pixels half the disparities can be calculated simultaneously (\Cref{fig:cuda_impl_census} (bottom)). %
In this, the census descriptor of the reference pixels is first loaded by all threads of a thread block into the shared memory. %
Then all threads load the census descriptors of the matching pixel given a certain disparity. %
This means that the 32 threads of a thread block load the census descriptors $CT(x-i,y),...,CT(x-i-31,y)$ from the matching image into the shared memory. %
This is repeated until for all disparities $[0, d_\mathrm{max}]$ the census descriptors of the matching image are loaded into the shared memory. %
Given all the census descriptors, the threads of a thread block compute the Hamming distance simultaneously for different pixels and store the result at the corresponding position in the cost volume. %
Since the matching cost of the different disparities for one pixel lie directly next to each other in the cost volume, the dimension of the CUDA warp is chosen in such a way that the memory access from the GPU can be pooled together. %

\paragraph{Inverted and truncated version of the normalized cross-correlation:}

Using the \gls*{NCC} as a cost function is computationally more expensive than relying on the Hamming distance of the \gls*{CT}. %
While the computation of the \gls*{CT} and the subsequent Hamming distance only requires some comparative and bit-level operations, the computation of the \gls*{NCC} needs the calculation of a mean and variance of the two input patches. %
Since the mean and the variance of all possible patches inside an image can be precomputed and then reused, we divide the calculation of the \gls*{NCC} and the process of image matching into two separate stages, similar to the calculation of the \gls*{CT} and the Hamming distance.

In the first step, we calculate the mean and variance for all patches in the left and right input image. %
In this, we instantiate a kernel with the same configuration as when calculating the \gls*{CT}, iterating over all pixels in the left and right image, and compute the necessary data for a patch of a given size, centered around the current pixel. %
Similar to the process of computing the \gls*{CT}, we thus first enhance the input images with the independent patch information, storing the patch-mean and patch-variance, together with the pixel value of the center pixel, in a special struct for each pixel of the input image.
Just as when calculating the Hamming distance of the \acrlong*{CT}, we then use the pixel and the patch data to perform the image matching based on the inverted and truncated \acrlong*{NCC} and fill the resulting cost volume in the second stage. %
Again, we instantiate a kernel with the same parameters, as when doing the image matching with the Hamming distance. %
We have implemented the \gls*{NCC} for a patch size of $5\times5$\px and $9\times9$\px. %

\subsubsection{Semi-Global Matching optimization}
\label{sec:methodology-gpu-sgm}

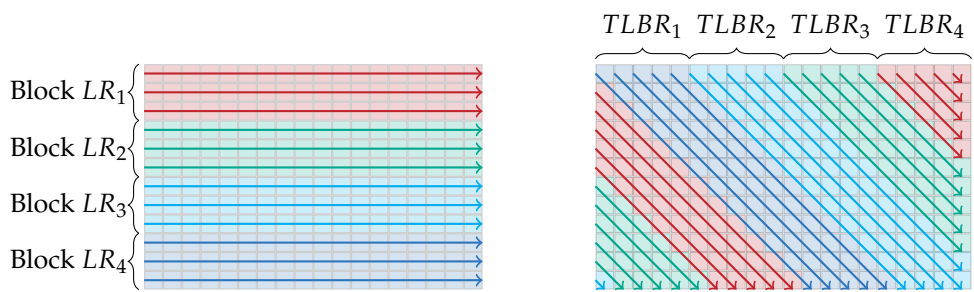
\begin{figure}[b!] %
	\centering
	\begin{tikzpicture}
		
	\foreach \x in {1,...,18}{
		\foreach \y in {1,...,3}{
			\node[rectangle, draw, opacity=0.2, fill=KITblue] (a-\x-\y) at (0.25 * \x, 0.25 * \y) {};
		}
	}
	
	\foreach \x in {1,...,18}{
		\foreach \y in {4,...,6}{
			\node[rectangle, draw, opacity=0.2, fill=KITcyanblue] (b-\x-\y) at (0.25 * \x, 0.25 * \y) {};
		}
	}
	
	\foreach \x in {1,...,18}{
		\foreach \y in {7,...,9}{
			\node[rectangle, draw, opacity=0.2, fill=KITgreen] (c-\x-\y) at (0.25 * \x, 0.25 * \y) {};
		}
	}
	
	\foreach \x in {1,...,18}{
		\foreach \y in {10,...,12}{
			\node[rectangle, draw, opacity=0.2, fill=KITred] (d-\x-\y) at (0.25 * \x, 0.25 * \y) {};
		}
	}
	
	\draw [decorate,decoration={brace,amplitude=4pt,raise=2pt}]
		(a-1-1.south west) -- (a-1-3.north west) node [black,midway,xshift=-1cm]   {Block $LR_4$};
	\draw [decorate,decoration={brace,amplitude=4pt,raise=2pt}]
		(b-1-4.south west) -- (b-1-6.north west) node [black,midway,xshift=-1cm]   {Block $LR_3$};
	\draw [decorate,decoration={brace,amplitude=4pt,raise=2pt}]
		(c-1-7.south west) -- (c-1-9.north west) node [black,midway,xshift=-1cm]   {Block $LR_2$};
	\draw [decorate,decoration={brace,amplitude=4pt,raise=2pt}]
		(d-1-10.south west) -- (d-1-12.north west) node [black,midway,xshift=-1cm] {Block $LR_1$};
		
	\foreach \y in {10,...,12}{
		\draw[->,thick, draw=KITred] (0.125,0.25 * \y) -- (4.625, 0.25 * \y);
	}
	
	\foreach \y in {7,...,9}{
		\draw[->,thick,draw=KITgreen] (0.125,0.25 * \y) -- (4.625, 0.25 * \y);
	}
	
	\foreach \y in {4,...,6}{
		\draw[->,thick,draw=KITcyanblue] (0.125,0.25 * \y) -- (4.625, 0.25 * \y);
	}
		
	\foreach \y in {1,...,3}{
		\draw[->,thick,draw=KITblue] (0.125,0.25 * \y) -- (4.625, 0.25 * \y);
	}


		\foreach \x in {1,...,20}{
		\foreach \y in {1,...,3}{
			\node[rectangle, draw, opacity=0.2] (e-\x-\y) at (6+0.25 * \x, 0.25 * \y) {};
		}
	}
	
	\foreach \x in {1,...,20}{
		\foreach \y in {4,...,6}{
			\node[rectangle, draw, opacity=0.2] (f-\x-\y) at (6+0.25 * \x, 0.25 * \y) {};
		}
	}
	
	\foreach \x in {1,...,20}{
		\foreach \y in {7,...,9}{
			\node[rectangle, draw, opacity=0.2] (g-\x-\y) at (6+0.25 * \x, 0.25 * \y) {};
		}
	}
	
	\foreach \x in {1,...,20}{
		\foreach \y in {10,...,12}{
			\node[rectangle, draw, opacity=0.2] (h-\x-\y) at (6+0.25 * \x, 0.25 * \y) {};
		}
	}
	
	\draw [decorate,decoration={brace,amplitude=4pt,raise=2pt}]
		(h-1-12.north west) -- (h-5-12.north east) node [black,midway,yshift=0.5cm]     {$TLBR_1$};
	\draw [decorate,decoration={brace,amplitude=4pt,raise=2pt}]
		(h-6-12.north west) -- (h-10-12.north east) node [black,midway,yshift=0.5cm]    {$TLBR_2$};
	\draw [decorate,decoration={brace,amplitude=4pt,raise=2pt}]
		(h-11-12.north west) -- (h-15-12.north east) node [black,midway,yshift=0.5cm]   {$TLBR_3$};
	\draw [decorate,decoration={brace,amplitude=4pt,raise=2pt}]
		(h-16-12.north west) -- (h-20-12.north east) node [black,midway,yshift=0.5cm]   {$TLBR_4$};

	\foreach \x in {1,...,12}{
			\node[rectangle, fill=KITblue, opacity=0.2] (i-1-\x) at (6 + \x * 0.25 + 0.00, 3.25 - \x * 0.25) {};
			\node[rectangle, fill=KITblue, opacity=0.2] (i-2-\x) at (6 + \x * 0.25 + 0.25, 3.25 - \x * 0.25) {};
			\node[rectangle, fill=KITblue, opacity=0.2] (i-3-\x) at (6 + \x * 0.25 + 0.50, 3.25 - \x * 0.25) {};
			\node[rectangle, fill=KITblue, opacity=0.2] (i-4-\x) at (6 + \x * 0.25 + 0.75, 3.25 - \x * 0.25) {};
			\node[rectangle, fill=KITblue, opacity=0.2] (i-5-\x) at (6 + \x * 0.25 + 1.00, 3.25 - \x * 0.25) {};
	}
	
	\foreach \x in {1,...,12}{
			\node[rectangle, fill=KITcyanblue, opacity=0.2] (i-1-\x) at (6 + \x * 0.25 + 1.25, 3.25 - \x * 0.25) {};
			\node[rectangle, fill=KITcyanblue, opacity=0.2] (i-2-\x) at (6 + \x * 0.25 + 1.50, 3.25 - \x * 0.25) {};
			\node[rectangle, fill=KITcyanblue, opacity=0.2] (i-3-\x) at (6 + \x * 0.25 + 1.75, 3.25 - \x * 0.25) {};
			\node[rectangle, fill=KITcyanblue, opacity=0.2] (i-4-\x) at (6 + \x * 0.25 + 2.00, 3.25 - \x * 0.25) {};
	}
	
	\foreach \x in {1,...,11}{
		\node[rectangle, fill=KITcyanblue, opacity=0.2] (i-3-\x) at (6 + \x * 0.25 + 2.25, 3.25 - \x * 0.25) {};
	}
	\node[rectangle, fill=KITcyanblue, opacity=0.2] (i-2) at (e-1-1) {};
	
	\foreach \x in {1,...,10}{
		\node[rectangle, fill=KITgreen, opacity=0.2] (j-1-\x) at (6 + \x * 0.25 + 2.50, 3.25 - \x * 0.25) {};
	}
	\foreach \x in {1,...,9}{
		\node[rectangle, fill=KITgreen, opacity=0.2] (j-2-\x) at (6 + \x * 0.25 + 2.75, 3.25 - \x * 0.25) {};
	}
	\foreach \x in {1,...,8}{
		\node[rectangle, fill=KITgreen, opacity=0.2] (j-3-\x) at (6 + \x * 0.25 + 3.00, 3.25 - \x * 0.25) {};
	}
	\foreach \x in {1,...,7}{
		\node[rectangle, fill=KITgreen, opacity=0.2] (j-4-\x) at (6 + \x * 0.25 + 3.25, 3.25 - \x * 0.25) {};
	}
	\foreach \x in {1,...,6}{
		\node[rectangle, fill=KITgreen, opacity=0.2] (j-5-\x) at (6 + \x * 0.25 + 3.50, 3.25 - \x * 0.25) {};
	}
  \foreach \x in {1,...,2}{
		\node[rectangle, fill=KITgreen, opacity=0.2] (j-6-\x) at (6 + \x * 0.25, 0.75 - \x * 0.25) {};
	}
  \foreach \x in {1,...,3}{
		\node[rectangle, fill=KITgreen, opacity=0.2] (j-7-\x) at (6 + \x * 0.25, 1.00 - \x * 0.25) {};
	}
	\foreach \x in {1,...,4}{
		\node[rectangle, fill=KITgreen, opacity=0.2] (j-8-\x) at (6 + \x * 0.25, 1.25 - \x * 0.25) {};
	}
	\foreach \x in {1,...,5}{
		\node[rectangle, fill=KITgreen, opacity=0.2] (j-9-\x) at (6 + \x * 0.25, 1.50 - \x * 0.25) {};
	}
	\foreach \x in {1,...,6}{
		\node[rectangle, fill=KITgreen, opacity=0.2] (j-10-\x) at (6 + \x * 0.25, 1.75 - \x * 0.25) {};
	}
	
	 \foreach \x in {1,...,7}{
		\node[rectangle, fill=KITred, opacity=0.2] (k-6-\x) at (6 + \x * 0.25, 2.00 - \x * 0.25) {};
	}
  \foreach \x in {1,...,8}{
		\node[rectangle, fill=KITred, opacity=0.2] (k-7-\x) at (6 + \x * 0.25, 2.25 - \x * 0.25) {};
	}
	\foreach \x in {1,...,9}{
		\node[rectangle, fill=KITred, opacity=0.2] (k-8-\x) at (6 + \x * 0.25, 2.50 - \x * 0.25) {};
	}
	\foreach \x in {1,...,10}{
		\node[rectangle, fill=KITred, opacity=0.2] (k-9-\x) at (6 + \x * 0.25, 2.75 - \x * 0.25) {};
	}
	\foreach \x in {1,...,11}{
		\node[rectangle, fill=KITred, opacity=0.2] (k-10-\x) at (6 + \x * 0.25, 3.00 - \x * 0.25) {};
	}

	\foreach \x in {1,...,5}{
		\node[rectangle, fill=KITred, opacity=0.2] (g-1-\x) at (6 + \x * 0.25 + 3.75, 3.25 - \x * 0.25) {};
	}
	\foreach \x in {1,...,4}{
		\node[rectangle, fill=KITred, opacity=0.2] (g-2-\x) at (6 + \x * 0.25 + 4.00, 3.25 - \x * 0.25) {};
	}
	\foreach \x in {1,...,3}{
		\node[rectangle, fill=KITred, opacity=0.2] (g-3-\x) at (6 + \x * 0.25 + 4.25, 3.25 - \x * 0.25) {};
	}
	\foreach \x in {1,...,2}{
		\node[rectangle, fill=KITred, opacity=0.2] (g-4-\x) at (6 + \x * 0.25 + 4.50, 3.25 - \x * 0.25) {};
	}
	\foreach \x in {1,...,1}{
		\node[rectangle, fill=KITred, opacity=0.2] (g-5-\x) at (6 + \x * 0.25 + 4.75, 3.25 - \x * 0.25) {};
	}

	\draw[->, thick, draw=KITcyanblue] (6.125,0.25) -- (6.25,0.125);
	\foreach \x in {2,...,6}{
		\draw[->,thick,draw=KITgreen] (6.125, \x * 0.25) -- (6 + \x * 0.25,0.125);
	}
	\foreach \x in {7,...,11}{
		\draw[->,thick,draw=KITred] (6.125, \x * 0.25) -- (6 + \x * 0.25,0.125);
	}
	
	\foreach \x in {1,...,5}{
		\draw[->,thick,draw=KITblue] (6 + 0.25 * \x - 0.125,3) -- (6 + 2.75 + \x * 0.25,0.125);
	}
	\foreach \x in {6,...,9}{
		\draw[->,thick,draw=KITcyanblue] (6 + 0.25 * \x - 0.125,3) -- (6 + 2.75 + \x * 0.25,0.125);
	}

	\draw[->,thick,draw=KITcyanblue] (8.375, 3) -- (11.00,0.375);
	
	\foreach \x in {1,...,5} {
		\draw[->,thick,draw=KITgreen] (8.375 + \x * 0.25, 3) -- (11.00,0.375 + \x * 0.25);
	}
	
	\foreach \x in {6,...,10} {
		\draw[->,thick,draw=KITred] (8.375 + \x * 0.25, 3) -- (11.00,0.375 + \x * 0.25);
	}
	
\end{tikzpicture} %
	\caption { %
		Multiple CUDA blocks processing the \gls*{SGM} path aggregation. %
		Each block calculates 16 different lines along one path direction. %
	} %
	\label{fig:cuda_sgm_impl_1} %
\end{figure} %

The calculation of the eight different \gls*{SGM} path costs is done sequentially on CUDA hardware. %
The parallelization of the cost aggregation on one path direction is realized on two different levels. %
First, each CUDA block calculates the costs for 16 different lines along one path direction (\Cref{fig:cuda_sgm_impl_1}). %
If a diagonal line reaches the image border, the values are reset and the calculation is resumed on the other side of the image. %
This ensures that all calculations of one path direction takes the same time. %
Additionally, for each image point, the costs for the disparities $d_0, ...\ ,(\frac{d_\mathrm{max}}{2}-1)$ and $\frac{d_\mathrm{max}}{2}, ...\ ,d_\mathrm{max}$ are calculated in parallel by two iterations. %
This ensures that all threads within one warp access a contiguous area in the memory, allowing the memory transactions to be more efficient. %
However, this requires a synchronization of the threads within a CUDA warp after the costs for all disparities have been calculated, in order to find the minimum path cost, which is necessary for further processing. %
To find the minimum of the aggregated costs, we utilize the map reduce method as illustrated in \Cref{sec:appendix_mapreduce}. %
After the calculation and aggregation of the different \gls*{SGM} path costs, the \gls*{WTA} disparity with the minimum aggregated costs is to be found. %
This is done by assigning a specific image region to each CUDA warp and again utilizing the map reduce method mentioned above. %

\subsubsection{Consistency check}

\begin{figure}[b!] %
	\centering %
	\resizebox{0.75\textwidth}{!}{\begin{tikzpicture}
	
\foreach \x in {1,...,10}{
	\node[rectangle, draw, opacity=0.5] (c-\x) at (0.25 * \x, 2) {};
}

\foreach \x in {11,...,20}{
	\node[rectangle, draw, opacity=0.5] (c-\x) at (1 + 0.25 * \x, 2) {};
}

\foreach \x in {21,...,30}{
	\node[rectangle, draw, opacity=0.5] (c-\x) at (2 + 0.25 * \x, 2) {};
}

\foreach \x in {31,...,40}{
	\node[rectangle, draw, opacity=0.5] (c-\x) at (3 + 0.25 * \x, 2) {};
}

\node (t-1) at (3.1,2)  {$...$};
\node (t-2) at (6.6,2)  {$...$};
\node (t-3) at (10.1,2) {$...$};

\coordinate (a) at (0.125,1.5);
\coordinate (b) at (0.125,2.5);
\draw[dashed] (a) -- (b);

\coordinate (c) at (3.625,1.5);
\coordinate (d) at (3.625,2.5);
\draw[dashed] (c) -- (d);

\coordinate (e) at (7.125,1.5);
\coordinate (f) at (7.125,2.5);
\draw[dashed] (e) -- (f);

\coordinate (g) at (10.625,1.5);
\coordinate (h) at (10.625,2.5);
\draw[dashed] (g) -- (h);

\node[anchor=north] (t1) at (0.125, 1.5) {$\bar{\mathcal{S}}((x,y),0)$};
\node[anchor=north] (t2) at (3.625, 1.5) {$\bar{\mathcal{S}}((x+1,y),0)$};
\node[anchor=north] (t3) at (7.125, 1.5) {$\bar{\mathcal{S}}((x+2,y),0)$};
\node[anchor=north] (t4) at (10.625,1.5) {$\bar{\mathcal{S}}((x+3,y),0)$};

\node[fill=KITred] (b1) at (c-1) {};
\node[fill=KITred] (b2) at (c-12){};
\node[fill=KITred] (b3) at (c-23){};
\node[fill=KITred] (b4) at (c-34){};

\node (d1) at (2.2,3.4) {$d_\mathrm{max} + 1$};
\node (d2) at (5.8,3.4) {$d_\mathrm{max} + 1$};
\node (d3) at (9.5,3.4) {$d_\mathrm{max} + 1$};

\draw (b1) to[out=75,in=115] (b2);
\draw (b2) to[out=75,in=115] (b3);
\draw (b3) to[out=75,in=115] (b4);

\draw[dashed, draw=KITred, thick] (b1.north) -- (0.25, 3.5);
\draw[dashed, draw=KITred, thick] (b2.north) -- (4,    3.5);
\draw[dashed, draw=KITred, thick] (b3.north) -- (7.75, 3.5);
\draw[dashed, draw=KITred, thick] (b4.north) -- (11.5, 3.5);

\node[anchor=south, text=KITred] (f1) at (0.25,3.5) {$\bar{\mathcal{S}}^{\mathrm{R}}((x,y),0)$};
\node[anchor=south, text=KITred] (f2) at (4, 3.5)   {$\bar{\mathcal{S}}^{\mathrm{R}}((x,y),1)$};
\node[anchor=south, text=KITred] (f3) at (7.75,3.5) {$\bar{\mathcal{S}}^{\mathrm{R}}((x,y),2)$};
\node[anchor=south, text=KITred] (f4) at (11.5,3.5) {$\bar{\mathcal{S}}^{\mathrm{R}}((x,y),3)$};

\end{tikzpicture}} %
	\caption { %
		In the consistency check, an additional disparity map, which corresponds to the matching image, is approximated from the aggregated cost volume $\bar{\mathcal{S}}$. %
		The required entries are not situated directly next to each other, which hinders an efficient memory access. %
	} %
	\label{fig:impl_consistency_probl} %
\end{figure}
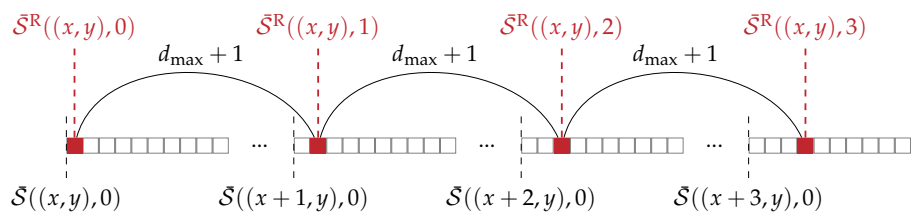

The key aspect in the consistency check is the calculation of the approximated disparity map $D^{\mathrm{R}}$, corresponding to the matching image. %
This is approximated from the calculated aggregated cost volume of the reference image $\bar{\mathcal{S}}$. %
In this, each entry of $D^{\mathrm{R}}$ is calculated according to \Cref{eq:right_disparity}. %
The difficulty, that arises in this process, is the access of non-adjacent areas in $\bar{\mathcal{S}}$, as illustrated by \Cref{fig:impl_consistency_probl}. %
Between all entries of the aggregated cost volume $\bar{\mathcal{S}}$, that are to be used for the calculation of $D^{\mathrm{R}}$, always lie $d_{max} + 1$ entries, which are of no interest. %

In the GPU implementation for the consistency check, each instantiation of a CUDA kernel computes a specified region in the approximated disparity map $D^\mathrm{R}$.
In this, the data of the aggregated cost volume is first copied into shared memory for quicker access. %
Since the data does not lie next to each other, the access to the global memory by the different CUDA threads cannot be pooled together. %
When all data is available in the shared memory, we again use the map reduce method to find the minimum. %
After the disparity map $D^\mathrm{R}$ for the matching image is approximated, each thread performs the consistency check according to \Cref{eq:LR_check} for a specific pixel in the final disparity map $D^{\mathrm{L}}$. %

\subsubsection{Median filter}

The execution of the median filter on the GPU is straightforward. %
To each CUDA thread block a region in the final disparity map is assigned for which the filter is to be processed. %
In this, the necessary data is first copied to the shared memory. %
Pixels that lie outside the image get a disparity value of 0xFFFF assigned, making them irrelevant for the filtering. %
With all the data in the shared memory, each thread of the thread block calculates the median for a pixel. %
In this, the first five iterations of the bubble sort algorithm are performed to sort the values in the $3\times3$\px neighborhood. %
After the fifth iteration, the five highest values are correctly sorted and the median can be extracted. %

\subsection{Vectorized SIMD processing with NEON intrinsic set on ARM CPUs}
\label{sec:methodology-cpu}

\glsreset{SIMD}

\begin{figure}[t!] %
	\centering %
	\resizebox{0.8\columnwidth}{!}{\begin{tikzpicture}

\tikzstyle{thread} = [fill=KITcyanblue, rectangle, 
    minimum height=1.5em, minimum width=15em, fill opacity=0.2, text opacity=1.0]

\node[rectangle, minimum height=9em, minimum width=14em, draw] (backnode) at (0.9,7) {};
\node[rectangle, fill=KITblue, minimum height=2.125, minimum width=14em, fill opacity=0.2, text opacity=1.0] (thread1) at (0.9,8.35) {Thread 1};
\node[rectangle, fill=KITcyanblue, minimum height=2.125, minimum width=14em, fill opacity=0.2, text opacity=1.0] (thread2) at (0.9,7.45) {Thread 2};
\node[rectangle, fill=white, minimum height=2.125, minimum width=14em, fill opacity=0.2, text opacity=1.0] (thread3) at (0.9,6.55) {$\vdots$};
\node[rectangle, fill=KITred, minimum height=2.125, minimum width=14em, fill opacity=0.2, text opacity=1.0] (threadN) at (0.9,5.65) {Thread $n$};

\node[fill=KITyellow, rectangle, minimum width = 10em, fill opacity = 0.2, text opacity=1.0] (instructions) at (8,9) {SIMD Instructions};
\node[fill=KITyellow, rectangle, minimum width = 3em,  fill opacity = 0.2, text opacity=1.0] (pu1) at (8,8) {PU};
\node[fill=KITyellow, rectangle, minimum width = 3em,  fill opacity = 0.2, text opacity=1.0] (pu2) at (8,7) {PU};
\node[fill=KITyellow, rectangle, minimum width = 3em,  fill opacity = 0.2, text opacity=1.0] (pu3) at (8,6) {PU};
\node[fill=KITyellow, rectangle, minimum width = 3em,  fill opacity = 0.2, text opacity=1.0] (pu4) at (8,5) {PU};

\node[minimum width=9em, fill=KITyellow, rectangle, fill opacity = 0.2, text opacity=1.0, rotate=90] (mem) at (6.3,6.5) {Memory};

\draw[ultra thick]  (9.5,8.75) -- (9.5,5);
\draw[->, ultra thick] (9.5,8) -- (pu1);
\draw[->, ultra thick] (9.5,7) -- (pu2);
\draw[->, ultra thick] (9.5,6) -- (pu3);
\draw[->, ultra thick] (9.5,5) -- (pu4);

\draw[->, ultra thick] (pu1) -- (6.7,8);
\draw[->, ultra thick] (pu2) -- (6.7,7);
\draw[->, ultra thick] (pu3) -- (6.7,6);
\draw[->, ultra thick] (pu4) -- (6.7,5);


\node[draw, dashed, rectangle, minimum height = 13em, minimum width = 10.8em] (box) at (7.9,7.0) {};
\node[] (cnt) at (8,9.7) {Thread 2 - NEON Vector-Processor};

\draw[dashed, ultra thick, color=KITcyanblue] (thread2.east) -- (box.north west);
\draw[dashed, ultra thick, color=KITcyanblue] (thread2.east) -- (box.south west);

\node[rectangle, fill=KITblue, draw] (cpu1) at (-3,8.35) {CPU 1};
\node[rectangle, fill=KITcyanblue, draw] (cpu2) at (-3,7.45) {CPU 2};
\node[rectangle, fill=KITred, draw] (cpuN) at (-3,5.65) {CPU $n$};

\draw[->, ultra thick] (cpu1.east) -- (thread1.west);
\draw[->, ultra thick] (cpu2.east) -- (thread2.west);
\draw[->, ultra thick] (cpuN.east) -- (threadN.west);

\end{tikzpicture}} %
	\caption{ %
		In the optimization of the \gls*{SGM} stereo algorithm for the execution on embedded CPUs, two different parallelization strategies are utilized. %
		The implementation uses multiple threads to evenly distribute the processing on the available CPU cores. %
		Each thread uses the AMD NEON instruction set to perform a vectorized \gls*{SIMD} processing by using the NEON processing units (PU). %
	} %
	\label{fig:impl_cpu_general} %
\end{figure}
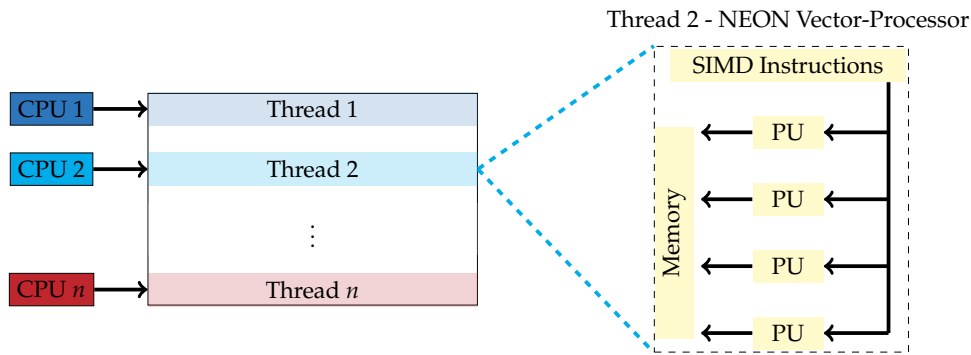

In order to efficiently deploy the real-time processing pipeline for the estimation of dense disparity maps on an embedded CPU, such as the 8-core ARMv8.2 on the NVIDIA Jetson Xavier AGX, we utilize two strategies of parallelization as illustrated in \Cref{fig:impl_cpu_general}, namely: %
\begin{enumerate} %
	\item a thread-level parallelization, and%
	\item a vectorized data processing with the \gls*{SIMD} NEON intrinsics. %
\end{enumerate}
Our implementation uses eight concurrent threads to efficiently utilize the available CPU cores. %
In each step of the processing pipeline, in exception to the \gls*{SGM} optimization step, each thread operates on a different image stripe, thus operating isolated and independently from the other threads. %
At two locations in the processing pipeline of the stereo algorithm, \ie before and after the \gls*{SGM} optimization, the concurrent threads need to be synchronized, since the \gls*{SGM} optimization relies on a different thread barrier than the other steps of the pipeline. %
In the other steps of the pipeline, the threads do not require any synchronization and can thus be processed fully concurrently. %
As part of our second parallelization strategy, each thread uses the ARM NEON instruction set \citep{Arm2013neon} to perform a vectorized \gls*{SIMD} processing on the vector-processors of the CPU (\cf \Cref{sec:appendix_cpu_processing}). %

In the following sections, we will discuss how we have optimized each step of the processing pipeline for execution on an ARMv8 CPU, utilizing thread parallelism and \gls*{SIMD} processing. %
Since the use of the \gls*{NCC} as a cost function for the image matching is computationally more expensive, and the frame rates achieved by the use of the CPU are anyway much lower than those achieved on the GPU, we refrained from implementing the \gls*{NCC} for a vectorized \gls*{SIMD} processing on the CPU. %

\subsubsection{Calculating the census transformation and its Hamming distance}
\label{sec:methodology-cpu-ct}
\glsreset{CT}

\begin{figure}[t] %
	\centering %
	\resizebox{0.6\columnwidth}{!}{\begin{tikzpicture}

\foreach \x in {1,...,48}
	\foreach \y in {8,...,20} 
		\node[rectangle, draw] (r-\x-\y) at (0.25 * \x, 0.25 * \y) {};
		
\foreach \x in {1,...,48}
	\foreach \y in {4,...,5} 
		\node[rectangle, draw, opacity=0.1] (r-\x-\y) at (0.25 * \x, 0.25 * \y) {};
		
\foreach \x in {1,...,48}
	\foreach \y in {6,...,7} 
		\node[rectangle, draw, opacity=0.3] (r-\x-\y) at (0.25 * \x, 0.25 * \y) {};
		
\foreach \x in {1,...,20}
	\foreach \y in {1,...,5} 
		\node[draw, rectangle, fill opacity=0.2, fill=KITred] (n-\x-\y) at (3.5 + 0.25 * \x, 2.5 + 0.25 * \y) {};
		
\foreach \y in {1,...,16} 
		\node[draw, rectangle, fill opacity=1.0, fill=KITblue] (c-\y) at (4 + 0.25 * \y, 3.25) {};
		
\foreach \x in {1,...,48}
	\foreach \y in {20,...,14} 
		\node[rectangle, fill=KITgreen, fill opacity=0.2] (d-\x-\y) at (0.25 * \x, 0.25 * \y) {};
		
\foreach \y in {1,...,16} 
	\node[rectangle, fill=KITgreen, fill opacity=0.2] (d-\y) at (0.25 * \y, 3.25) {};
	
\foreach \y in {33,...,48} 
	\node[rectangle, fill=KITyellow, fill opacity=0.2] (z-\y) at (0.25 * \y, 3.25) {};
		
\node[rectangle, fill=KITblue,   fill opacity=1.0] (label1) at (0.25, 0.5)   {};
\node[rectangle, fill=KITred,    fill opacity=0.2] (label2) at (0.25, 0)   {};
\node[rectangle, fill=KITgreen,  fill opacity=0.2] (label3) at (7.0,  0.5)   {};
\node[rectangle, fill=KITyellow, fill opacity=0.2] (label4) at (7.0,  0)   {};

\node[anchor=west] (text1) at (0.5, 0.5) {Currently treated reference pixel};
\node[anchor=west] (text2) at (0.5, 0) {Currently treated neighborhood pixel};
\node[anchor=west] (text3) at (7.25,0.5) {Previously processed pixel};
\node[anchor=west] (text4) at (7.25,0) {Pixel to be processed};

\draw[dashed, thick] (r-1-20.west)   -- (0.125, 5.75);
\draw[dashed, thick] (r-17-20.west)  -- (4.125, 5.75);
\draw[dashed, thick] (r-33-20.west)  -- (8.125, 5.75);
\draw[dashed, thick] (r-48-20.east) -- (12.125, 5.75);

\node[anchor=west] (it1) at (1, 5.5) {Iteration 1};
\node[anchor=west] (it2) at (5, 5.5) {Iteration 2};
\node[anchor=west] (it3) at (9, 5.5) {Iteration 3};

\end{tikzpicture}} %
	\caption{ %
		For the optimized CPU implementation, the \acrlong*{CT} is processed for multiple pixels simultaneously by utilizing the NEON vector processing units. %
		In this, a sliding window is used, which processes the image data from left-to-right and from top-to-bottom. %
	} %
	\label{fig:neon_impl_census_1} %
\end{figure}
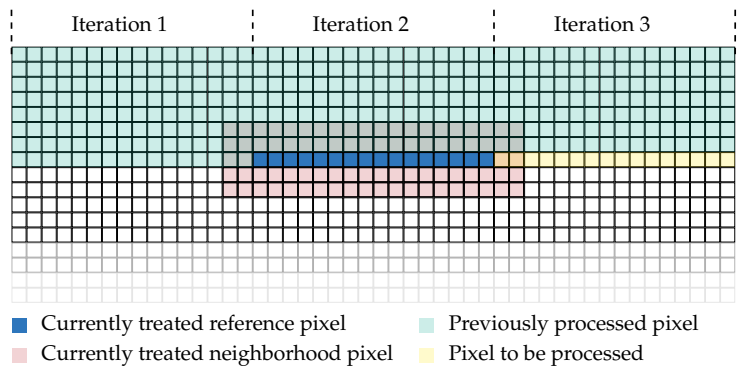 %
As illustrated by \Cref{fig:neon_impl_census_1}, we can calculate the \gls*{CT} in each thread for 16 pixels simultaneously, using the \gls*{SIMD} vector-registers. %
In this, we load the 16 reference pixels with eight bits each into one vector-register. %
In addition, we load the corresponding 24$\times$16 neighbor pixels into one vector-register each, resulting in a total use of 25 vector-registers with 16 lanes. %
We process the full image by sliding the illustrated window from left-to-right and top-to-bottom over the image. %
In doing so, there is a good chance that the image data, which is needed by the next iteration, is already cached. %
In the calculation of the \gls*{CT}, we omit the comparison of the reference pixel with itself. %
This allows us to represent the 24\,bits of the resulting \gls*{CT} bitstring by three bytes and thus store the census descriptors of all 16 pixels in three vector-registers.
This is also why we only consider a $5\times5$\px neighborhood in the optimized implementation of the \gls*{CT} for the ARM CPU. %
In order to also calculate the \gls*{CT} at the image border, where a part of the neighborhood lies outside of the image, we would need to introduce a conditional statement, which is not recommended when using \gls*{SIMD} vectorization. %
Thus, we only calculate the \gls*{CT} up to two pixels with respect to the image border, reducing the image size by four pixels in each dimension for the subsequent processing. %
\begin{figure}[b!] %
	\centering %
	\resizebox{0.75\columnwidth}{!}{\begin{tikzpicture}


\foreach \x in {1,...,2}{
	\foreach \y in {1,...,10}{
		{\node[rectangle, draw, opacity=0.1] (r-\x-\y) at (0.25 * \x, 0.25 * \y) {};}
	}
}

\foreach \x in {3,...,4}{
	\foreach \y in {1,...,10}{
		{\node[rectangle, draw, opacity=0.3] (r-\x-\y) at (0.25 * \x, 0.25 * \y) {};}
	}
}

\foreach \x in {5,...,22}{
	\foreach \y in {1,...,10}{
		{\node[rectangle, draw] (r-\x-\y) at (0.25 * \x, 0.25 * \y) {};}
	}
}

\node[rectangle, fill=KITblue, fill opacity=0.2] (ref) at (r-20-8) {};

\foreach \x in {1,...,2}{
	\foreach \y in {1,...,10}{
		{\node[rectangle, draw, opacity=0.1] (m-\x-\y) at (7 + 0.25 * \x, 0.25 * \y) {};}
	}
}

\foreach \x in {3,...,4}{
	\foreach \y in {1,...,10}{
		{\node[rectangle, draw, opacity=0.3] (m-\x-\y) at (7 + 0.25 * \x, 0.25 * \y) {};}
	}
}

\foreach \x in {5,...,22}{
	\foreach \y in {1,...,10}{
		{\node[rectangle, draw] (m-\x-\y) at (7 + 0.25 * \x, 0.25 * \y) {};}
	}
}
		
\foreach \x in {1,...,16} {
	\node[rectangle, fill=KITred, fill opacity=0.2] (n-\x) at (8 + \x * 0.25, 2) {};
}

\node[draw, rounded rectangle, fill=KITblue, fill opacity=0.2, text opacity=1.0, minimum width=4.5cm] (refVec) at (3.1,4) {Reference bitstrings};
\node[draw, rounded rectangle, fill=KITred, fill opacity=0.2, text opacity=1.0, minimum width=4.5cm] (matVec) at (10.1,4) {Matching bitstrings};

\foreach \x in {1,...,16} {
	\draw[->, thick, draw=KITred] (n-\x.north) -- (8 + \x * 0.25, 3.65);
}

\coordinate (refEdge1) at (5,3.2);
\coordinate (refEdge2) at (1.25,3.2);

\draw[draw=KITblue, thick] (ref.north) -| (refEdge1.south) -- (refEdge2.east);

\foreach \x in {1,...,16} {
	\draw[->, thick, draw=KITblue] (5.25 - 0.25 * \x, 3.2) -- (5.25 - 0.25 * \x,3.65);
}

\node[rectangle, draw, fill=KITgreen, fill opacity=0.2, text opacity=1.0] (xor) at (6.5,4){XOR};	
\draw[->,draw=KITblue, thick] (refVec.east) -- (xor.west);
\draw[->,draw=KITred, thick]  (matVec.west) -- (xor.east);

\node[draw, rounded rectangle, fill=KITgreen, fill opacity=0.2, text opacity=1.0, minimum width=4.5cm] (tempVec1) at (6.5,5.3) {Temporary vector 1};
\coordinate (tmpEdge1) at (6.5,4.6);
\coordinate (tmpEdge2) at (4.625,4.6);
\coordinate (tmpEdge3) at (8.4,4.6);
\draw[draw=KITgreen, thick] (xor.north) -| (tmpEdge1.south) -- (tmpEdge2.east);
\draw[draw=KITgreen, thick] (xor.north) -| (tmpEdge1.south) -- (tmpEdge3.west);

\foreach \x in {1,...,16} {
	\draw[->, thick, draw=KITgreen] (4.375 + 0.25 * \x, 4.6) -- (4.375 + 0.25 * \x, 4.9);
}

\node[rectangle, draw, fill=KITorange, fill opacity=0.2, text opacity=1.0] (vcnt) at (6.5,6.15){VCNT};	
\draw[->,draw=KITorange, thick] (tempVec1.north) -- (vcnt.south);

\node[draw, rounded rectangle, fill=KITorange, fill opacity=0.2, text opacity=1.0, minimum width=4.5cm] (tempVec2) at (6.5,7.3) {Temporary vector 2};
\coordinate (tmpEdge4) at (6.5,6.6);
\coordinate (tmpEdge5) at (4.625,6.6);
\coordinate (tmpEdge6) at (8.4,6.6);
\draw[draw=KITorange, thick] (vcnt.north) -| (tmpEdge4.south) -- (tmpEdge5.east);
\draw[draw=KITorange, thick] (vcnt.north) -| (tmpEdge4.south) -- (tmpEdge6.west);

\foreach \x in {1,...,16} {
	\draw[->, thick, draw=KITorange] (4.375 + 0.25 * \x, 6.6) -- (4.375 + 0.25 * \x, 6.9);
}

\node[draw, rounded rectangle, fill=KITyellow, fill opacity=0.2, text opacity=1.0, minimum width=4.5cm] (mem) at (6.5,8.5) {Memory};

\foreach \x in {1,...,16} {
	\draw[->, thick, draw=KITorange] (4.375 + 0.25 * \x, 7.6) -- (4.375 + 0.25 * \x, 8.1);
}

\node (ctLeft)  at (3,-0.5)  {$I^\mathrm{L}$};
\node (ctRight) at (10,-0.5) {$I^\mathrm{R}$};



\end{tikzpicture}} %
	\caption { %
		Overview on the calculation of the Hamming distance with \gls*{SIMD} intrinsics. %
		Each census descriptor is loaded from $I^\mathrm{L}$ and $I^\mathrm{R}$ in separate vector-registers, on which a XOR operation is applied. %
		The number of set bits inside a vector-register is counted by using the NEON hardware instruction VCNT (Vector Count Set Bits). %
	} %
	\label{fig:neon_impl_hamming_1} %
\end{figure}
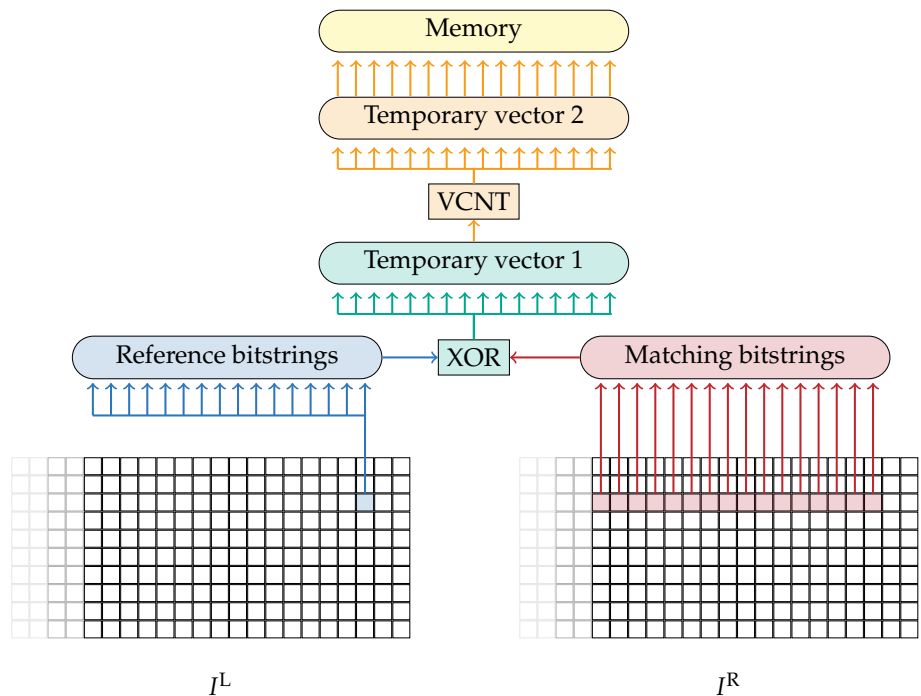

To calculate the Hamming distance between two census descriptors, namely the one from the pixel of the reference image and the corresponding pixel in the matching image, we apply the XOR operator and count how many bits are set to 1 in the final output. %
To count the number of bits which are set, the NEON instruction set offers a population count (VCNT) which can be applied to each vector-lane. %
As illustrated by \Cref{fig:neon_impl_hamming_1}, the resulting matching cost, \ie the Hamming distance of the \gls*{CT}, is stored to a three-dimensional cost volume. 
By utilizing the \gls*{SIMD} instructions and the 32 vector-register, a maximum of 64 matching costs can be calculated simultaneously in each thread. %
Thus, each thread processes 16 disparities and four lines simultaneously in one iteration. %

In case the currently calculated disparity is bigger than the x-coordinate of the reference pixel, the corresponding matching pixel lies outside the image. %
In order to efficiently handle this case, we additionally store the disparity for which the matching costs are currently being calculated as well as the x-coordinate of the currently processed pixel in two additional vector-registers. %
In each iteration, both of the above mentioned registers are compared against each other and the result is stored in a third register. %
If the disparity is greater than the x-coordinate of the pixel, all bits inside the vector-lanes will be set. %
Finally, if we apply an OR operation between the register with the matching cost and the register with the comparison result, the matching cost of each disparity that spans over the image boundary will be set to 0xFF and thus will not contribute in the subsequent search for the optimum. %

\subsubsection{Semi-Global Matching optimization}
\label{sec:methodology-cpu-sgm}

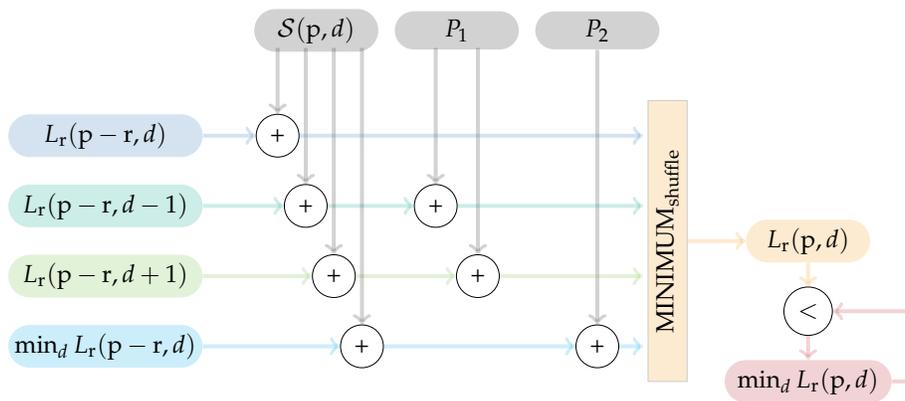
\begin{figure}[t!] %
	\centering %
	\resizebox{0.75\columnwidth}{!}{\begin{tikzpicture}

\node[rounded rectangle, fill=KITblue, minimum width=3cm, fill opacity=0.2, text opacity=1.0]
		 (dVec) at (1,5) {$L_\mathrm{r}(\mathrm{p-r}, d)$};
\node[rounded rectangle, fill=KITgreen, minimum width=3cm, fill opacity=0.2, text opacity=1.0]
		 (dVec1) at (1,4) {$L_\mathrm{r}(\mathrm{p-r}, d-1)$};
\node[rounded rectangle, fill=KITpalegreen, minimum width=3cm, fill opacity=0.2, text opacity=1.0]
		 (dVec2) at (1,3) {$L_\mathrm{r}(\mathrm{p-r}, d+1)$};
\node[rounded rectangle, fill=KITcyanblue, minimum width=3cm, fill opacity=0.2, text opacity=1.0]
		 (dVec3) at (1,2) {$\min_d L_\mathrm{r}(\mathrm{p-r}, d)$};
\node[rounded rectangle, fill=KITblack, minimum width=2cm, fill opacity=0.2, text opacity=1.0]
		 (p0) at (4,6.5) {$\mathcal{S}(\mathrm{p}, d)$};
\node[rounded rectangle, fill=KITblack, minimum width=2cm, fill opacity=0.2, text opacity=1.0]
		 (p1) at (6,6.5) {$P_1$};
\node[rounded rectangle, fill=KITblack, minimum width=2cm, fill opacity=0.2, text opacity=1.0]
		 (p2) at (8,6.5) {$P_2$};

\node[circle, draw] (plus1) at (5.7,4) {+};
\node[circle, draw] (plus2) at (6.3,3) {+};
\node[circle, draw] (plus3) at (8,2)   {+};

\node[circle, draw] (plus4) at (3.45,5)   {+};
\node[circle, draw] (plus5) at (3.85,4)   {+};
\node[circle, draw] (plus6) at (4.25,3)   {+};
\node[circle, draw] (plus7) at (4.65,2)   {+};

\draw[->, ultra thick, draw=KITblue,      opacity=0.2]  (dVec.east)  -- (plus4.west);
\draw[->, ultra thick, draw=KITgreen, opacity=0.2]      (dVec1.east) -- (plus5.west);
\draw[->, ultra thick, draw=KITpalegreen, opacity=0.2]  (dVec2.east) -- (plus6.west);
\draw[->, ultra thick, draw=KITcyanblue, opacity=0.2]   (dVec3.east) -- (plus7.west);

\draw[->, ultra thick, draw=KITgreen, opacity=0.2]      (plus5.east) -- (plus1.west);
\draw[->, ultra thick, draw=KITpalegreen, opacity=0.2]  (plus6.east) -- (plus2.west);
\draw[->, ultra thick, draw=KITcyanblue, opacity=0.2]   (plus7.east) -- (plus3.west);

\draw[->, ultra thick, opacity=0.2, draw=KITblack] (3.45, 6.25)    -- (plus4.north);
\draw[->, ultra thick, opacity=0.2, draw=KITblack] (3.85, 6.25)    -- (plus5.north);
\draw[->, ultra thick, opacity=0.2, draw=KITblack] (4.25, 6.25)    -- (plus6.north);
\draw[->, ultra thick, opacity=0.2, draw=KITblack] (4.65, 6.25)    -- (plus7.north);

\draw[->, ultra thick, opacity=0.2, draw=KITblack] (5.7, 6.25)    -- (plus1.north);
\draw[->, ultra thick, opacity=0.2, draw=KITblack] (6.3, 6.25)    -- (plus2.north);
\draw[->, ultra thick, opacity=0.2, draw=KITblack] (p2.south)    -- (plus3.north);

\draw[->, ultra thick, draw=KITblue,      opacity=0.2]  (plus4.east)  -- (8.65,5);
\draw[->, ultra thick, draw=KITgreen,     opacity=0.2]  (plus1.east) -- (8.65,4);
\draw[->, ultra thick, draw=KITpalegreen, opacity=0.2]  (plus2.east) -- (8.65,3);
\draw[->, ultra thick, draw=KITcyanblue,  opacity=0.2]  (plus3.east) -- (8.65,2);

\node[rectangle, draw, opacity=0.2, fill=KITorange, minimum width=4cm, rotate=90, text opacity=1.0] (minShuffle) at (9,3.5) {$\text{MINIMUM}_\text{shuffle}$};

\node[rounded rectangle, fill=KITorange, minimum width=2cm, fill opacity=0.2, text opacity=1.0]
		 (result) at  (11,3.5) {$L_\mathrm{r}(\mathrm{p},d)$};
\node[rounded rectangle, fill=KITred, minimum width=2cm, fill opacity=0.2, text opacity=1.0]
		 (currMin) at (11,1.5)   {$\min_d L_\mathrm{r}(\mathrm{p}, d)$};

\draw[->, ultra thick, draw=KITorange,  opacity=0.2]  (minShuffle.south) -- (result.west);

\node[circle, draw] (lt) at (11, 2.5) {$<$};

\draw[->, ultra thick, draw=KITorange,  opacity=0.2]  (result.south) -- (lt.north);
\draw[->, ultra thick, draw=KITred,  opacity=0.2]     (lt.south) -- (currMin.north);

\coordinate (edge1) at (12.5,1.5);
\coordinate (edge2) at (12.5,2.5);

\draw[->, ultra thick, draw=KITred,  opacity=0.2] (currMin.east) -| (edge1) -| (edge2) -- (lt.east);

\end{tikzpicture}} %
	\caption { %
		Schematic overview of the implementation of the \gls*{SGM} aggregation at a single pixel on a path. %
		All components of the \gls*{SGM} path aggregation (\cf \Cref{eq:sgm_path}) are calculated simultaneously. %
		A final minimum operation will yield the result which is stored in the aggregated cost volume.
	} %
	\label{fig:neon_impl_paths} %
\end{figure}

The optimized implementation of the \gls*{SGM} algorithm can be divided into two separate steps. %
First, each thread calculates the \gls*{SGM} path costs for each path that is assigned to it according to \Cref{eq:sgm_path}. %
As illustrated by \Cref{fig:neon_impl_paths}, the four vector-registers are first filled with the results of the previous iteration, so that the vector-register $L_\mathrm{r}(\mathrm{p-r}, d)$ will hold the previous path costs at the same disparity level, the vector-registers $L_\mathrm{r}(\mathrm{p-r}, d-1)$ and $L_\mathrm{r}(\mathrm{p-r}, d+1)$ will hold the previous path costs at the disparity level $\pm 1$, and the vector-register $\min_d L_\mathrm{r}(\mathrm{p-r}, d)$ will hold the minimum path costs over all disparity levels at the considered pixel. %
According to \Cref{eq:sgm_path}, the current matching costs from the cost volume as well as the penalties are added to the different vector-registers. %
Again, the NEON instruction set provides a method to get the minimum from the four vector-registers, namely $\text{MINIMUM}_\text{shuffle}$. %
The result is stored in the allocated memory and is compared to the path costs of the other disparities in order to get the minimum path cost for the next iteration. %
\begin{figure}[b!] %
	\centering %
	\subfloat[Straight paths]{\begin{tikzpicture}


\foreach \x in {1,...,18}{
	\foreach \y in {1,...,10}{
		\node[rectangle, draw, opacity=0.5] (r-\x-\y) at (0.25 * \x, 0.25 * \y) {};
	}
}

\node[rectangle, fill=KITred, opacity=0.5]  (px1) at (r-5-4) {};
\node[rectangle, fill=KITred, opacity=0.5]  (px2) at (r-5-3) {};
\node[rectangle, fill=KITcyanblue, opacity=0.5] (px5) at (r-6-4) {};
\node[rectangle, fill=KITcyanblue, opacity=0.5] (px6) at (r-6-3) {};

\draw[->, thick, draw=KITred] (px5) -- (r-18-4.east);
\draw[->, thick, draw=KITred] (px6) -- (r-18-3.east);

\foreach \x in {18,...,1}{
	\foreach \y in {5,...,10}{
		\node[rectangle, fill=KITgreen, opacity=0.2] (b-\x-\y) at (0.25 * \x, 0.25 * \y) {};
	}
}

\foreach \x in {1,...,4}{
	\node[rectangle, fill=KITgreen, opacity=0.2] (a-\x) at (0.25 * \x, 1) {};
	\node[rectangle, fill=KITgreen, opacity=0.2] (a-\x) at (0.25 * \x, 0.75) {};
}

\node[rectangle, fill=KITblack, opacity=0.5] (px3) at (r-1-10) {};
\node[rectangle, fill=KITblack, opacity=0.5] (px4) at (r-1-9)  {};

\foreach \x in {1,...,18}{
	\node[rectangle, fill=KITorange, opacity=0.5] (u-\x) at (0.25 * \x, 0.25) {};
	\node[rectangle, fill=KITorange, opacity=0.5] (u-\x) at (0.25 * \x, 0.5) {};
}

\draw[dashed, ->, thick] (0,2.625) -- (0,0.25);

\end{tikzpicture} \label{fig:neon_impl_path_move_1}} \qquad %
	\subfloat[Diagonal paths]{\begin{tikzpicture}
	

\foreach \x in {1,...,18}{
	\foreach \y in {1,...,10}{
		\node[rectangle, draw, opacity=0.5] (r-\x-\y) at (0.25 * \x, 0.25 * \y) {};
	}
}


\foreach \y in {1,...,7}{
	\node[rectangle, fill=KITgreen, opacity=0.2] (v1-\y) at (r-1-\y) {};
}

\foreach \y in {1,...,6}{
	\node[rectangle, fill=KITgreen, opacity=0.2] (v2-\y) at (r-2-\y) {};
}

\foreach \y in {1,...,4}{
	\node[rectangle, fill=KITgreen, opacity=0.2] (v3-\y) at (r-3-\y) {};
}

\foreach \y in {1,...,3}{
	\node[rectangle, fill=KITgreen, opacity=0.2] (v4-\y) at (r-4-\y) {};
}

\foreach \y in {1,...,2}{
	\node[rectangle, fill=KITgreen, opacity=0.2] (v5-\y) at (r-5-\y) {};
}

\node[rectangle, fill=KITgreen, opacity=0.2] (v6-1) at (r-6-1) {};

\foreach \y in {10,...,5}{
	\node[rectangle, fill=KITgreen, opacity=0.2] (v7-\y) at (r-18-\y) {};
}

\foreach \y in {10,...,6}{
	\node[rectangle, fill=KITgreen, opacity=0.2] (v8-\y) at (r-17-\y) {};
}

\foreach \y in {10,...,7}{
	\node[rectangle, fill=KITgreen, opacity=0.2] (v9-\y) at (r-16-\y) {};
}

\foreach \y in {10,...,8}{
	\node[rectangle, fill=KITgreen, opacity=0.2] (v10-\y) at (r-15-\y) {};
}

\foreach \y in {10,...,9}{
	\node[rectangle, fill=KITgreen, opacity=0.2] (v11-\y) at (r-14-\y) {};
}

\node[rectangle, fill=KITgreen, opacity=0.2] (v12-21) at (r-13-10) {};
\node[rectangle, fill=KITgreen, opacity=0.2] (v12-22) at (r-13-9) {};
\node[rectangle, fill=KITgreen, opacity=0.2] (v12-23) at (r-12-10) {};

\foreach \y in {8,...,1}{
	\node[rectangle, fill=KITorange, opacity=0.5] (z0-\y) at (2.25 - \y * 0.25, \y * 0.25) {};
}

\foreach \y in {10,...,3}{
	\node[rectangle, fill=KITorange, opacity=0.5] (z1-\y) at (\y * 0.25 + 2, 3.25 - \y * 0.25) {};
}


\node[rectangle, fill=KITred, opacity=0.5]   (px1) at (r-3-5)   {};
\node[rectangle, fill=KITred, opacity=0.5]   (px2) at (r-14-8)  {};
\node[rectangle, fill=KITcyanblue, opacity=0.5]   (px1) at (r-4-4)   {};
\node[rectangle, fill=KITcyanblue, opacity=0.5]   (px2) at (r-15-7)  {};
\node[rectangle, fill=KITblack, opacity=0.5] (px3) at (r-1-1)   {};
\node[rectangle, fill=KITblack, opacity=0.5] (px4) at (r-18-10) {};

\draw[->, thick, draw=KITred] (px1.south east) -- (r-7-1.south east);
\draw[->, thick, draw=KITred] (px2.south east) -- (r-18-4.south east);

\draw[dashed, ->, thick] (0,0.125) -| (0,2.75) -- (0.5,2.75);
\draw[dashed, ->, thick] (4.625,2.75) -- (0.75,2.75);

\end{tikzpicture} \label{fig:neon_impl_path_move_2} } %
\begin{tikzpicture}
\node[rectangle, fill=KITred,   opacity=0.5]    (rec1) at (0.3,0.25) {};
\node[rectangle, fill=KITgreen, opacity=0.2]    (rec2) at (0.3,0.75) {};
\node[rectangle, fill=KITcyanblue,opacity=0.5]  (rec3) at (0.3,1.25) {};
\node[rectangle, fill=KITorange, opacity=0.5]   (rec4) at (0.3,1.75) {};
\node[rectangle, fill=KITblack, opacity=0.5]    (rec5) at (0.3,2.25) {};

\node[anchor=west] (text1) at (rec1.east) {Current pixel};
\node[anchor=west] (text2) at (rec2.east) {Processed pixel};
\node[anchor=west] (text3) at (rec3.east) {Next pixel};
\node[anchor=west] (text4) at (rec4.east) {Next path};
\node[anchor=west] (text5) at (rec5.east) {Starting point};

\end{tikzpicture}
{
	\caption{Different traversal strategies in the aggregation of the \gls*{SGM} path costs} %
}
\end{figure}
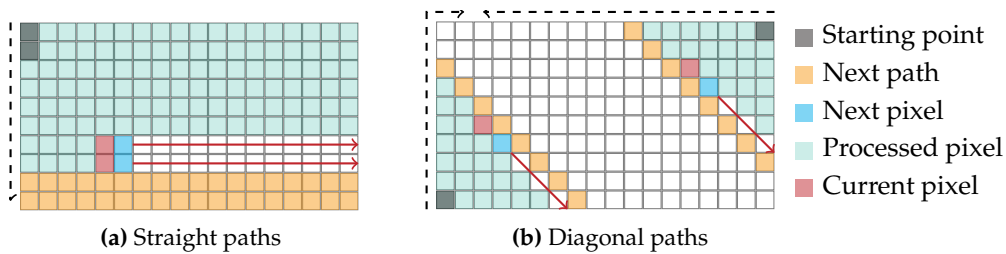
In each thread, the costs for the disparities of two pixels are calculated simultaneously. %
In this, the implementation of the horizontal and vertical paths ($L_\mathrm{LR}$, $L_\mathrm{RL}$, $L_\mathrm{TB}$, $L_\mathrm{BT}$) differ from the implementation of the diagonal paths ($L_\mathrm{TLBR}$, $L_\mathrm{TRBL}$, $L_\mathrm{BLTR}$, $L_\mathrm{BRTL}$). %
On the straight paths, each thread processes two pixels on two neighboring rows or columns (\Cref{fig:neon_impl_path_move_1}).  %
However, due to the different lengths of the diagonal paths, this cannot be applied to the processing of the same. %
Instead, on the diagonal paths, each thread processes two pixels lying on opposite sides of the image. %
This is illustrated by \Cref{fig:neon_impl_path_move_2}. %

In the second step of the \gls*{SGM} aggregation, each thread sums up all \gls*{SGM} path costs and finds the disparity with the minimum cost, \ie the \gls*{WTA} cost, for each image stripe assigned to it. %
The different path costs are first copied into different vector-registers and then summed up. %
While summing up, each thread additionally stores the currently processed disparity as well as the minimum aggregated cost and the corresponding \gls*{WTA} disparity in additional vector-registers. %
Afterwards, the aggregated costs are compared to the minimum costs which are updated if necessary. %
If the minimum costs are updated, the corresponding \gls*{WTA} disparity is updated accordingly. %
The final aggregated path costs are stored into an aggregated cost volume $\bar{\mathcal{S}}$, which is needed for the subsequent consistency check, while the final \gls*{WTA} disparity is written into the disparity map. %

\subsubsection{Consistency check}
\label{sec:methodology_cpu_constency}

In the optimization of the consistency check for the CPU, we encounter the same difficulty as in the implementation of the approximated consistency check for the GPU, namely that the required data from the aggregated cost volume does not lie physically next to each other (\cf \Cref{fig:impl_consistency_probl}).
This makes it inefficient to load the data into vector-registers first and then process it with NEON intrinsics. %
There are two ways to solve this problem: %
The first possibility, which is proposed by \citet{Spangenberg2014large}, is to first transform the aggregated cost volume $\bar{\mathcal{S}}$ into a temporary volume in such a way that the data which is required to compute the approximated disparity map $D^{\mathrm{R}}$ will physically lie next to each other in memory. %
Afterwards, the vector-registers can be filled with the data and the approximated disparity map $D^\mathrm{R}$, corresponding to the matching image, can be calculated efficiently with \gls*{SIMD} instructions. %
The second option, which we chose, is to refrain from using \gls*{SIMD} instructions for the consistency check. %
In our implementation of the consistency check, we only use a thread-level parallelization in which each thread is processing a different part of the cost volume. %
This saves us the need to rearrange the cost volume necessary to use \gls*{SIMD} instructions. %

\subsubsection{Median filter}
\label{sec:methodology_cpu_median}

\begin{figure}[b!] %
	\centering %
	\input{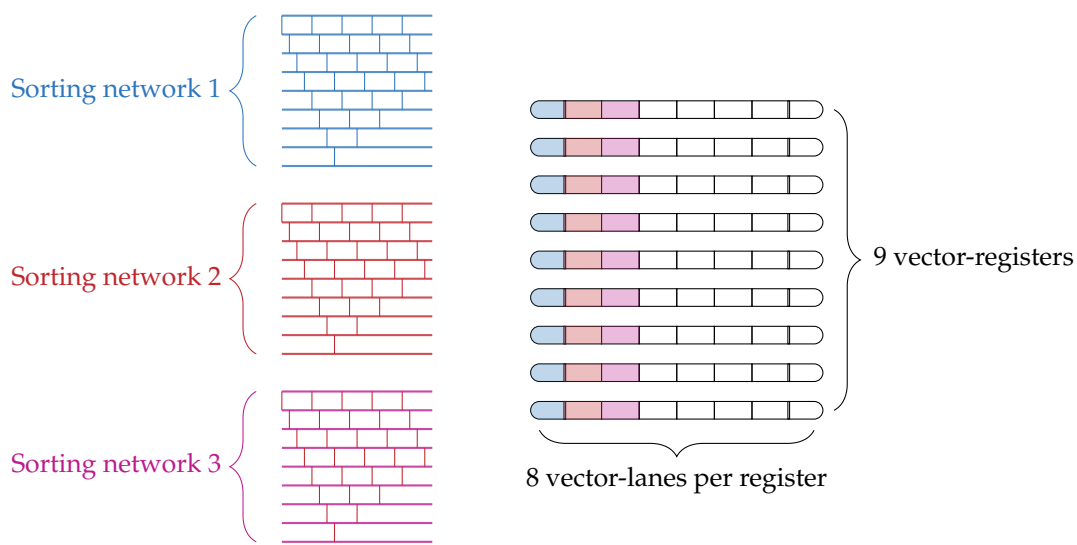} %
	\caption { %
		Implementation of sorting networks using vectorized \gls*{SIMD} processing. %
		The nine wires of a sorting network are mapped to vector-lanes of nine different vector-registers. %
	} %
	\label{fig:neon_impl_sorting_network} %
\end{figure} %

In our pipeline, we deploy a final median filter after the consistency check to remove any small outliers that still remain in the disparity map $D$. %
This requires a sorting of all disparity values within the local neighborhood. %
We utilize the concept of sorting networks (\cf \Cref{sec:appendix_sortingnetworks}), which relies on wires and comparators, to allow for a parallel sorting with \gls*{SIMD} intrinsics. %
In the implementation for the vectorized processing of the 3$\times$3 median filter, the wires of the sorting network are realized using the vector-lanes of the vector-register. %
In this, the wires of the network are distributed among different vector-registers, utilizing one vector-lane from each register. %
Thus, for the implementation of one sorting network for the median filter with nine wires, nine different vector-lanes distributed over nine different vector-registers are utilized (\Cref{fig:neon_impl_sorting_network}). %

The comparators of the sorting network are implemented by using two comparison instructions of the NEON instruction set, that compare each vector-lane of two vector-registers and store the minimum or maximum in a third register. %
Thus, for each vector-lane we first extract the minimum and maximum value from the two source vector-registers. %
Then, the minimum is copied to the register which represents the upper lane of the sorting network, while the maximum is copied to the register which represents the lower lane. %
By this vectorized parallelization the median filter is computed for 16 pixels simultaneously. %
Inherent to the nature of the BubbleSort algorithm, it is only necessary to calculate the first five iterations to get the median of a 3$\times$3\px neighborhood. %

\section{Results}
\label{sec:experiments}
\sloppy

The first part of our experiments (\Cref{sec:Experiments-Quantitative}) comprises a quantitative evaluation of the accuracy and performance of our optimized implementations on the \KITTI 2015 stereo benchmark \citep{Menze2015object}, as well as the Middlebury 2014 stereo benchmark \citep{Scharstein2014high}. %
In this, we have also evaluated and studied the effects of different configurations of the processing pipeline for real-time disparity estimation, \eg the effect of reducing the number of \gls*{SGM} paths or the improvement of the additional subpixel refinement. %
We compare the results of our implementations with state-of-the-art approaches and analyze their performance with respect to the power consumption of the embedded system. %

In the second part (\Cref{sec:Experiments-Qualitative}), we qualitatively discuss the results of the use-case specific experiments we have conducted. %
In this, we have deployed our approaches on a low-cost \gls*{UAV} and performed real-time disparity estimation based on a stereo camera system, which is pointing forward in the direction of flight. %
The resulting disparity map can then be used for reactive obstacle avoidance as proposed in our previous work \citep{Ruf2018realtime} or for facade and close-range object reconstruction.

\subsection{Quantitative evaluation of accuracy on public stereo benchmarks}
\label{sec:Experiments-Quantitative}

The quantitative assessment of the performance of our approaches comprises an evaluation with respect to their accuracy in \Cref{sec:Experiments-QuantitativeAccuracy}, as well as studies on the effects of the subpixel disparity refinement in \Cref{sec:Experiments-QuantitativeRefinement} and the improvements gained by an accurate left-right consistency check in \Cref{sec:Experiments-QuantitativeConsistency}. %
For the evaluation of the accuracy of our approach and its different configurations, we have used the training set of the \KITTI 2015 stereo benchmark \citep{Menze2015object}, which consists of 200 stereo image pairs and ground truth disparity maps captured by a \gls*{LIDAR} sensor from on top of a car driving around urban areas, as well as the Middlebury 2014 stereo benchmark \citep{Scharstein2014high}. %
The latter one allows a more thorough evaluation on the accuracy of our approach, and the effects of different optimizations in the processing pipeline, since it consists of 15 high-resolution stereo pairs of indoor scenes, together with highly accurate and dense ground truth disparity maps captured by a structured light sensor. %
Moreover, in \Cref{sec:Experiments-QuantitativeSpeed}, we study the processing speed and power consumption of our approaches, together with the effects of reducing the aggregation paths of the \gls*{SGM} optimization. %
For our experiments, we have deployed our approaches on the NVIDIA Jetson Xavier AGX with an 8-core 64\,bit ARMv8.2 CPU and a 512-core Volta GPU. %
All measurements with respect to accuracy, timings and power consumption were done on this hardware. %

The standard evaluation routine of the \KITTI 2015 stereo benchmark \citep{Menze2015object} states the accuracy as the amount of erroneous pixels (D1-all), averaged over all $m$ ground truth pixels in the evaluation set, for which the estimated disparity $d$ differs by 3 or more pixels with respect to the ground truth $\hat{d}$: %
\begin{equation} %
	\text{D1-all} = \frac{1}{m} \sum_{i = 1}^m \left[|d-\hat{d}| \geq 3 \right], %
	\label{eq:accuracyKitti}
\end{equation} %
with $[\cdot]$ being the Iverson bracket. %
Since the ground truth was generated from a \acrshort*{LIDAR} sensor, mounted at a slightly different position as the camera, for which the disparity map is estimated, the ground truth also provides disparity values in areas which are occluded in the second camera image and, in turn, usually only contains limited information in the estimated disparity map. %
Although the \KITTI benchmark also provides ground truth maps which only contain non-occluded (noc) areas, the standard evaluation protocol uses the occluded (occ) dataset, which we have also used for our evaluation in \Cref{tab:Kitti2015_accuracy}. %
Furthermore, the benchmark distinguished between the results of the actual estimated (Est) disparity maps and interpolated versions of them (All). %
The latter ones allow a comparison between disparity maps of different density, by applying a background interpolation to fill the pixels in the estimated disparity map for which no data is available. %
However, since our approach uses a left-right consistency check and a median filter to explicitly remove outliers and inconsistent areas, we are more interested in the results achieved by the actual estimate. %
Nonetheless, for comparison, we also provide the results achieved by the interpolated disparity maps, as well as the information on the density of the non-interpolated map, if available, which states the amount of pixels in the estimated disparity map which contain valid estimates. %

Similar to the evaluation routine of the \KITTI 2015 stereo benchmark, the Middlebury benchmark ranks the algorithms based on four different accuracy levels, namely the amount of pixels whose error is greater than $0.5$ (bad$0.5$), $1$ (bad$1$), $2$ (bad$2$) and $4$ (bad$4$) \px with respect to all $m$ ground truth pixels in the evaluation set: %
\begin{equation} %
	\text{bad}\theta = \frac{1}{m} \sum_{i = 1}^m \left[|d-\hat{d}| > \theta \right], %
	\label{eq:accuracyMiddlebury}
\end{equation} %
with $d$ and $\hat{d}$ again denoting the estimated and ground truth disparity respectively, and $[\cdot]$ being the Iverson bracket. %
The data is provided in full (F) image resolution with up to $3000\times2000$\px and a disparity range of up to $800$\px, as well as half (H) and quarter (Q) image resolution. %
The official evaluation is always performed on the full image resolution. %
Thus, if the results are generated on a dataset with a smaller resolution, the results are first being up-sampled before being evaluated. %

\subsubsection{Accuracy}
\label{sec:Experiments-QuantitativeAccuracy}
\begin{table}[t]
\centering
\resizebox{\textwidth}{!}{%
\begin{tabular}{|l|l|l|r|r|r|r|} \toprule
\multirow{2}{*}{Approach} & \multirow{2}{*}{Configuration} & \multirow{2}{*}{HW Device} & Resolution & \multicolumn{3}{c|}{Accuracy} \\ \cline{5-7}
& & & \scriptsize{(in \px)} & D1-all (Est.)  & D1-all (All) & Density \\ \midrule\midrule
\citet{Zhao2020fp} & $\text{\gls*{CT}}_{5\times5}$ - \gls*{SGM} &\gls*{FPGA} & $1242\times375$ & - & $11.8\,\%$ & - \\
\citet{Zhao2020fp} & $\text{\gls*{CT}}_{7\times7}$ - \gls*{SGM} &\gls*{FPGA} & $1242\times375$ & - & $9.5\,\%$ & - \\
\citet{Ruf2018realtime} & $\text{\gls*{CT}}_{5\times5}$ - \gls*{SGM} &\gls*{FPGA} & $640\times360$ & $4.6\,\%$ & $31.2\,\%$ & $46.4\,\%$ \\
\citet{Rahnama2018r3sgm} & $\text{\gls*{CT}}_{5\times5}$ - \acrshort*{MGM} &\gls*{FPGA} & $1242\times375$ & $6.7\,\%$ & $13.6\,\%$ & $81.0\,\%$ \\
\citet{Rahnama2018r3sgm} & $\text{\gls*{CT}}_{13\times13}$ - \acrshort*{MGM} &\gls*{FPGA} & $1242\times375$ & $4.8\,\%$ & $9.9\,\%$ & $85.0\,\%$ \\
\citet{Cui2019real}$\dagger$ & $\text{\gls*{NCC}}_{9\times9}$-nativ & GPU & $1242\times375$ & - & $16.6\,\%$ & - \\
\citet{Cui2019real}$\dagger$ & $\text{\gls*{NCC}}_{9\times9}$-optimized & GPU & $1242\times375$ & - & $13.1\,\%$ & - \\
\citet{Chang2020zigzag} & Z$^2$-ZNCC & GPU & $1242\times375$ & $7.6\,\%$ & $7.7\,\%$ & $99.9\,\%$ \\
\citet{Hernandez2016embedded} & $\text{CT}_{9\times7}$ - \gls*{SGM} & GPU & $1242\times375$ & $8.2\,\%$ & $8.2\,\%$ & $\mathbf{100\,\%}$ \\
\midrule
\RESSTAC - CUDA & $\text{\gls*{CT}}_{5\times5}$ - \gls*{SGM} & GPU & $640\times480$ & $5.4\,\%$ & $8.4\,\%$ & $94.5\,\%$ \\
\RESSTAC - CUDA & $\text{\gls*{CT}}_{5\times5}$ - \gls*{SGM} & GPU & $1242\times375$ & $4.3\,\%$ & $8.3\,\%$ & $88.8\,\%$  \\
\RESSTAC - CUDA & $\text{\gls*{CT}}_{9\times7}$ - \gls*{SGM} & GPU & $640\times480$ & $5.1\,\%$ & $7.9\,\%$ & $94.6\,\%$ \\
\RESSTAC - CUDA & $\text{\gls*{CT}}_{9\times7}$ - \gls*{SGM} & GPU & $1242\times375$ & $\mathbf{4.0\,\%}$ & $7.7\,\%$ & $90.0\,\%$ \\
\RESSTAC - CUDA & $\text{\gls*{NCC}}_{5\times5}$ - \gls*{SGM} & GPU & $640\times480$ & $5.3\,\%$ & $7.8\,\%$ & $94.8\,\%$ \\
\RESSTAC - CUDA & $\text{\gls*{NCC}}_{5\times5}$ - \gls*{SGM} & GPU & $1242\times375$ & $4.3\,\%$ & $8.1\,\%$ & $90.0\,\%$\\
\RESSTAC - CUDA & $\text{\gls*{NCC}}_{9\times9}$ - \gls*{SGM} & GPU & $640\times480$ & $5.9\,\%$ & $8.2\,\%$ & $94.7\,\%$\\
\RESSTAC - CUDA & $\text{\gls*{NCC}}_{9\times9}$ - \gls*{SGM} & GPU & $1242\times375$ & $4.8\,\%$ & $8.3\,\%$ & $91.1\,\%$ \\
\RESSTAC - NEON & $\text{\gls*{CT}}_{5\times5}$ - \gls*{SGM} & CPU & $640\times480$ & $5.0\,\%$ & $7.9\,\%$ & $94.5\,\%$\\
\RESSTAC - NEON & $\text{\gls*{CT}}_{5\times5}$ - \gls*{SGM} & CPU & $1242\times375$ & $4.6\,\%$ & $8.5\,\%$ & $90.0\,\%$\\
\midrule
\citet{Schoenberger2018sgm} & $\text{\gls*{NCC}}_{7\times7}$ - \gls*{SGM}-Forest & CPU & $1242\times375$ & $4.3\,\%$ & $\mathbf{4.4\,\%}$ & $99.9\,\%$ \\
OpenCV-SGBM & $\text{SAD}_{3\times3}$ - \gls*{SGM} & CPU & $1242\times375$ & $5.9\,\%$ & $10.9\,\%$ & $90.4\,\%$ \\
\citet{Hirschmueller2008} & \gls*{CT} - \gls*{SGM} & GPU & $1242\times375$ & $6.4\,\%$ & $6.4\,\%$ & $\mathbf{100\,\%}$ \\
\bottomrule
\end{tabular}}
\caption{
	Accuracy achieved by different algorithms and configurations on the \KITTI 2015 stereo benchmark \citep{Menze2015object}. %
	While the upper part lists algorithms that are optimized and deployed on embedded hardware, the three algorithms at the bottom are listed as a reference and a baseline. %
	The results achieved by different configurations of our implementation are listed in the middle section. %
	The accuracy is stated as the amount of erroneous pixels (D1-all), for which the estimated disparity differs by 3 or more pixels with respect to the ground truth. %
	The \KITTI 2015 benchmark distinguishes between the result of the actual estimated (Est) disparity map and an interpolated version of it (All), in which the pixels, for which no disparity is available, get interpolated by a simple background interpolation. %
	As an evaluation ground truth, we have considered all available pixels, not only the non-occluded ones. %
	The density indicates, how many pixels inside the computed disparity maps have an estimate. %
	$\dagger$: The accuracy stated is computed with respect to the non-occluded pixels in the ground truth.
	} %
\label{tab:Kitti2015_accuracy}
\end{table}

\begin{figure}[t!]
     \centering
     \subfloat{\includegraphics[width=0.495\textwidth]{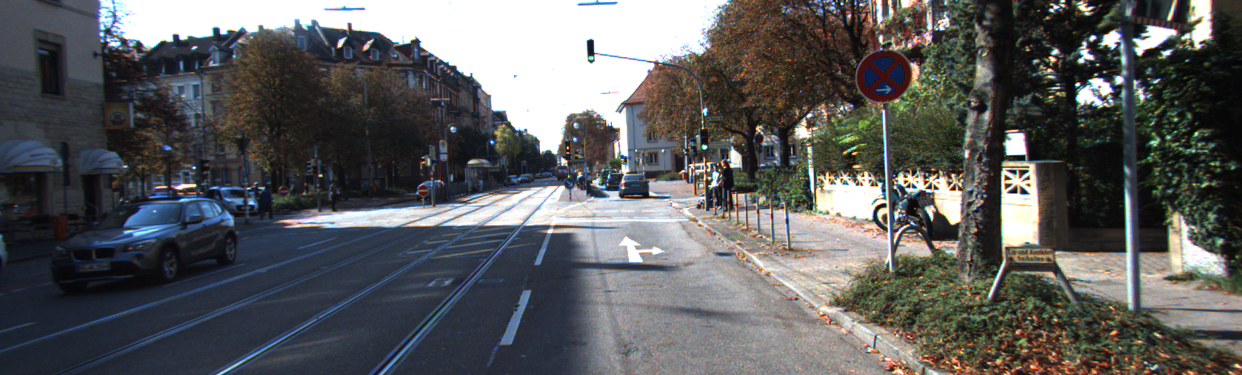}} \hfill
	 \subfloat{\includegraphics[width=0.495\textwidth]{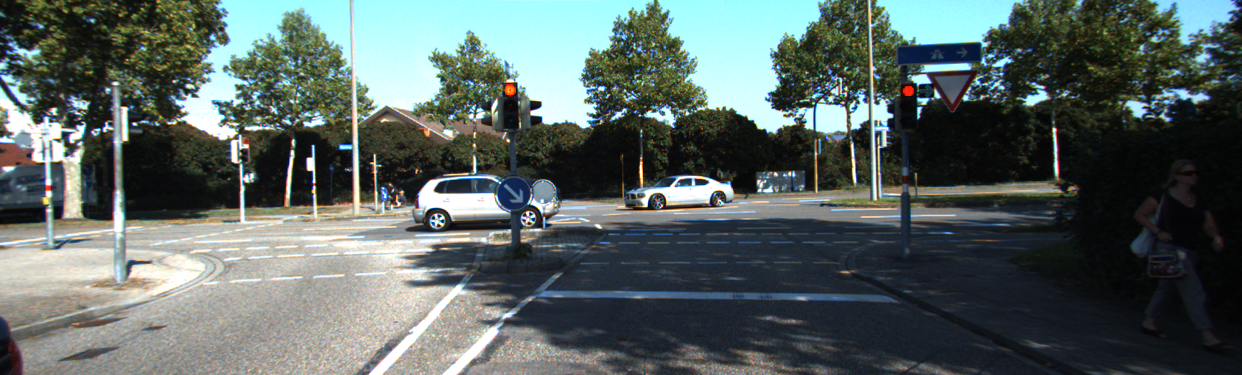}} \\[-1.5ex]
     \subfloat{\includegraphics[width=0.495\textwidth]{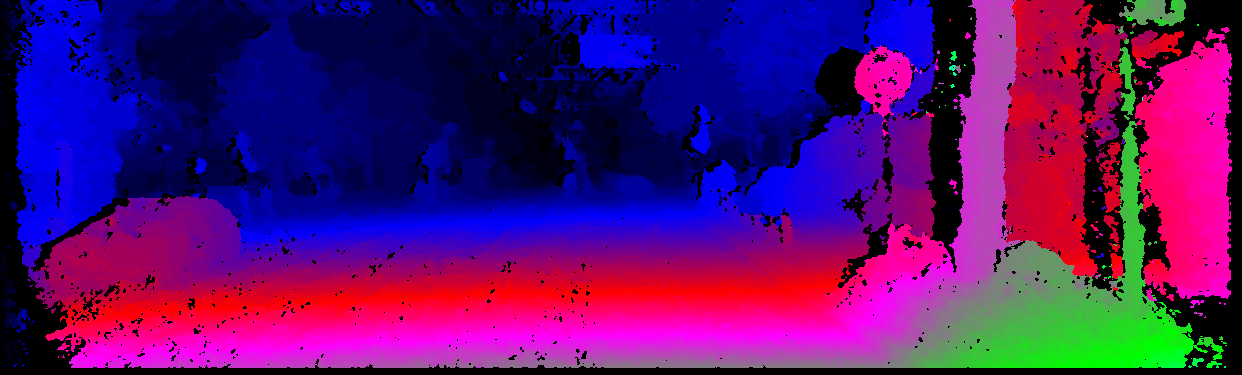}} \hfill
	 \subfloat{\includegraphics[width=0.495\textwidth]{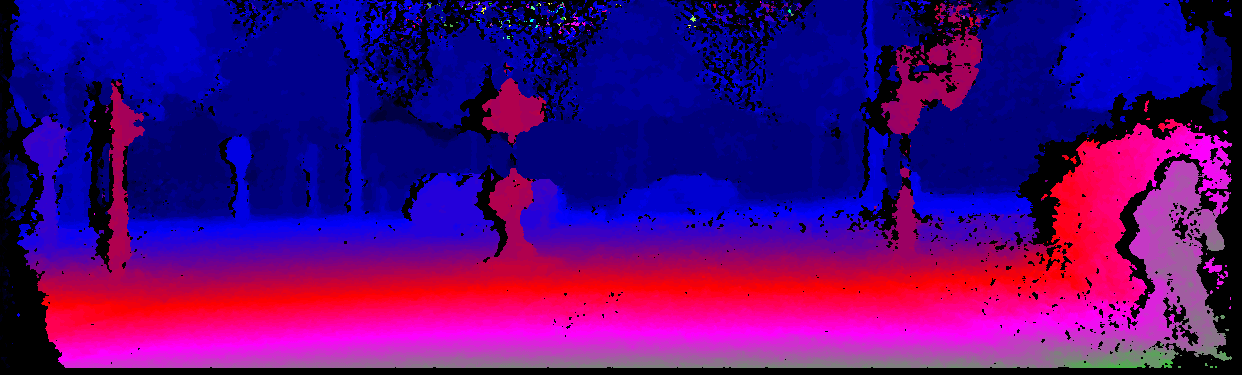}} \\[-1.5ex]
     \subfloat{\includegraphics[width=0.495\textwidth]{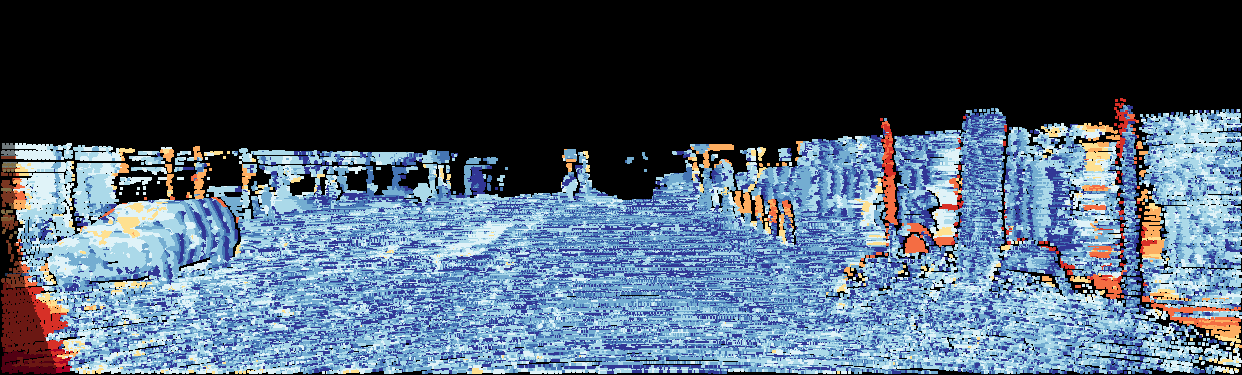}} \hfill
	 \subfloat{\includegraphics[width=0.495\textwidth]{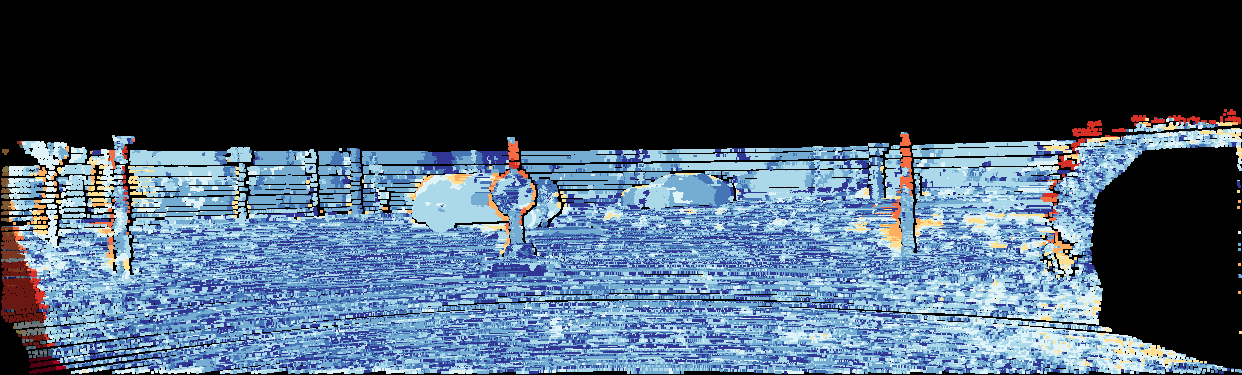}} \\[-1.5ex]
     \subfloat{\includegraphics[width=0.495\textwidth]{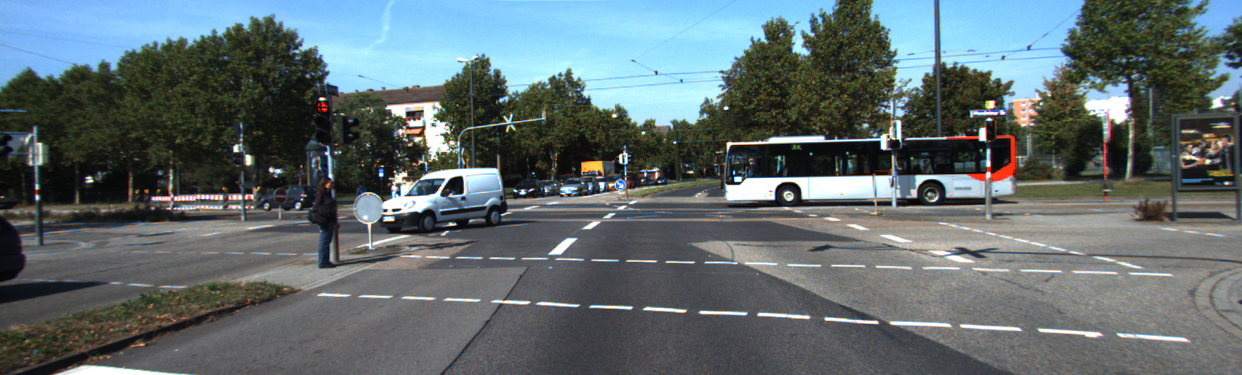}} \hfill
     \subfloat{\includegraphics[width=0.495\textwidth]{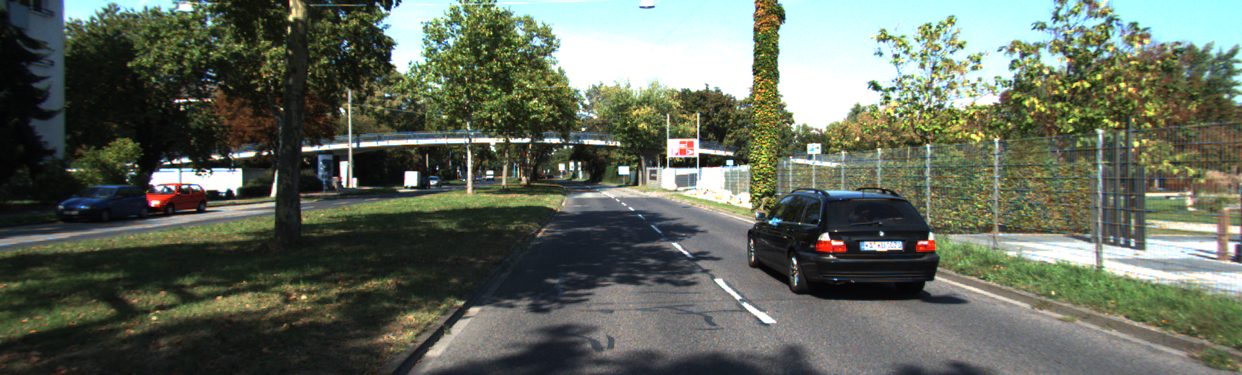}} \\[-1.5ex]
     \subfloat{\includegraphics[width=0.495\textwidth]{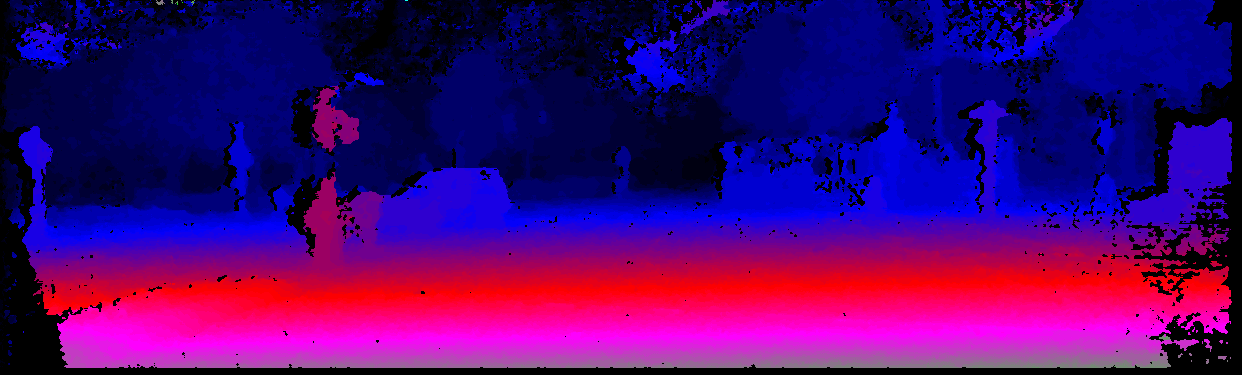}} \hfill
     \subfloat{\includegraphics[width=0.495\textwidth]{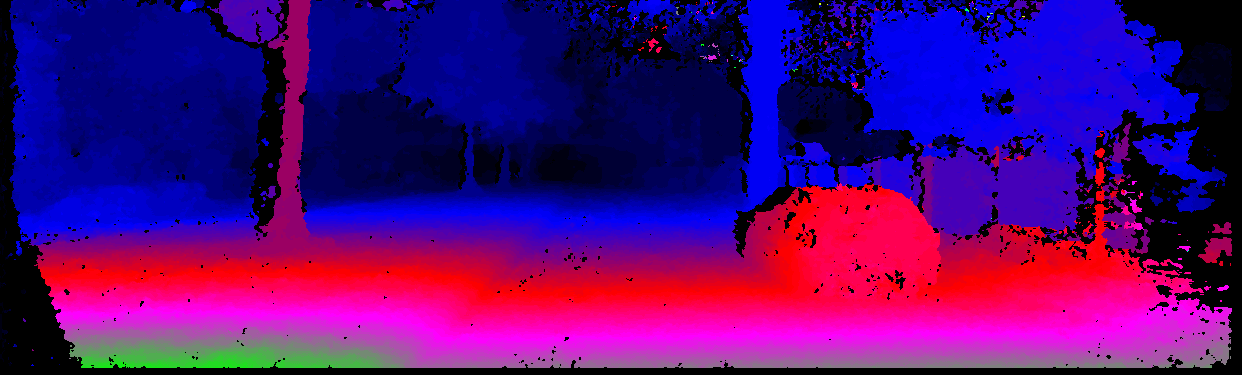}} \\[-1.5ex]
     \subfloat{\includegraphics[width=0.495\textwidth]{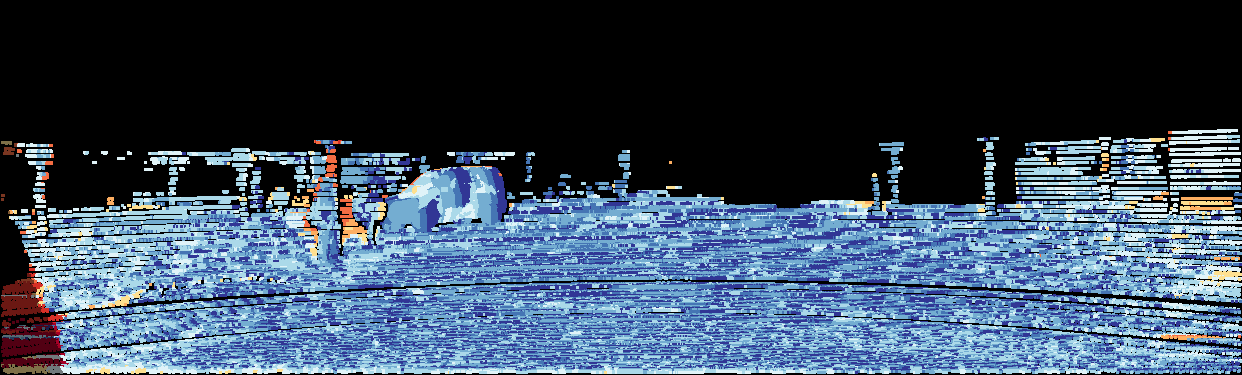}} \hfill
	 \subfloat{\includegraphics[width=0.495\textwidth]{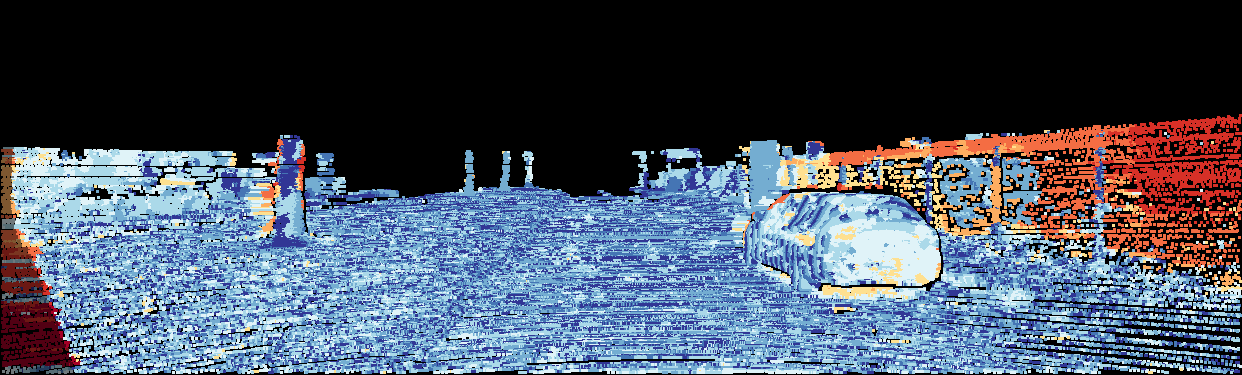}} \\
     \caption{
     	Four exemplary results from the \KITTI 2015 stereo benchmark, computed with the  $\text{\gls*{CT}}_{9\times7}$ - \gls*{SGM} configuration on the original image size. %
     	\textbf{Rows 1 \& 4}: Reference image. %
     	\textbf{Rows 2 \& 5}: Estimated disparity maps. %
     	\textbf{Rows 3 \& 6}: Color-coded error image between the prediction and the ground truth. %
     	Error images use a log-color scale as described by \citet{Menze2015object}, marking correct pixels in estimates and wrong estimates in red color tones.
     }
     \label{fig:Kitti2015_result}
\end{figure}

In this section, we will first list the accuracy achieved on the \KITTI 2015 stereo benchmark, which is followed by the evaluation of the accuracy on the 
Middelbury 2014 stereo benchmark. %

\paragraph{\KITTI 2015 stereo benchmark:}

\Cref{tab:Kitti2015_accuracy} lists the quantitative results of the accuracy of different configurations of our implementations, as well as those of other approaches and implementations, which are achieved on the \KITTI 2015 stereo benchmark. %
While the results of our approach were achieved on the training set of the benchmark, the results of the approaches from literature were taken either from the corresponding publication or the official listing of the benchmark, which lists the results achieved on the actual test set. %
The upper part of the \Cref{tab:Kitti2015_accuracy} lists algorithms and configurations, which are optimized for the deployment and execution on embedded hardware. %
Not all of these are variants of the \acrlong*{SGM} stereo algorithm. %
Yet, they serve as a good comparison since they were deployed on the same or similar hardware as ours. %
The three algorithms at the bottom of the list serve as a baseline to our approaches. %
While the one from \citet{Hirschmueller2008} reveals the accuracy achieved by the original \gls*{SGM} algorithm implemented on the GPU with the \acrlong*{CT} of unknown size as a cost function, the SGBM variant of the OpenCV library is widely spread and easy to use, however, not optimized for embedded processing. %
The algorithm of \citet{Schoenberger2018sgm} is evaluated on both the \KITTI 2015 stereo benchmark, as well as the Middlebury 2014 stereo benchmark. %
They propose to use a random forest classifier to learn to efficiently fuse the different scan-line optimizations of the \gls*{SGM} algorithm, in order to reduce the number of optimization paths for embedded processing more efficiently. %

We have measured the accuracy of our approach using different cost functions and different support regions. %
For each configuration, we have evaluated the results achieved on the original image size provided by the \KITTI benchmark, \ie $1242\times375$\px, as well as on images with VGA resolution. %
In the case of VGA resolution, we have down-sampled the original images to a resolution of $640\times480$\px, performed the stereo disparity estimation and up-sampled the resulting disparity maps to the original image size with a nearest-neighbor interpolation and a scaling of the disparities by the horizontal scale factor. %
Thus, we have always used the original image size for the accuracy measurements. %
For the selection of the \gls*{SGM} penalties, we have empirically evaluated different values and selected those with the best result.  %
In terms of the best configurations, these are $P_1 = 27\ \text{and}\ P_2 = 86$ for the $\text{\gls*{CT}}_{9\times7}$ and $P_1 = 90\ \text{and}\ P_2 = 880$ for the $\text{\gls*{NCC}}_{5\times5}$. %
An excerpt on the qualitative results for the best configuration of our approach is presented in \Cref{fig:Kitti2015_result}. %

The results in \Cref{tab:Kitti2015_accuracy} reveal, that the use of the \acrlong*{CT} with a support region of $9\times 7$\px and its Hamming distance as a matching cost function, achieves the best results, in the evaluation of both the actual estimate and the interpolated version. %
As can be expected, the use of down-sampled versions of the input images yield less accurate results and yet, achieves a higher density in the resulting disparity maps. %
Furthermore, the use of the \acrlong*{NCC} as a cost function achieves slightly less accurate results, while leading to a smaller throughput as discussed in \Cref{sec:Experiments-QuantitativeSpeed}. %
The implementation on the CPU using NEON \gls*{SIMD} intrinsics with a \gls*{CT} of size $5\times 5$\px achieves similar and, in case of the smaller image resolution, slightly better results than its GPU counterpart.
In summary, when considering the actual estimate, our configuration $\text{\gls*{CT}}_{9\times7}$ - \gls*{SGM} executed on the GPU, with the original image size of $1242\times375$\px, outperforms the baseline implementations as well as the other algorithms optimized for embedded hardware. %
When evaluating on the interpolated disparity maps, our approaches achieve similar and mostly better results than the other embedded algorithms. %
The superiority of the baseline algorithms from \citet{Hirschmueller2008} and \citet{Schoenberger2018sgm} with respect to the quality of the disparity estimation is to be expected, since they were optimized with respect to accuracy and not speed or throughput. %

\paragraph{Middelbury 2014 stereo benchmark:}

\begin{table}[t!]
\centering
\resizebox{\textwidth}{!}{%
\begin{tabular}{|l|l|r|r|r|r|r|r|} \toprule
\multirow{2}{*}{Approach} & \multirow{2}{*}{Configuration} & \multirow{2}{*}{Resolution} & \multicolumn{5}{c|}{Accuracy} \\ \cline{4-8}
& & & bad$0.5$  & bad$1$ & bad$2$ & bad$4$ & Density\\ \midrule\midrule
\RESSTAC - CUDA & $\text{\gls*{CT}}_{5\times5}$ - \gls*{SGM} & Orig. Q 			& $77.0\,\%$ & $59.5\,\%$ & $35.4\,\%$ & $13.5\,\%$ & $93.0\,\%$ \\
\RESSTAC - CUDA & $\text{\gls*{CT}}_{9\times7}$ - \gls*{SGM} & Orig. Q 			& $77.3\,\%$ & $59.9\,\%$ & $35.7\,\%$ & $13.6\,\%$ & $93.3\,\%$ \\
\RESSTAC - CUDA & $\text{\gls*{NCC}}_{5\times5}$ - \gls*{SGM} & Orig. Q 			& $77.1\,\%$ & $60.1\,\%$ & $36.3\,\%$ & $14.4\,\%$ & $92.8\,\%$ \\
\RESSTAC - CUDA & $\text{\gls*{NCC}}_{9\times9}$ - \gls*{SGM} & Orig. Q 			& $77.1\,\%$ & $61.2\,\%$ & $38.7\,\%$ & $17.1\,\%$ & $91.9\,\%$ \\
\RESSTAC - NEON & $\text{\gls*{CT}}_{5\times5}$ - \gls*{SGM} & Orig. Q 			& $76.2\,\%$ & $59.0\,\%$ & $35.1\,\%$ & $13.4\,\%$ & $92.1\,\%$ \\
\midrule
\citet{Schoenberger2018sgm} & $\text{\gls*{NCC}}_{7\times7}$ - \gls*{SGM}-Forest & Orig. H 	& $\mathbf{43.1\,\%}$ & $\mathbf{14.8\,\%}$ & $\mathbf{7.0\,\%}$ & $\mathbf{3.7\,\%}$ & $100\,\%$ \\
OpenCV-SGBM & $\text{SAD}_{3\times3}$ - \gls*{SGM} & Orig. Q 								& $67.3\,\%$ & $42.1\,\%$ & $25.5\,\%$ & $17.3\,\%$ & $100\,\%$ \\
\citet{Hirschmueller2008} & \gls*{CT} - \gls*{SGM} & Orig. H 								& $51.5\,\%$ & $28.2\,\%$ & $17.7\,\%$ & $12.2\,\%$ & $100\,\%$ \\
\bottomrule
\end{tabular}}
\caption{
	Accuracy achieved by different algorithms and configurations on the Middlebury 2014 stereo benchmark \citep{Scharstein2014high}. %
	While the upper part lists the results of different configurations of our implementation, the three algorithms at the bottom are listed as a reference and a baseline. %
	The accuracy is stated as the amount of erroneous pixels, whose error is greater than $0.5$ (bad$0.5$), $1$ (bad$1$), $2$ (bad$2$) and $4$ (bad$4$) \px with respect to the ground truth. %
	The density indicates, how many pixels inside the computed disparity maps have an estimate. %
	The Middlebury 2014 stereo benchmark provides the image data in full (F), half (H) and quarter (Q) image resolution. %
	We have computed our results on the quarter image resolution and evaluated them according to the standard evaluation pipeline on the full resolution. %
} %
\label{tab:Middlebury2014_accuracy}
\end{table}

\begin{figure}[b!]
     \centering
     \subfloat{\includegraphics[width=0.33\textwidth]{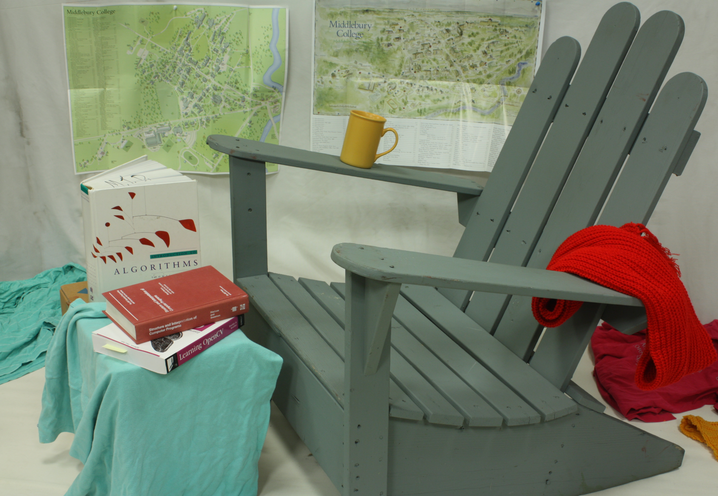}} \hfill
     \subfloat{\includegraphics[width=0.33\textwidth]{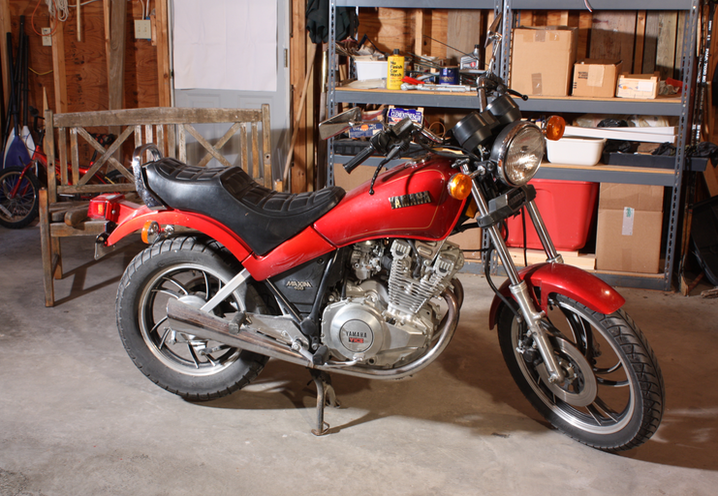}} \hfill
     \subfloat{\includegraphics[width=0.33\textwidth]{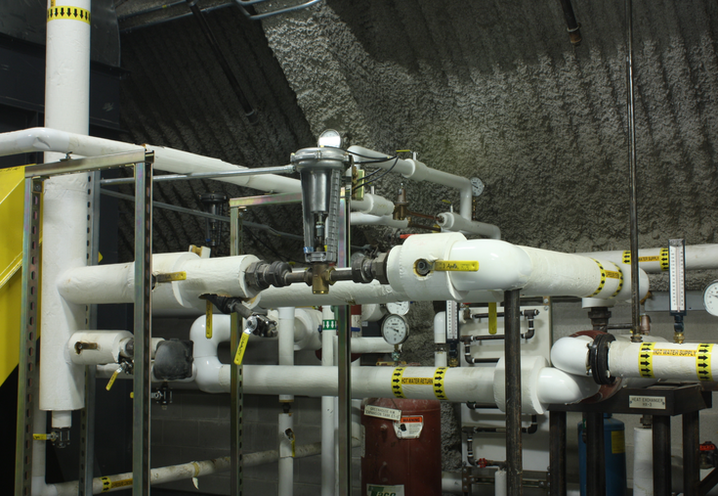}} \\[-1.5ex]
     \subfloat{\includegraphics[width=0.33\textwidth]{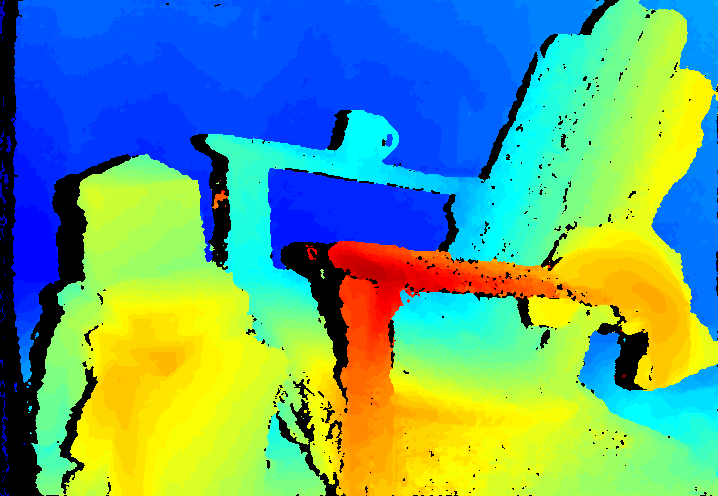}} \hfill
     \subfloat{\includegraphics[width=0.33\textwidth]{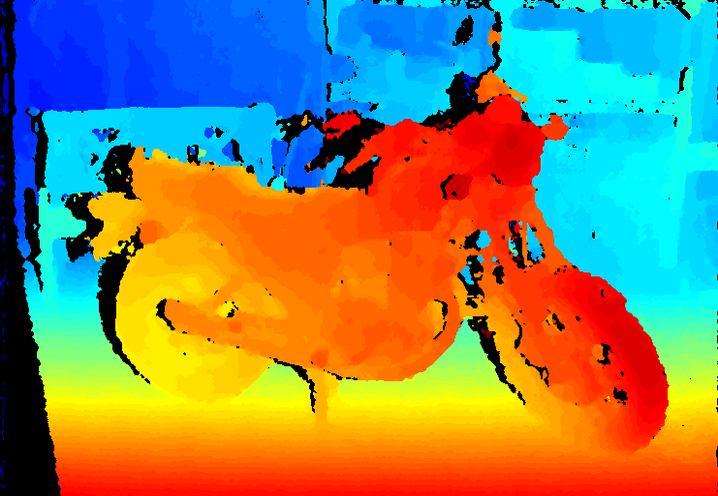}} \hfill
     \subfloat{\includegraphics[width=0.33\textwidth]{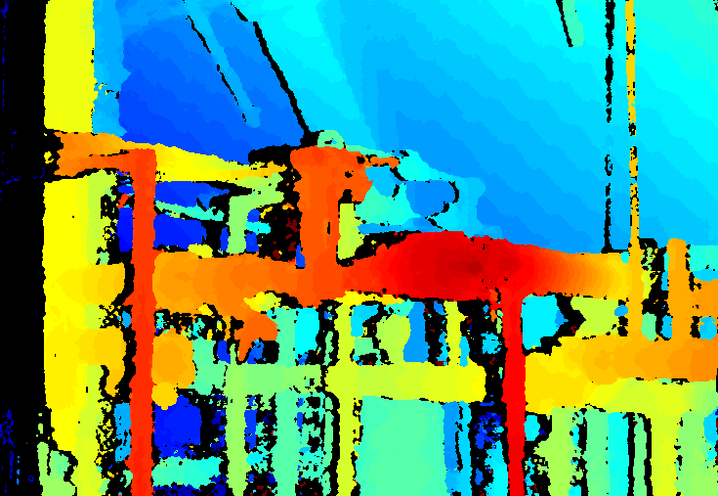}} \\[-1.5ex]
	 \subfloat{\includegraphics[width=0.33\textwidth]{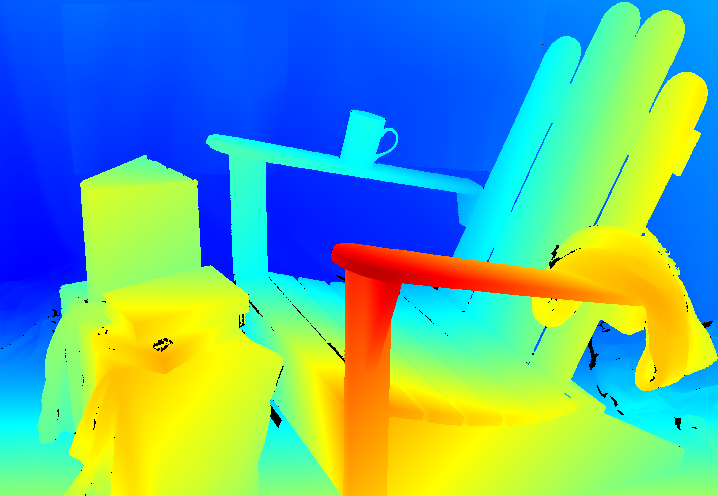}} \hfill
     \subfloat{\includegraphics[width=0.33\textwidth]{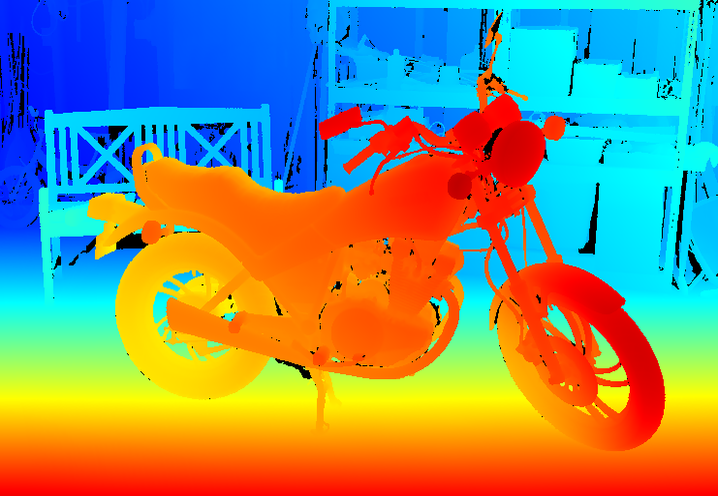}} \hfill
     \subfloat{\includegraphics[width=0.33\textwidth]{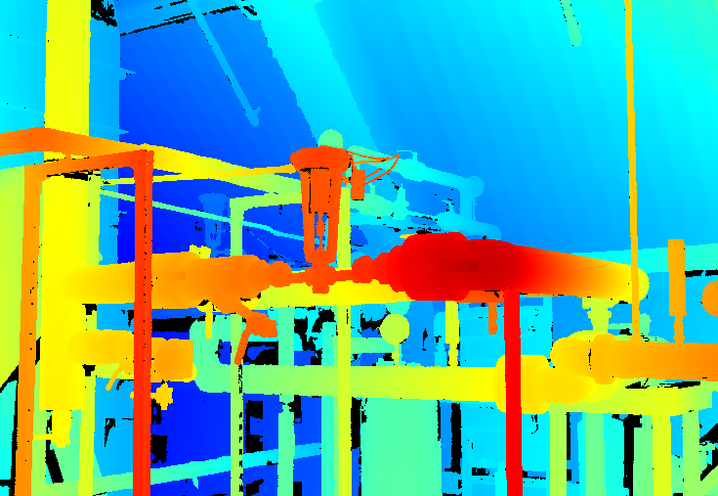}} \\
     \caption{
     	Three exemplary results from the Middlebury 2014 stereo benchmark, computed with the best performing base configuration, \ie $\text{\gls*{CT}}_{5\times5}$ - \gls*{SGM} on the quarter image resolution. %
     	\textbf{Row 1}: Reference image.
     	\textbf{Row 2}: Estimated disparity map.
     	\textbf{Row 3}: Ground truth disparity map.
     }
     \label{fig:Middlebury2014_result}
\end{figure}

The results achieved by our approach on the training set of the benchmark are listed in \Cref{tab:Middlebury2014_accuracy}, with a qualitative excerpt of the results achieved by the best configuration presented in \Cref{fig:Middlebury2014_result}. %
None of the other approaches for stereo processing on embedded hardware, which were listed in the evaluation on the \KITTI benchmark, have also been evaluated on the Middlebury 2014 stereo benchmark and, thus, these are not listed in this evaluation. %
However, results on the non-embedded baseline algorithms are available and are again listed in the lower part of \Cref{tab:Middlebury2014_accuracy}. %
We have evaluated the same configurations as in the evaluation on the \KITTI benchmark. %
Since our approach can only handle a disparity range of up to 256\px, we computed the disparity maps on the provided quarter image resolution (Orig. Q). %
Just as in the standard evaluation routine of the benchmark, the results listed were found after up-sampling the disparity maps to the original image resolution with a nearest-neighbor interpolation and scaling the containing disparities with a factor of 4. %
Again, for the selection of the \gls*{SGM} penalties, we have empirically evaluated different values and selected those which yield the best results, being $P_1 = 11\ \text{and}\ P_2 = 39$ for the $\text{\gls*{CT}}_{5\times5}$ and $P_1 = 140\ \text{and}\ P_2 = 730$ for the $\text{\gls*{NCC}}_{5\times5}$. %

\Cref{tab:Middlebury2014_accuracy} does not yield any satisfying results. %
This, however, is not surprising, since we have only used quarter of the image resolution to compute the results and up-sampled them by a factor of 4 for evaluation, introducing a lot of errors due to interpolation. %
As stated by \citet{Scharstein2014high}, the aim of this benchmark is on providing new challenges for modern stereo algorithms in terms of image resolution, accuracy and scene complexity, and not necessarily on the evaluation of optimizations with respect to computational efficiency and run-time.
Nonetheless, the accuracy in the ground truth and the evaluation protocol of the Middlebury 2014 stereo benchmark allows for an evaluation of the improvement gained by a subpixel disparity refinement as done in the following section.

\subsubsection{The effect of subpixel disparity refinement}
\label{sec:Experiments-QuantitativeRefinement}

As described in \Cref{sec:methodology-post-processing}, a subpixel disparity refinement can be computed for each pixel in the disparity map, by fitting a parabola through the matching costs of the winning disparity and its two neighbors. %
This achieves an increase in accuracy of up to $0.8$\percent in case of the \KITTI benchmark (\Cref{tab:Kitti2015_disparityRefinement}), and up to $9$\percent in case of the Middlebury benchmark (\Cref{tab:Middlebury2014_disparityRefinement}), and yet only requires a small computational overhead (\cf \Cref{tab:throughput}). %
\begin{table}[b!]
\centering
\resizebox{0.75\textwidth}{!}{%
\begin{tabular}{|l|l|r|r|r|} \toprule
\multirow{2}{*}{Approach} & \multirow{2}{*}{Configuration} & Resolution & \multicolumn{2}{c|}{Accuracy} \\ \cline{4-5}
& & \scriptsize{(in \px)} & D1-all (Est.)  & D1-all (All) \\ \midrule\midrule
\RESSTAC - CUDA & $\text{\gls*{CT}}_{5\times5}$ - \gls*{SGM} - fine & $1242\times375$ & $3.5\,\%$\,\scriptsize{($-0.8$)} & $8.0\,\%$\,\scriptsize{($-0.4$)} \\
\RESSTAC - CUDA & $\text{\gls*{CT}}_{9\times7}$ - \gls*{SGM} - fine & $1242\times375$ & $3.3\,\%$\,\scriptsize{($-0.7$)} & $7.4\,\%$\,\scriptsize{($-0.3$)} \\
\RESSTAC - CUDA & $\text{\gls*{NCC}}_{5\times5}$ - \gls*{SGM} - fine & $1242\times375$ & $3.7\,\%$\,\scriptsize{($-0.6$)} & $7.7\,\%$\,\scriptsize{($-0.4$)} \\
\RESSTAC - CUDA & $\text{\gls*{NCC}}_{9\times9}$ - \gls*{SGM} - fine & $1242\times375$ & $4.3\,\%$\,\scriptsize{($-0.5$)} & $7.9\,\%$\,\scriptsize{($-0.4$)}  \\
\bottomrule
\end{tabular}}
\caption{
	Accuracy achieved by selected configurations of our approach with an additional subpixel disparity refinement on the training set of the \KITTI 2015 stereo benchmark \citep{Menze2015object}. %
	The corresponding differences to the accuracies listed in \Cref{tab:Kitti2015_accuracy} are given in parentheses. %
} %
\label{tab:Kitti2015_disparityRefinement}
\end{table}

\begin{table}[b!]
\centering
\resizebox{\textwidth}{!}{%
\begin{tabular}{|l|l|r|r|r|r|r|} \toprule
\multirow{2}{*}{Approach} & \multirow{2}{*}{Configuration} & \multirow{2}{*}{Resolution} & \multicolumn{4}{c|}{Accuracy} \\ \cline{4-7}
& & & bad $0.5$  & bad $1$ & bad $2$ & bad $4$ \\ \midrule\midrule
\RESSTAC - CUDA & $\text{\gls*{CT}}_{5\times5}$ - \gls*{SGM} - fine & Orig. Q 			& $72.4\,\%$\,\scriptsize{($-4.6$)} & $52.1\,\%$\,\scriptsize{($-7.4$)} & $26.1\,\%$\,\scriptsize{($-9.3$)} & $10.4\,\%$\,\scriptsize{($-3.1$)} \\
\RESSTAC - CUDA & $\text{\gls*{CT}}_{9\times7}$ - \gls*{SGM} - fine & Orig. Q 			& $73.0\,\%$\,\scriptsize{($-4.3$)} & $52.7\,\%$\,\scriptsize{($-7.2$)}  & $26.6\,\%$\,\scriptsize{($-9.1$)} & $10.7\,\%$\,\scriptsize{($-2.9$)} \\
\RESSTAC - CUDA & $\text{\gls*{NCC}}_{5\times5}$ - \gls*{SGM} - fine & Orig. Q 			& $73.8\,\%$\,\scriptsize{($-3.3$)} & $53.8\,\%$\,\scriptsize{($-6.3$)}  & $27.8\,\%$\,\scriptsize{($-8.5$)} & $12.1\,\%$\,\scriptsize{($-2.0$)} \\
\RESSTAC - CUDA & $\text{\gls*{NCC}}_{9\times9}$ - \gls*{SGM} - fine & Orig. Q 			& $73.8\,\%$\,\scriptsize{($-3.3$)} & $55.2\,\%$\,\scriptsize{($-6.0$)}  & $30.6\,\%$\,\scriptsize{($-8.1$)} & $14.8\,\%$\,\scriptsize{($-2.3$)} \\
\bottomrule
\end{tabular}
}
\caption{
	Accuracy achieved by selected configurations of our approach with an additional subpixel disparity refinement on the training set of the Middlebury 2014 stereo benchmark \citep{Scharstein2014high}. %
	The corresponding differences to the accuracies listed in \Cref{tab:Middlebury2014_accuracy} are given in parentheses. %
} %
\label{tab:Middlebury2014_disparityRefinement}
\end{table}

\subsubsection{Accurate left-right consistency check}
\label{sec:Experiments-QuantitativeConsistency}

To evaluate the effects of only approximating the disparity map corresponding to the right image of the stereo pair, which is needed for the left-right consistency check (\cf \Cref{sec:methodology-post-processing}), we have implemented a more exact and more costly consistency check for the GPU. %
In this, we switch and flip the reference and matching image and calculate a second disparity map for the original matching image. %
This leads to a more accurate disparity map $D^{\mathrm{R}}$ for the right input image, which, in turn, is used by \Cref{eq:LR_check} of the consistency check. %
With a more accurate $D^{\mathrm{R}}$ it is assumed that the consistency check is more effective in filtering outliers, but does the high computational overhead of fully calculating two disparity maps justify the increase in accuracy? %
The results of our evaluation do not indicate a significant improvement. %
Our studies reveal an increase in accuracy of only $0.4-1.0$\percent, when calculating the disparity map of the right image from scratch compared to only approximating it from the cost volume corresponding to the left disparity map. %
However, the throughput is nearly halved when using a more accurate consistency check, as illustrated by configurations with the prefix \textit{exact-cc} (exact consistency check) in  \cref{tab:throughput}.


\subsubsection{Throughput, frame rates and power consumption}
\label{sec:Experiments-QuantitativeSpeed}

\begin{figure}[b!] %
	\resizebox{\columnwidth}{!}{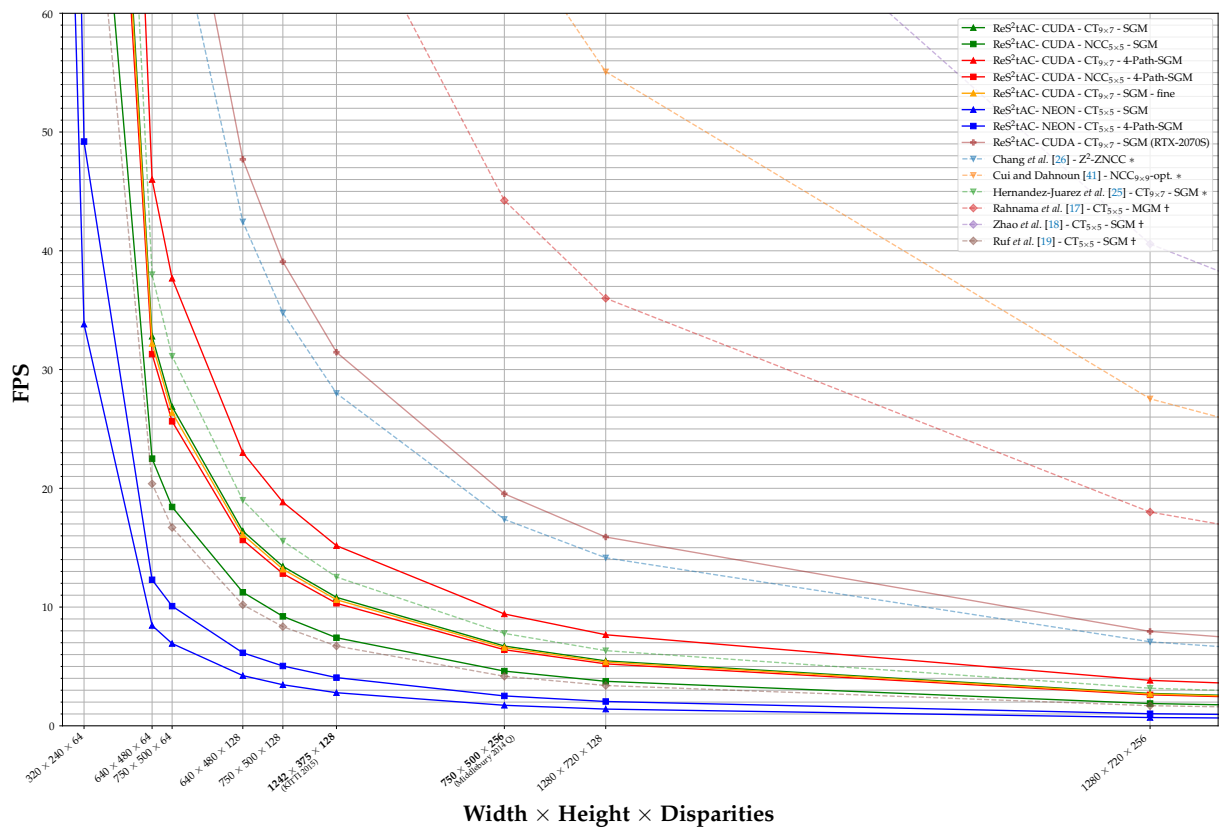}%
	\caption{\glsreset{FPS} %
		Expected \gls*{FPS} for sets of different image sizes and disparity ranges, based on the throughput achieved by different configurations and approaches listed in \Cref{tab:throughput} and \Cref{tab:throughput_4path}. %
		Image resolution and disparity ranges corresponding to the \KITTI 2015 and Middlebury 2014 Q benchmark are printed in bold. %
		$\ast$: Approaches from literature deployed on embedded \gls*{GPU} hardware. %
		$\dagger$: Approaches from literature deployed on embedded \gls*{FPGA} hardware. %
	}
	\label{fig:fps_curves}
\end{figure}

A typical measure to quantify the processing speed of a stereo algorithm is the number of \gls*{FPS} which can be calculated. %
However, since the \gls*{FPS} greatly depends on the image size of the output and the disparity range, we instead reason on the efficiency of our approaches based on their throughput, which is measured in \gls*{MDEs}: %
\begin{equation} %
\label{eq:mdes} %
\text{\MDEs} = \frac{W \cdot H \cdot |\Gamma|}{\text{run-time}},
\end{equation} %
with $W$ and $H$ being the width and the height of the resulting disparity map, and $|\Gamma|$ being the size of the disparity range. %
Given the throughput achieved by a certain configuration or algorithm, it is possible to deduce the expected frame rates for a set of image size and disparity range:
\begin{equation} %
\label{eq:fps} %
\text{\fps} = \frac{\text{\MDEs}}{W \cdot H \cdot |\Gamma|},
\end{equation} %
as done in \Cref{fig:fps_curves}. %
Furthermore, the throughput allows for a better comparison between different algorithms with respect to their processing speed, due to its independence of a fixed image size and disparity range. %

In \Cref{tab:throughput}, we have listed the throughput achieved by different configurations of our approach, as well as the throughput of selected embedded algorithms from literature. %
While all measurements of our approach were done on the NVIDIA Jetson Xavier AGX with maximum performance, the related work optimized for the execution on GPU were deployed on the NVIDIA Jetson TX1 and TX2. %
In case of the related work, which has not explicitly stated the throughput of the respective algorithm, we have used the frame rate and the corresponding image size to calculate the values according to \Cref{eq:fps}. %
For the measurement of the run-time of our approaches, we have considered the whole pipeline, including the upload of the stereo image pair and the download of the disparity image to and from the device memory of the \gls*{GPU}. %

Firstly, the measurements reveal the higher computational efficiency of the \acrlong*{CT} as a matching cost function with respect to the \acrlong*{NCC}. %
And secondly, they show the small computational overhead of the subpixel disparity refinement (fine), as discussed in \Cref{sec:Experiments-QuantitativeRefinement}. %
However, the measurements also unveil that our optimizations are less efficient than those that can be found in the literature. %
As expected, the implementations which are optimized and deployed on \gls*{FPGA} architectures are superior to those running on an embedded GPU. %
Furthermore, the superiority in terms of throughput of algorithms, such as those from \citet{Cui2019real} and \citet{Chang2020zigzag}, that do not rely on a complex regularization scheme like the \gls*{SGM} is also to be expected. %
Nonetheless, our approach has a lower throughput than a similar implementation of \citet{Hernandez2016embedded}, while simultaneously being deployed on a more powerful system. %

\begin{table}[t!]
\centering
\begin{tabular}{|l|l|l|r|} \toprule
\multirow{2}{*}{Approach} & \multirow{2}{*}{Configuration} & \multirow{2}{*}{HW Device} & Throughput \\
& & & \scriptsize{(in \MDEs)} \\
\toprule
\citet{Zhao2020fp} & $\text{\gls*{CT}}_{5\times5}$ - \gls*{SGM} &\gls*{FPGA} & $9589.9$ \\
\citet{Zhao2020fp} & $\text{\gls*{CT}}_{7\times7}$ - \gls*{SGM} &\gls*{FPGA} & $8743.7$ \\
\citet{Ruf2018realtime} & $\text{\gls*{CT}}_{5\times5}$ - \gls*{SGM} &\gls*{FPGA} & $400.8$ \\
\citet{Rahnama2018r3sgm} & $\text{\gls*{CT}}_{5\times5}$ - \acrshort*{MGM} & \gls*{FPGA} & $4246.9$ \\
\citet{Cui2019real} & $\text{\gls*{NCC}}_{9\times9}$-optimized & GPU (TX2) & $6497.1$ \\
\citet{Chang2020zigzag} & Z$^2$-ZNCC & GPU (TX2) & $1669.2$ \\
\citet{Hernandez2016embedded} & $\text{CT}_{9\times7}$ - \gls*{SGM} & GPU (TX1) & $747.1$ \\
\midrule
\RESSTAC-CUDA & $\text{\gls*{CT}}_{5\times5}$ - \gls*{SGM} & GPU (AGX) & $652.7$ \\
\RESSTAC-CUDA & $\text{\gls*{CT}}_{9\times7}$ - \gls*{SGM} & GPU (AGX) & $644.9$ \\
\RESSTAC-CUDA & $\text{\gls*{CT}}_{5\times5}$ - \gls*{SGM} - fine & GPU (AGX) & $640.9$ \\
\RESSTAC-CUDA & $\text{\gls*{CT}}_{9\times7}$ - \gls*{SGM} - fine & GPU (AGX) & $633.1$ \\
\RESSTAC-CUDA & $\text{\gls*{CT}}_{5\times5}$ - \gls*{SGM} - exact-cc & GPU (AGX) & $365.7$ \\
\RESSTAC-CUDA & $\text{\gls*{CT}}_{9\times7}$ - \gls*{SGM} - exact-cc & GPU (AGX) & $361.8$ \\
\RESSTAC-CUDA & $\text{\gls*{NCC}}_{5\times5}$ - \gls*{SGM} & GPU (AGX) & $442.4$ \\
\RESSTAC-CUDA & $\text{\gls*{NCC}}_{9\times9}$ - \gls*{SGM} & GPU (AGX) & $344.1$ \\
\RESSTAC-NEON & $\text{\gls*{CT}}_{5\times5}$ - \gls*{SGM} & CPU & $166.2$ \\

\bottomrule
\end{tabular}
\caption{
	\glsreset{MDEs}
	Throughput achieved by our approach and selected embedded algorithms from literature.
	The throughput is measured in \gls*{MDEs}. %
	All the measurements for our approach were done on the NVIDIA Jetson Xavier AGX board, with the power setting set to maximum performance.
	} %
\label{tab:throughput}
\end{table}

\begin{table}[b!]
\centering
\begin{tabular}{|l|l|l|r|r|r|} \toprule
\multirow{2}{*}{Approach} & \multirow{2}{*}{Configuration} & \multirow{2}{*}{HW Device} & Throughput & \multicolumn{2}{c|}{Accuracy} \\ \cline{5-6}
& & & \scriptsize{(in \MDEs)} & D1-all (Est.) & Density \\ 
\toprule
\RESSTAC-CUDA & $\text{\gls*{CT}}_{5\times5}$ - 4-Path-\gls*{SGM} & GPU (AGX) & $924.1$\,\scriptsize{($\times1.42$)} & $4.3\,\%$\,\scriptsize{($+0.0$)} & $88.1\,\%$\,\scriptsize{($-0.7$)} \\
\RESSTAC-CUDA & $\text{\gls*{CT}}_{9\times7}$ - 4-Path-\gls*{SGM} & GPU (AGX) & $904.4$\,\scriptsize{($\times1.40$)} & $4.2\,\%$\,\scriptsize{($+0.2$)} & $89.2\,\%$\,\scriptsize{($-0.8$)} \\
\RESSTAC-CUDA & $\text{\gls*{NCC}}_{5\times5}$ - 4-Path-\gls*{SGM} & GPU (AGX) & $615.4$\,\scriptsize{($\times1.39$)} & $4.3\,\%$\,\scriptsize{($+0.0$)} & $88.6\,\%$\,\scriptsize{($-1.4$)} \\
\RESSTAC-CUDA & $\text{\gls*{NCC}}_{9\times9}$ - 4-Path-\gls*{SGM} & GPU (AGX) & $436.5$\,\scriptsize{($\times1.27$)} & $4.6\,\%$\,\scriptsize{($+0.2$)} & $89.5\,\%$\,\scriptsize{($-1.6$)} \\
\RESSTAC-NEON & $\text{\gls*{CT}}_{5\times5}$ - 4-Path-\gls*{SGM} & CPU & $241.8$\,\scriptsize{($\times1.45$)} & $4.8\,\%$\,\scriptsize{($+0.2$)} & $89.3\,\%$\,\scriptsize{($-0.7$)} \\

\bottomrule
\end{tabular}
\caption{ %
	Throughput and accuracy achieved with a reduced number of aggregation paths in the \gls*{SGM} optimization. %
	Instead of eight aggregation paths, only the two horizontal and the two vertical paths were used. %
	The accuracy in terms of error rate and density was measured on the \KITTI 2015 stereo benchmark. %
	} %
\label{tab:throughput_4path}
\end{table}

\begin{figure}[b!] %
	\resizebox{\columnwidth}{!}{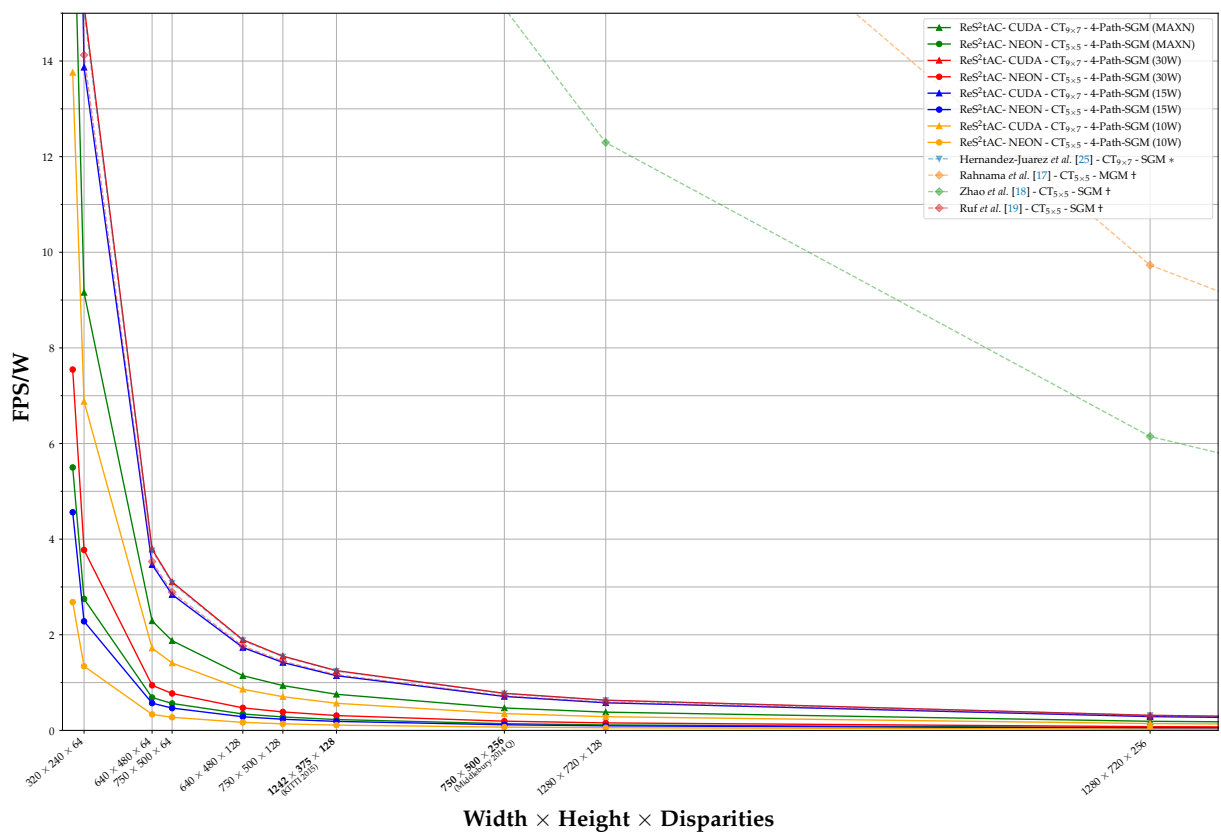}%
	\caption{ %
		Expected \gls*{FPSW} for sets of different image sizes and disparity ranges, based on the throughput and the power consumption achieved by different configurations and approaches. %
		In terms of our approach, the power settings of the AGX were varied in order to measure its efficiency. 
		Image resolution and disparity ranges corresponding to the \KITTI 2015 and Middlebury 2014 Q benchmark are printed in bold. %
		$\ast$: Approaches from literature deployed on embedded \gls*{GPU} hardware. %
		$\dagger$: Approaches from literature deployed on embedded \gls*{FPGA} hardware. %
	}
	\label{fig:fpsPerW_curves}
\end{figure}

A common way to further increase the throughput of the \gls*{SGM} algorithm is to reduce the number of aggregation paths. %
Most of the implementations aggregated the matching costs for each pixel along eight concentric paths. %
However, studies \citep{Banz2010realtime, Hernandez2016embedded} suggest that a reduction of the aggregation paths from eight to four does not have a significant negative impact on the accuracy of the resulting disparity map, while greatly increasing the processing speed. %
This is also supported by our experiments, in which we have left out the diagonal aggregation paths of the \gls*{SGM} optimization, since they are the longest ones, and only regularized the cost volume with the two horizontal and the two vertical aggregation paths. %
The results of our experiments are listed in \Cref{tab:throughput_4path}, showing an increase in the throughput by a factor of up $1.45$, while reducing the accuracy and the disparity by a maximum of 0.2\percent and 1.6\percent respectively. %
In this, we outperform the approach of \citet{Hernandez2016embedded}, which is comparable to ours, in terms of accuracy and throughput. %

With the throughput listed in \Cref{tab:throughput,tab:throughput_4path}, we calculated the expected \gls*{FPS}, which are to be achieved for sets of different image sizes and disparity ranges according to \Cref{eq:fps}, and plotted these as curves in \Cref{fig:fps_curves}. %
In this, we have not plotted all of our configurations, but selected one for each cost function and hardware, as well as the corresponding versions with only four paths in the \gls*{SGM} optimization.
Additionally, we have selected one configuration that performs a subpixel disparity refinement for comparison. %
We have also plotted the \gls*{FPS} curves for the related approaches from literature deployed on \gls*{FPGA} ($\dagger$) and \gls*{GPU} ($\ast$) architectures, as well as one curve achieved by our approach on a high-end desktop NVIDIA RTX 2070 Super GPU. %
We have printed the image sizes and disparity ranges corresponding to the \KITTI 2015 and Middlebury 2014 Q benchmark in bold. %
The curves provide a good visual representation of the throughput listed in \Cref{tab:throughput,tab:throughput_4path} and show that with decreasing complexity, \ie a reduced image size and disparity range, the frame rates increase rapidly. %
Thus, the curves reveal how different configurations, approaches and utilized hardware compare in terms of processing speed. %

\glsreset{FPSW}

A key characteristic of embedded systems is their power consumption and, depending on what platform the system is deployed, this can be a very crucial characteristic. %
The simple metric of \gls*{FPSW} helps to quantify the efficiency of image processing algorithms with respect to the power consumption of the system on which they are deployed. %
The NVIDIA Jetson Xavier AGX, on which we have deployed our approach, allows to set four different power settings, namely:
\begin{description}
\item[MAXN] This is the setting enabling the maximum performance. %
With this, all eight cores of the ARM CPU are activated and can clock up to a maximum of $2.3$\GHz.
The maximum clock rate of the GPU is set to $1.4$\GHz. %
This is the setting with which all previous experiments were conducted. %
\item[30\W] In this, again all eight cores of the CPU are enabled. %
However, they are restricted to a maximum clock rate of $1.2$\GHz. %
Furthermore, the clock rate of the GPU is restricted to $905$\MHz.
\item[15\W] In this setting, four cores of the CPU are enabled which clock at a maximum rate of $1.2$\GHz, while the GPU clocks up to $675$\MHz.
\item[10\W] In the smallest setting, only two cores of the CPU are enabled with a maximum of $1.2$\GHz and the clock rate of the GPU is restricted to only $522$\MHz.
\end{description}

In \Cref{fig:fpsPerW_curves}, we have plotted the \gls*{FPSW} which are expected to be achieved by different configurations of our approach, as well as by some approaches form the literature, again, depending on different image sizes and disparity ranges. %
These calculations are based on the throughput and power consumption measured or stated. %
In terms of our approach, we have selected the two configurations reaching the highest throughput on the GPU and the CPU. %
We have varied the power setting and measured the throughput and power consumption, the latter being provided by internal sensors of the AGX. %
Note that the actual power consumption on the AGX does not coincide with the statement of the power setting, as the latter one only indicates an upper bound on the consumption.
From literature, we have selected approaches which provide a value on the power consumption in addition to the throughput or frame rate. %
Unfortunately, in terms of related work that has deployed stereo algorithms on embedded GPUs, this was only done by \citet{Hernandez2016embedded}. %
\begin{figure}[t!] %
  \centering
	\resizebox{0.85\columnwidth}{!}{
\begingroup%
  \makeatletter%
  \providecommand\color[2][]{%
    \errmessage{(Inkscape) Color is used for the text in Inkscape, but the package 'color.sty' is not loaded}%
    \renewcommand\color[2][]{}%
  }%
  \providecommand\transparent[1]{%
    \errmessage{(Inkscape) Transparency is used (non-zero) for the text in Inkscape, but the package 'transparent.sty' is not loaded}%
    \renewcommand\transparent[1]{}%
  }%
  \providecommand\rotatebox[2]{#2}%
  \newcommand*\fsize{\dimexpr\f@size pt\relax}%
  \newcommand*\lineheight[1]{\fontsize{\fsize}{#1\fsize}\selectfont}%
  \ifx\svgwidth\undefined%
    \setlength{\unitlength}{842.51607099bp}%
    \ifx\svgscale\undefined%
      \relax%
    \else%
      \setlength{\unitlength}{\unitlength * \real{\svgscale}}%
    \fi%
  \else%
    \setlength{\unitlength}{\svgwidth}%
  \fi%
  \global\let\svgwidth\undefined%
  \global\let\svgscale\undefined%
  \makeatother%
  \begin{picture}(1,0.5784383)%
    \lineheight{1}%
    \setlength\tabcolsep{0pt}%
    \put(0,0){\includegraphics[width=\unitlength,page=1]{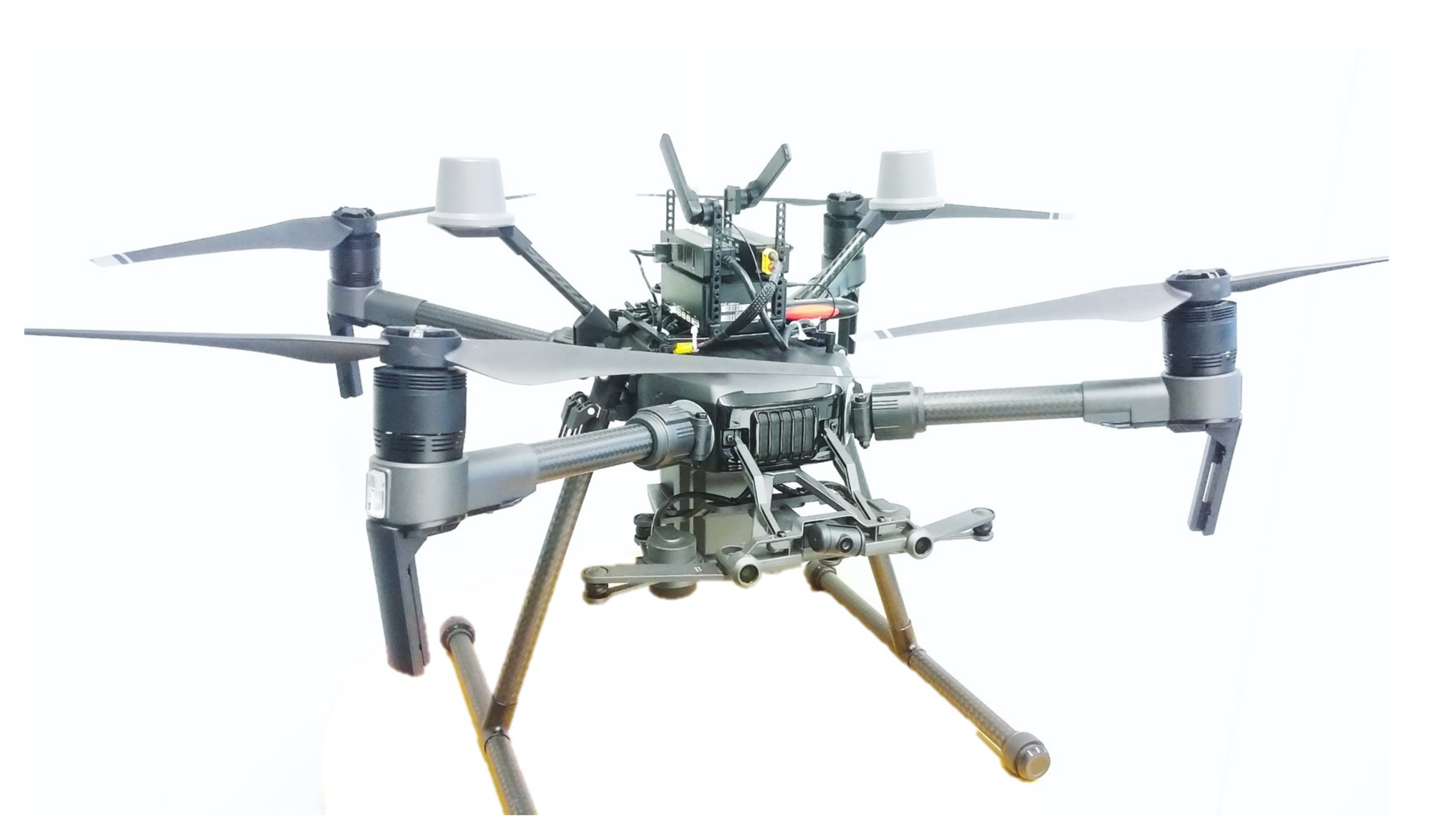}}%
    \put(0.18559452,0.53456235){\color[rgb]{0,0,0}\makebox(0,0)[lt]{\lineheight{1.25}\smash{\begin{tabular}[t]{l}\huge{DJI Manifold 2-G}\end{tabular}}}}%
    \put(0.75256249,0.1064586){\color[rgb]{0,0,0}\makebox(0,0)[lt]{\lineheight{1.25}\smash{\begin{tabular}[t]{l}\huge{Stereo Vision Sensor}\end{tabular}}}}%
    \put(0,0){\includegraphics[width=\unitlength,page=2]{DJIMatrice210.pdf}}%
  \end{picture}%
\endgroup%
}%
	\caption{ %
		The system used for use-case specific experiments for real-time stereo processing with an embedded CUDA device on board a low-cost \gls*{UAV}: %
		A DJI Matrice 210 v2 RTK equipped with a DJI Manifold 2-G processing unit. %
		The integrated stereo vision sensor is used as stereo camera. %
	}
	\label{fig:djiMatrice}
\end{figure}

The curves in \Cref{fig:fps_curves,fig:fpsPerW_curves} clearly reveal the superiority of \gls*{FPGA}-based approaches over those deployed on GPUs. %
Not only do they achieve much higher frame rates, but also require significantly less power and, in turn, also achieve higher \gls*{FPSW}. %
However, the emerging embedded GPUs also achieve quite reasonable frame rates with respect to their power consumption, and depending where the systems are deployed, \eg quadrotor-based systems, the power required by the GPU is negligible, when compared to that required by the rotors. %
But more on this is provided in the discussion on what the results mean for our use-case (\cf \Cref{sec:discussion}). %
Interestingly, the curves in \Cref{fig:fpsPerW_curves} show that both the CUDA and NEON implementation of our approach achieve the best efficiency on the 30\W power setting. %
Even the CUDA implementation being run on the 15\W power setting is still more efficient than the same run on the setting with maximum performance. %
We assume that this is the result of clocking down the CPU, since it is less efficient than the GPU.
Nonetheless, the 30\W and 15\W power setting reduce the throughput by approximately 29\percent and 42\percent, respectively, compared to that achieved on MAXN.

\subsection{Qualitative and quantitative evaluation of real-time stereo processing on-board low-cost UAVs}
\label{sec:Experiments-Qualitative}

As part of our use-case specific experiments, in which we want to bring real-time stereo processing with an embedded CUDA device on board a low-cost \gls*{UAV}, we have equipped a DJI Matrice 210v2 RTK with a DJI Manifold 2-G processing unit, which is based on the NVIDIA Jetson TX2 architecture equipped with a 4-core 64\,bit ARMv8 CPU and a 256-core Pascal GPU. %
As a stereo camera we have used the integrated stereo vision sensor, which can be accessed by the Manifold through the DJI onboard SDK. %
Our setup is illustrated in \Cref{fig:djiMatrice}. %

\begin{table}[b!]
\centering
\resizebox{\textwidth}{!}{%
\begin{tabular}{|l|l|l|r|r|r|r|} \toprule
\multirow{3}{*}{Approach} & \multirow{3}{*}{Configuration} & \multirow{3}{*}{HW Device} & Throughput & \multicolumn{2}{c|}{Frame Rate} \\ \cline{5-6}
& & & \scriptsize{(in \MDEs)} & \scriptsize{at $640\times480\times64$\px} & \scriptsize{at $320\times240\times64$\px}\\ 
& & & & \scriptsize{(in \fps)} & \scriptsize{(in \fps)} \\ 
\toprule
\RESSTAC-CUDA & $\text{\gls*{CT}}_{9\times7}$ - 4-Path-\gls*{SGM} & GPU (TX2) & $304.7$  & $15.5$ & $62.0$  \\
\RESSTAC-NEON & $\text{\gls*{CT}}_{5\times5}$ - 4-Path-\gls*{SGM} & CPU & $102.2$ & $5.2$ & $20.8$ \\
\bottomrule
\end{tabular}
}
\caption{ %
	Throughput and frame rate achieved by two configurations on the DJI Manifold, equipped with a Jetson TX2. %
	} %
\label{tab:throughput_dji}
\end{table}

In the maximum power setting (MAXN), the ARM Cortex A57 CPU of the TX2 inside the Manifold has four cores with a maximum clock rate of 2\GHz, while the built-in Tegra GPU clocks up to a maximum of $1.3$\GHz.
The integrated vision sensor provides non-rectified, grayscale stereo image pairs with an image resolution of up to $640\times480$\px at a frame rate of $20$\fps. %
We have used the standard calibration routine of OpenCV to calibrate the stereo sensor and, in turn, precompute the rectification maps, needed to transform the input images into a rectified stereo pair prior to the actual stereo processing. %
The integrated stereo vision sensor also provides precomputed disparity maps at a frame rate of $10$\fps and with an image resolution of $320\times240$\px \citep{DJI2021osdk}. %

\begin{figure}[t!]
     \centering
     \subfloat{\includegraphics[width=0.49\textwidth]{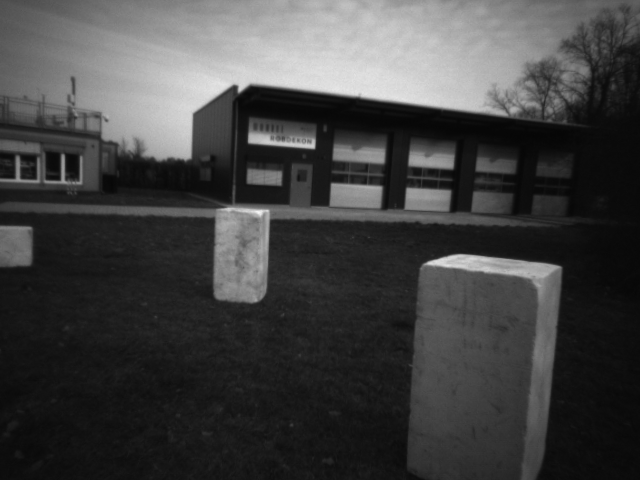}} \hfill
	 \subfloat{\includegraphics[width=0.49\textwidth]{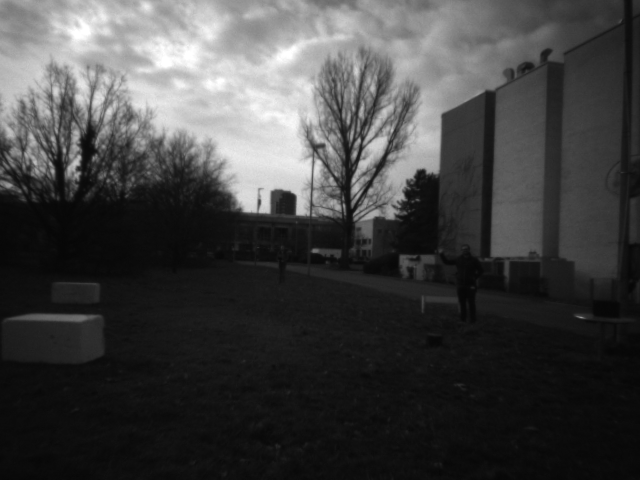}} \hfill
	 \subfloat{\includegraphics[width=0.49\textwidth]{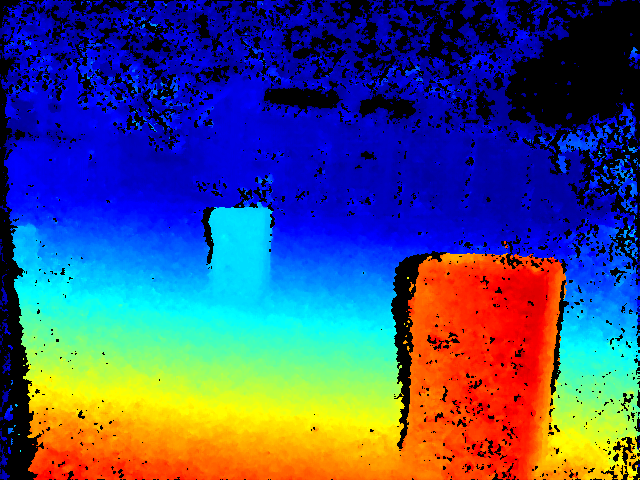}} \hfill
	 \subfloat{\includegraphics[width=0.49\textwidth]{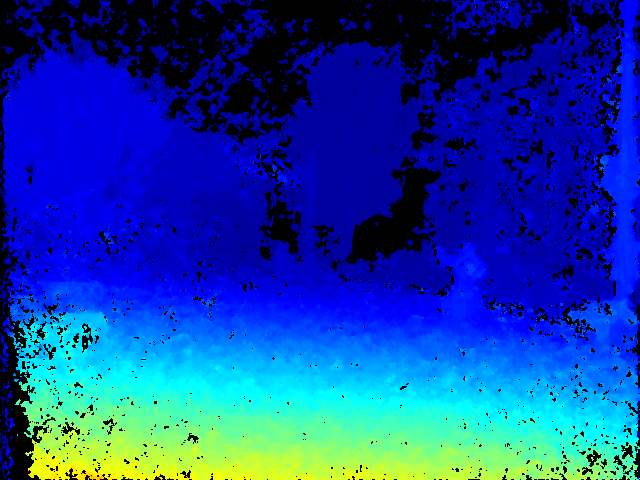}} \hfill
     \caption{
     	Qualitative results of our approach run on the DJI Manifold 2-G on top of the DJI Matrice 210v2 RTK, using the data of the stereo vision sensor as input. %
     	In this, the \gls*{UAV} was flying 1-2\m above the ground in order to demonstrate the ability of our approach to appropriately estimate the scene depth. %
     	\textbf{Top row:} Rectified reference images. %
		\textbf{Bottom row:} Corresponding disparity maps, color-coded with the jet color map, going from red (near), over yellow and green to blue (far).
     }
     \label{fig:DJI_result}
\end{figure}

We have deployed and tested two configurations of our approach, one running on the GPU and the other on the CPU, namely: \RESSTAC-CUDA with $\text{\gls*{CT}}_{9\times7}$ - 4-Path-\gls*{SGM} and \RESSTAC-NEON with $\text{\gls*{CT}}_{5\times5}$. %
Both configurations rely on the Hamming distance of the \acrlong*{CT} as a cost function and use only four paths in the \gls*{SGM} optimization in order to reach a higher throughput. %
The throughput and frame rates, as well as qualitative results achieved by the two configurations are listed in \Cref{tab:throughput_dji} and \Cref{fig:DJI_result} respectively. %

\section{Discussion}
\label{sec:discussion}

In the following, we will discuss the findings of our experiments with respect to three different aspects, namely accuracy (\Cref{sec:discussion_accuracy}), processing speed (\Cref{sec:discussion_speed}) and power consumption (\Cref{sec:discussion_power}) in the light of possible applications, before finishing this paper with a short summary, a conclusion and an outlook. %

\subsection{Accuracy}
\label{sec:discussion_accuracy}

The quantitative evaluation on the \KITTI 2015 stereo benchmark (\Cref{tab:Kitti2015_accuracy}) reveals a high and state-of-the-art accuracy of our approach, both in the actual estimated disparity map (Est), in which inconsistent regions are removed, as well as the interpolated versions (All). %
The latter are being used as part of the standard evaluation routine of the benchmark. %
Understandably, the accuracy of the interpolated disparity maps is used as a primary ranking in the benchmark, since it allows to compare disparity maps with different densities, however, in the assessment of the performance of our approach, the accuracies of the filtered disparity maps are of greater importance. %
In particular, when looking at the qualitative results (\Cref{fig:Kitti2015_result}) and the stated densities of the estimated disparity maps, it becomes clear that the few areas, that are being removed by the post-filtering mostly arise in occluded areas and are thus legitimately removed, since it is not possible for the algorithm to reason about the depth in areas which are only seen by one camera. %

Unfortunately, the results of the Middlebury 2014 stereo benchmark render the accuracy of our approach far from the state-of-the-art. %
With an error rate of over $35$\percent for three out of four accuracy levels, the results are not really satisfying. %
Apart from the high-resolution ground truth and the high accuracy levels in the evaluation, we assume that these poor results can also be attributed to the fact that the resolution, with which the disparity maps are being evaluated, is $4\times$ bigger than the input resolution, introducing a lot of error in the process of upscaling and interpolation. %
Thus, we have also evaluated the results on ground truth data with only a quarter of the original image resolution and found that the error rate is on average reduced by way over $50$\percent. %
For the configuration with a $5\times5$ \acrlong*{CT}, being the best achieving configuration on this dataset, the accuracies calculated are $35.8$\percent (bad0.5), $14.2$\percent (bad1), $7.4$\percent (bad2) and $4.9$\percent (bad4), which is in the comparable range to that obtained on the \KITTI benchmark. %

As the name of our approach suggests, it is intended for real-time, rather than high accuracy, stereo processing. %
Our main use-case is the deployment on low-cost \glspl*{UAV}, equipped with embedded ARM or CUDA hardware, with the purpose of collision detection and avoidance or real-time 3D mapping and scene reconstruction. %
In the light of this use-case, we believe that the \KITTI benchmark is more appropriate, since it comprises image data that depict real-world scenes in a quality that can also be expected from cameras mounted on \glspl*{UAV}. %
Moreover, for collision avoidance or real-time 3D mapping, it is not necessary to have disparity maps with high subpixel accuracy as evaluated by the Middlebury benchmark. %
It is more important to reliably detect the location of objects in the perceived scene and extensively reconstruct their appearance. %
Not only do the quantitative results prove the accuracy of the disparity maps, the qualitative presentation of some of the disparity maps also shows that our approach is able to reveal objects, which are only visible by a second glance, such as the person on the the right of \Cref{fig:Kitti2015_result} (row 2, column 2) or \Cref{fig:DJI_result} (column 2). %

With respect to the \KITTI 2015 benchmark, most of the configurations of our approach outperform the other approaches that perform real-time stereo estimation on embedded hardware. %
Even compared to the baseline algorithms, our results are compatible, especially when considering the frame rates we achieve. %
Furthermore, our experiments on the effects of subpixel disparity refinement (\cf \Cref{tab:Kitti2015_disparityRefinement,tab:Middlebury2014_disparityRefinement}) have shown, that its use can increase the the accuracy by up to $35$\percent without significantly decreasing the throughput (\cf \Cref{tab:throughput}). %
We therefore will consider using the post-refinement as part of the standard pipeline. %

\subsection{Processing speed}
\label{sec:discussion_speed} 
\glsreset{COTS}
Compared to the related work from literature, the throughput and processing speed of our approach are not very impressive. %
We expected to reach a significantly lower throughput than approaches running on \glspl*{FPGA} \citep{Zhao2020fp, Rahnama2018r3sgm}, as well as a slightly lower throughput compared to approaches that do not rely on a computationally expensive optimization scheme but run on an embedded \gls*{GPU} \citep{Cui2019real, Chang2020zigzag}. %
However, the $\text{\gls*{CT}}_{9\times7}$ - \gls*{SGM} configuration of our approach has a lower throughput than that of a comparable configuration by \citet{Hernandez2016embedded}, while at the same time running on a hardware generation that is two times newer and that has twice the number of CUDA cores. %
This is not very satisfactory and something we will need to further look into in the scope of future work. %
One difference between the work of \citet{Hernandez2016embedded} and ours is that in our time measurements we include the data transfer to and from the memory of the GPU. %
\citet{Hernandez2016embedded} argue that the processing can be overlapped with the computation and thus is not relevant for the computation of the throughput. %
However, in our case, the data transfer only takes up $4$-$6$\percent of the processing time, which is not enough to reach the throughput of \citet{Hernandez2016embedded}, if omitted.
Other optimization steps are the utilization of \gls*{SIMD} instructions to vectorize the cost aggregation with the CUDA kernels, or to streamline the aggregation of the last path and the disparity computation to reduce memory access, which should lead to a $1.35\times$ performance speed-up \citep{Hernandez2016embedded}. %
A third and significant difference between our approach and that of \citet{Hernandez2016embedded} with respect to performance is that \citet{Hernandez2016embedded} refrain from any post-processing like left-right consistency check or median filter. %
This allows to reach a higher throughput but, in turn, reduces the accuracy of the results, leading to an error-rate which is twice as high as ours in the actual estimate (\cf \Cref{tab:Kitti2015_accuracy}). %

Furthermore, in the light of our use-case, addressing on-board stereo processing on low-cost \glspl*{UAV}, we believe that to this end the throughput of our approach is sufficient, as another limiting factor is the camera sensor and the data throughput between the sensor and processing board that is provided by \gls*{COTS} systems. %
Since a \gls*{FPGA} is typically located closer to the sensor with a direct and high-bandwidth connection, which allows to stream the image data directly into the memory of the \gls*{FPGA}, a high data throughput is of greater importance. %
However, embedded systems equipped with a \gls*{GPU}, like the NVIDIA Jetson series, that are mounted on \gls*{COTS} \glspl*{UAV}, are usually connected to the sensor via USB or similar, with the CPU capturing the data and storing it in global memory, from where it gets transferred to the device memory of the \gls*{GPU} before it can be processed. %
This process does not allow for a high input frame rate (usually between $20$\fps and $30$\fps), as can be seen in the example of the DJI Matrice 210 (\cf \Cref{sec:Experiments-Qualitative}), and therefore does not require an extremely high throughput. %
Nonetheless, the more time spent on the estimation of the disparity map, the less time is available for the successive interpretation, \eg obstacle detection and avoidance. %
This raises our interest to further look into the optimization of the processing speed in the future. %

\subsection{Power consumption}
\label{sec:discussion_power} 

The final aspect which we want to discuss is the power consumption of our approach running on the NVIDIA Jetson Xavier AGX in the light of deployment on a rotor-based \gls*{UAV} and whether it is feasible to use embedded CPUs for stereo processing. %
The average power consumption of the complete NVIDIA Jetson Xavier AGX (\ie including the GPU, the CPU and the EMC) under the maximum power setting MAXN during the execution of our CUDA- and NEON-based approaches is approximately $20.1$\W and $17.9$\W respectively. %
The DJI Matrice 210v2 RTK is powered by two batteries with a total energy of $349.2$\Wh, which allows a maximum flight time of $33$ minutes when no payload is attached \citep{DJI2020matrice}. %
Thus, during flight, the bare DJI Matrice 210v2 RTK consumes around $634.9$\W per minute. %
This power consumption obviously increases with each gram of payload that is being attached. %
Given these measurements, we can calculate that the power consumption of the AGX under the highest power setting only makes up $3.2$\percent and $2.8$\percent with respect to the total power consumption of the DJI Matrice, thus reducing the flight time by a maximum of $1$ minute, when our CUDA- and NEON-based approaches are executed respectively. %
This is an upper bound on the relative power consumption of the AGX, as the power consumption during flight of the DJI increases when payload is attached. %
The use of other power settings, that will increase the \fpsPerWatt ratio (\cf \Cref{fig:fpsPerW_curves}), but also decrease the absolute frame rate, is dependent on the use-case and whether the gain of few seconds in flight time is more valuable than a major reduction in frame rate. %

The power consumption during the execution of our CUDA-based approach is higher than during the execution of our NEON-based approach. %
This is expected as the GPU, which consumes more power than the CPU, is clocked down when not in use. %
However, the reduced power consumption does not stand in relation to the loss in processing speed of the NEON-based approach compared to the ones based on CUDA. %
This is clearly revealed by the curves in \Cref{fig:fpsPerW_curves}, depicting that the NEON-based approaches running on the CPU have the worst \fpsPerWatt ratio. %
Thus, we conclude that the use of embedded \glspl*{GPU} is preferred over embedded CPUs. 
However, some drones are only equipped with an embedded ARM CPU, \eg VOXL\footnote{\url{https://www.modalai.com/pages/voxl}}, for which a vectorized stereo processing with NEON intrinsics is an option. %
\section{Conclusions}
\label{sec:conclusion}

\sloppy

In conclusion, we present an approach for real-time stereo processing on embedded ARM and CUDA devices, such as those attached to modern low-cost \gls*{COTS} \glspl*{UAV}. %
In this, we have optimized a disparity estimation algorithm for embedded CUDA \glspl*{GPU}, such as the NVIDIA Tegra, by \acrlong*{GPGPU}, as well as for embedded ARM CPUs by utilizing the NEON intrinsics for vectorized \gls*{SIMD} processing. %
We have demonstrated, that our approach reaches state-of-the-art accuracies when evaluated on public stereo benchmark datasets. %
Our CUDA-based implementation for stereo processing on embedded GPUs reaches real-time performance, even though it does not outperform related work in terms processing speed. %
The frame rates of our NEON-based implementation, however, outperform all related work on stereo processing on embedded CPUs. %
In a use-case specific scenario, we have further demonstrated the suitability of our approach for real-time stereo estimation on a low-cost \gls*{COTS} \glspl*{UAV} for the task of obstacle detection and 3D mapping by deploying it on a DJI Matrice 210v2 RTK equipped with a DJI Manifold 2-G. %

We have shown, that in case of rotor-based \glspl*{UAV} a modern embedded \gls*{GPU} is a suitable alternative to an embedded \gls*{FPGA}, especially due to its shorter and thus less expensive developments cycles. %
Even though the GPU has a much greater power consumption than a \gls*{FPGA} and a significantly worse \gls*{FPSW} ratio, its power consumption is negligible compared to the energy needed by rotor-based \glspl*{UAV} during flight and will reduce the flight time of the DJI Matrice 210v2 RTK by a maximum of 1 minute. %
However, for embedded systems with stricter power constraints, a \gls*{FPGA}-based approach should be considered. %
Our experiments have also shown that, although the CPU requires less energy than the GPU, it has the worst \gls*{FPSW} ratio. %
Thus, our optimization based on NEON intrinsics for vectorized SIMD processing should only be used if neither GPU nor FPGA are available. %

Finally, we have also identified few aspects to consider in future work. %
For one, we will need to further investigate which part of our optimization for CUDA enabled GPUs can be further optimized, since our approach does not reach the processing speed of comparable approaches from the literature. %
And secondly, we should also consider other approaches, \eg deep-learning-based algorithms, that are capable of reaching higher accuracies than ours. %
Apart from that, our next steps are the extension of our approach by algorithms for real-time 3D mapping, as well as object and obstacle detection, in order to alleviate the perception of the environment around the \gls*{UAV} and in turn increase its autonomy. %

\vspace{6pt} 



\authorcontributions{Conceptualization, B.R. and J.M.; methodology, B.R. and J.M.; investigation, B.R.; writing--original draft preparation, B.R. and J.M.; writing--review and editing, B.R. and M.W.; supervision, S.H. and J.B.; All authors have read and agreed to the published version of the manuscript.}

\funding{ %
This work was funded by the European Commission under the H2020 Framework Programme for Research and Innovation as part of the \TULIPP project under grant agreement No 688403. %
We also acknowledge support by the KIT-Publication Fund of the Karlsruhe Institute of Technology.
} %

\acknowledgments{We would like to thank Mr. Raphael Senk and Mr. Shreyas Manjunath for their support in setting up and using the \gls*{UAV} with the on-board processing unit.}

\conflictsofinterest{The authors declare no conflict of interest.} 

%

\appendixtitles{yes} 
\appendix

\section{Stereoscopic vision}
\label{sec:appendix_stereo}

\begin{figure}[ht!] %
	\centering
	\resizebox{0.79\columnwidth}{!}{\subimport{figures/}{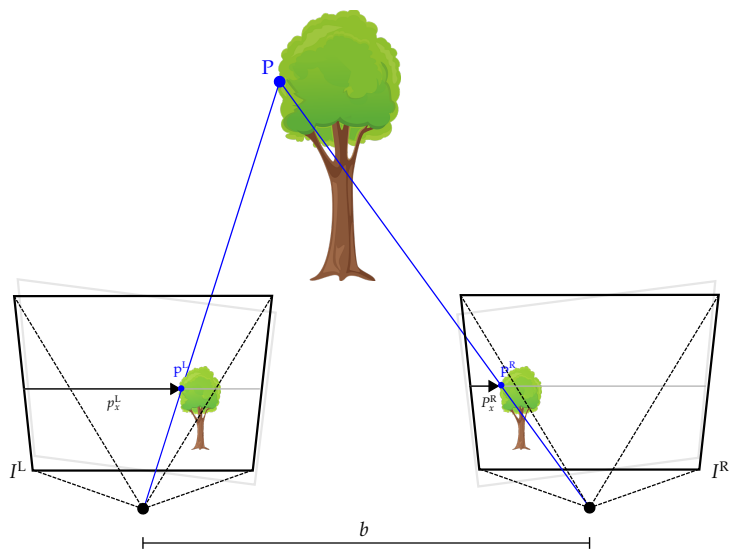}}%
	\caption{
		Illustration of a stereo camera setup. 
		Two cameras, placed apart from each other with a distance $b$ (baseline), observe a scene point $\mathrm{P}$ from two different vantage points. %
		The scene point will appear at different locations, $\mathrm{p^L}$ and $\mathrm{p^R}$, in the two camera images $I^\mathrm{L}$ and $I^\mathrm{R}$. %
		The difference between the $x$ coordinate of the two image points, $p^\mathrm{L}_x$ and $p^\mathrm{R}_x$ is called the disparity $d$. 
		If the camera rig is not calibrated, the two image planes are not rectified (light gray boxes), \ie the image planes are not co-planar. %
		By transforming $I^\mathrm{L}$ and $I^\mathrm{R}$ through rectification, the epipolar lines will align with the image row of the respective image points (gray lines).
	}
	\label{fig:stereo_setup}
\end{figure}

It is well known, that with the positions of two image points, $\mathrm{p^L}$ and $\mathrm{p^R}$, in the images of the left and right camera of a calibrated stereo camera setup, belonging to the same scene point $\mathrm{P}$, it is possible to compute the 3D position of $\mathrm{P}$ relative to the stereo rig. %
Thus, finding such correspondences is the essential task in the process of stereo processing, and in turn dense disparity or depth estimation. %
Generally speaking, a correspondence search between an image point in the one camera and the image projection of the same scene point in the other camera has to be done along the epipolar curve, which is the viewing ray, going through the image point of the first camera, projected into the image of the second camera. %
This correlation is formulated by the epipolar constraints. %
When the intrinsic parameters of the cameras are known, it is possible to account for image distortions, transforming the curve into an epipolar line. %
Furthermore, if the relative position and orientation between the two cameras of the stereo rig are known, the camera images can be rectified, \ie they can be transformed onto a common image plane, and in turn the epipolar lines can be transformed to coincide with the image row of the image points. %
Thus, the difference between the image positions of $\mathrm{p^L}$ and $\mathrm{p^R}$ is reduced to a horizontal shift, the so-called disparity $d_\mathrm{p} = \left| p^\mathrm{L}_x - p^\mathrm{R}_x \right|$. %
This is illustrated by \Cref{fig:stereo_setup}.

\section{Left-right consistency check}
\label{sec:appendix_consistencycheck}

\begin{figure}[ht] %
	\centering %
	\begin{tikzpicture}
	
	\node[rectangle, minimum width=4cm, minimum height=2.5cm, fill=KITcyanblue, fill opacity=0.2] (ref) at (2,5) {};
	\node[rectangle, minimum width=4cm, minimum height=2.5cm, fill=KITcyanblue, fill opacity=0.2] (mat) at (8,5) {};

	\draw [decorate,decoration={brace,amplitude=10pt,mirror,raise=4pt}]
	(ref.south west) -- (ref.south east) node [black,midway,yshift=-1cm] {$D^{\mathrm{L}}$};
	\draw [decorate,decoration={brace,amplitude=10pt,mirror,raise=4pt}]
	(mat.south west) -- (mat.south east) node [black,midway,yshift=-1cm] {$D^{\mathrm{R}}$};
	
	\draw[circle,minimum size=0.2cm,inner sep=0pt] (3,5.5) node[fill] (refPixel){};
	\draw[circle,minimum size=0.2cm,inner sep=0pt] (9,5.5) node[fill] (matPixel1){};
	\draw[circle,minimum size=0.2cm,inner sep=0pt] (7,5.5) node[fill] (matPixel2){};

	\draw [->,thick] (refPixel) to [out=45,in=135] (matPixel1);
	\draw [->,thick] (matPixel1) -- (matPixel2);
	
	\node[anchor=north] (t4) at (8,6) {$d^{\mathrm{L}}$};
	\node[anchor=north] (t1) at (2.8,5.4) {$d^{\mathrm{L}} = D^{\mathrm{L}}(x,y)$};
	\node[anchor=north west] (t2) at (9.1,5.4) {$D^{\mathrm{R}}(x,y)$};
	\node[anchor=north east] (t3) at (7.7,5.4) {$d^{\mathrm{R}} = D^{\mathrm{R}}(x-d^{\mathrm{L}},y)$};
	
	\node[] (t10) at (5,2.5) {$|d^{\mathrm{L}} - d^{\mathrm{R}}| \leq 1$};

\end{tikzpicture} %
	\caption { %
		With the disparity maps $D^{\mathrm{L}}$ and $D^{\mathrm{R}}$ for the reference and matching image respectively, a consistency check according to \Cref{eq:LR_check} is performed in order to find occlusions. %
	}
	\label{fig:cccheck_illus} %
\end{figure}
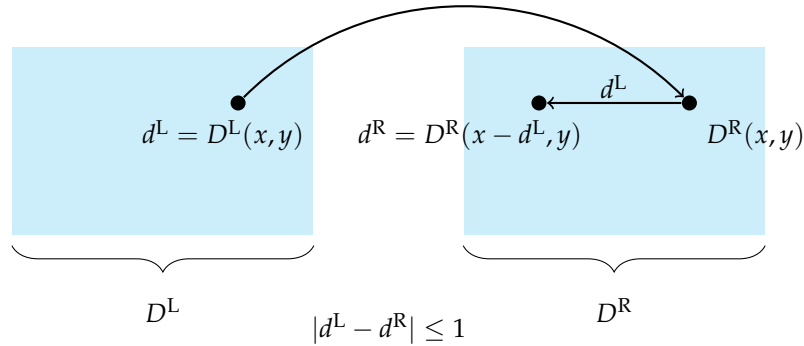

The existence of occluded pixels that arise from areas, which are observed by one camera and yet occluded in the field-of-view of the other camera, is inherent to disparity and depth maps that are computed from a conventional stereo setup, consisting of only two cameras. %
A typical approach to detect and filter such pixels is the left-right consistency check. %
As illustrated by \Cref{fig:cccheck_illus}, the disparities that are stored in the disparity map of the left image $D^{\mathrm{L}}$ are compared with the corresponding disparities in the right disparity map $D^{\mathrm{R}}$ and invalidated if they differ by a certain threshold, usually 1: %
 
\begin{equation} %
\label{eq:LR_check} %
D(\mathrm{p}) = %
\begin{cases} %
	D^\mathrm{L}(\mathrm{p}) ,\ \text{if}\ \left| D^\mathrm{L}(\mathrm{p}) - D^\mathrm{R}(p_x - D^\mathrm{L}(\mathrm{p}), p_y) \right| \leq 1 \\ %
	inv ,\ \text{otherwise}. %
\end{cases} %
\end{equation} %

\section{General purpose computing on CUDA-enabled GPUs}
\label{sec:appendix_gpu_programming}

\glsreset{GPU}
\glsreset{GPGPU}

\Glspl*{GPU} are designed for massively parallel processing. %
The number of processing units that are integrated into modern GPU hardware exceeds the number of cores available on a conventional CPU by far. %
Even embedded GPUs, like the one built into the NVIDIA Jetson Xavier AGX, have up to 100$\times$ more cores than high-end desktop CPUs. %
However, the processing units on a GPU are less powerful and flexible than those of a CPU. %
While the CPU is designed to do arbitrary processing tasks, the cores of the GPUs are intended for the parallel processing of small and dedicated instructions on a large amount of data simultaneously. %

Massively parallel general purpose processing on NVIDIA GPUs is alleviated by the CUDA-API, which allows the deployment of routines and functions for data processing in the form of so-called CUDA kernels on the GPU. %
When deployed, such a kernel is instantiated inside a high number of threads, which are being distributed among the available processing units and are each processing a different subset of the data. %
For a better abstraction and handling, the threads are logically grouped into thread-blocks, which share a local memory space and can thus exchange data between each other. %
Furthermore, the execution of all threads within one thread-block can be halted and synchronized. %
All thread-blocks are grouped into a grid. %
The grid size and the number of threads inside a grid is used to parameterize the instantiation of the kernel.
The actual execution of the threads on a processing unit is always done in groups of 32, which is referred to as a CUDA warp. %
For an efficient \gls*{GPGPU}, the developer is compelled to consider some design guidelines:
\begin{itemize} %
	\item Reduction of global memory access due to higher latency. %
	Instead, data which gets processed multiple times by threads in a thread-block should be cached inside the shared memory space.
	\item Pooling of global memory access and reduction of non-contiguous data storage. %
	\item Efficient and maximum utilization of hardware resources. %
\end{itemize} %

\section{Map reduce method}
\label{sec:appendix_mapreduce}

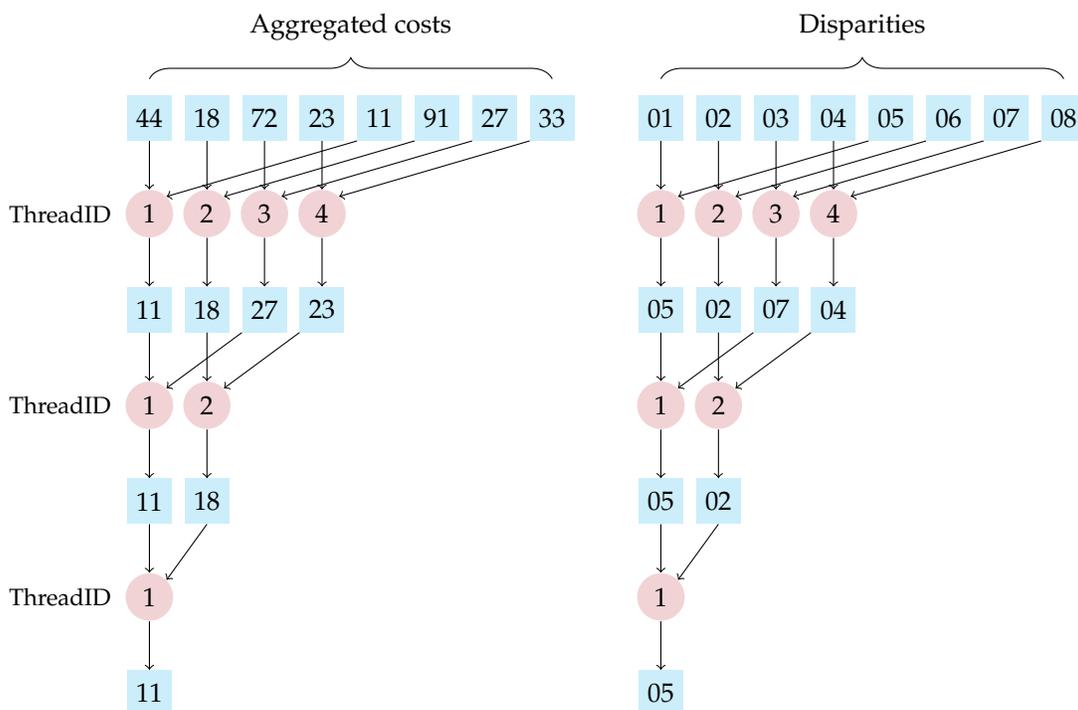
\begin{figure}[ht!] %
	\centering %
	\begin{tikzpicture}[scale=0.85]
	
	\node[rectangle, minimum width=0.4cm, minimum height=0.6cm, fill=KITcyanblue,
				opacity=0.2, text opacity=1.0] (t1) at (0.9,10) {$44$};
	\node[rectangle, minimum width=0.4cm, minimum height=0.6cm, fill=KITcyanblue,
			opacity=0.2, text opacity=1.0] (t2) at (1.8,10) {$18$};
	\node[rectangle, minimum width=0.4cm, minimum height=0.6cm, fill=KITcyanblue,
			opacity=0.2, text opacity=1.0] (t3) at (2.7,10) {$72$};
	\node[rectangle, minimum width=0.4cm, minimum height=0.6cm, fill=KITcyanblue,
			opacity=0.2, text opacity=1.0] (t4) at (3.6,10) {$23$};
	\node[rectangle, minimum width=0.4cm, minimum height=0.6cm, fill=KITcyanblue,
			opacity=0.2, text opacity=1.0] (t5) at (4.5,10) {$11$};
	\node[rectangle, minimum width=0.4cm, minimum height=0.6cm, fill=KITcyanblue,
			opacity=0.2, text opacity=1.0] (t6) at (5.4,10) {$91$};
	\node[rectangle, minimum width=0.4cm, minimum height=0.6cm, fill=KITcyanblue,
			opacity=0.2, text opacity=1.0] (t7) at (6.3,10)  {$27$};
	\node[rectangle, minimum width=0.4cm, minimum height=0.6cm, fill=KITcyanblue,
			opacity=0.2, text opacity=1.0] (t8) at (7.2,10) {$33$};
			
	\node[circle, fill=KITred, opacity=0.2, text opacity=1.0 ] (c1) at (0.9,8.5) {$1$};
	\node[circle, fill=KITred, opacity=0.2, text opacity=1.0 ] (c2) at (1.8,8.5) {$2$};
	\node[circle, fill=KITred, opacity=0.2, text opacity=1.0 ] (c3) at (2.7,8.5) {$3$};
	\node[circle, fill=KITred, opacity=0.2, text opacity=1.0 ] (c4) at (3.6,8.5) {$4$};
	
	\draw[->] (t1.south) -- (c1.north);
	\draw[->] (t5.south west) -- (c1.north east);
	\draw[->] (t2.south) -- (c2.north);
	\draw[->] (t6.south west) -- (c2.north east);
	\draw[->] (t3.south) -- (c3.north);
	\draw[->] (t7.south west) -- (c3.north east);
	\draw[->] (t4.south) -- (c4.north);
	\draw[->] (t8.south west) -- (c4.north east);
	
	\node[rectangle, minimum width=0.4cm, minimum height=0.6cm, fill=KITcyanblue,
				opacity=0.2, text opacity=1.0] (t11) at (0.9,7) {$11$};
	\node[rectangle, minimum width=0.4cm, minimum height=0.6cm, fill=KITcyanblue,
			opacity=0.2, text opacity=1.0] (t22) at (1.8,7) {$18$};
	\node[rectangle, minimum width=0.4cm, minimum height=0.6cm, fill=KITcyanblue,
			opacity=0.2, text opacity=1.0] (t33) at (2.7,7) {$27$};
	\node[rectangle, minimum width=0.4cm, minimum height=0.6cm, fill=KITcyanblue,
			opacity=0.2, text opacity=1.0] (t44) at (3.6,7) {$23$};
			
	\draw[->] (c1.south) -- (t11.north);
	\draw[->] (c2.south) -- (t22.north);
	\draw[->] (c3.south) -- (t33.north);
	\draw[->] (c4.south) -- (t44.north);
			
	\node[circle, fill=KITred, opacity=0.2, text opacity=1.0 ] (c11) at (0.9,5.5) {$1$};
	\node[circle, fill=KITred, opacity=0.2, text opacity=1.0 ] (c22) at (1.8,5.5) {$2$};

	\draw[->] (t11.south) -- (c11.north);
	\draw[->] (t33.south west) -- (c11.north east);
	\draw[->] (t22.south) -- (c22.north);
	\draw[->] (t44.south west) -- (c22.north east);
	
	\node[rectangle, minimum width=0.4cm, minimum height=0.6cm, fill=KITcyanblue,
		opacity=0.2, text opacity=1.0] (t111) at (0.9,4) {$11$};
	\node[rectangle, minimum width=0.4cm, minimum height=0.6cm, fill=KITcyanblue,
		opacity=0.2, text opacity=1.0] (t222) at (1.8,4) {$18$};
		
	\draw[->] (c11.south) -- (t111.north);
	\draw[->] (c22.south) -- (t222.north);
		
	\node[circle, fill=KITred, opacity=0.2, text opacity=1.0 ] (c111) at (0.9,2.5) {$1$};
	
	\draw[->] (t111.south)      -- (c111.north);
	\draw[->] (t222.south) -- (c111.north east);

	\node[rectangle, minimum width=0.4cm, minimum height=0.6cm, fill=KITcyanblue,
		opacity=0.2, text opacity=1.0] (t1111) at (0.9,1) {$11$};
	\draw[->] (c111.south) -- (t1111.north);
	
	\node[rectangle, minimum width=0.4cm, minimum height=0.6cm, fill=KITcyanblue,
				opacity=0.2, text opacity=1.0] (v1) at (0.9 + 8,10) {$01$};
	\node[rectangle, minimum width=0.4cm, minimum height=0.6cm, fill=KITcyanblue,
			opacity=0.2, text opacity=1.0] (v2) at (1.8 + 8,10) {$02$};
	\node[rectangle, minimum width=0.4cm, minimum height=0.6cm, fill=KITcyanblue,
			opacity=0.2, text opacity=1.0] (v3) at (2.7 + 8,10) {$03$};
	\node[rectangle, minimum width=0.4cm, minimum height=0.6cm, fill=KITcyanblue,
			opacity=0.2, text opacity=1.0] (v4) at (3.6 + 8,10) {$04$};
	\node[rectangle, minimum width=0.4cm, minimum height=0.6cm, fill=KITcyanblue,
			opacity=0.2, text opacity=1.0] (v5) at (4.5 + 8,10) {$05$};
	\node[rectangle, minimum width=0.4cm, minimum height=0.6cm, fill=KITcyanblue,
			opacity=0.2, text opacity=1.0] (v6) at (5.4 + 8,10) {$06$};
	\node[rectangle, minimum width=0.4cm, minimum height=0.6cm, fill=KITcyanblue,
			opacity=0.2, text opacity=1.0] (v7) at (6.3 + 8,10)  {$07$};
	\node[rectangle, minimum width=0.4cm, minimum height=0.6cm, fill=KITcyanblue,
			opacity=0.2, text opacity=1.0] (v8) at (7.2 + 8,10) {$08$};
			
	\node[circle, fill=KITred, opacity=0.2, text opacity=1.0 ] (b1) at (0.9 + 8,8.5) {$1$};
	\node[circle, fill=KITred, opacity=0.2, text opacity=1.0 ] (b2) at (1.8 + 8,8.5) {$2$};
	\node[circle, fill=KITred, opacity=0.2, text opacity=1.0 ] (b3) at (2.7 + 8,8.5) {$3$};
	\node[circle, fill=KITred, opacity=0.2, text opacity=1.0 ] (b4) at (3.6 + 8,8.5) {$4$};
	
	\node[] (text1) at (-0.5,8.5) {\small{ThreadID}};
	\node[] (text2) at (-0.5,5.5) {\small{ThreadID}};
	\node[] (text3) at (-0.5,2.5)   {\small{ThreadID}};
	
	\draw[->] (v1.south) -- (b1.north);
	\draw[->] (v5.south west) -- (b1.north east);
	\draw[->] (v2.south) -- (b2.north);
	\draw[->] (v6.south west) -- (b2.north east);
	\draw[->] (v3.south) -- (b3.north);
	\draw[->] (v7.south west) -- (b3.north east);
	\draw[->] (v4.south) -- (b4.north);
	\draw[->] (v8.south west) -- (b4.north east);
	
	\node[rectangle, minimum width=0.4cm, minimum height=0.6cm, fill=KITcyanblue,
				opacity=0.2, text opacity=1.0] (v11) at (0.9 + 8,7) {$05$};
	\node[rectangle, minimum width=0.4cm, minimum height=0.6cm, fill=KITcyanblue,
			opacity=0.2, text opacity=1.0] (v22) at (1.8 + 8,7) {$02$};
	\node[rectangle, minimum width=0.4cm, minimum height=0.6cm, fill=KITcyanblue,
			opacity=0.2, text opacity=1.0] (v33) at (2.7 + 8,7) {$07$};
	\node[rectangle, minimum width=0.4cm, minimum height=0.6cm, fill=KITcyanblue,
			opacity=0.2, text opacity=1.0] (v44) at (3.6 + 8,7) {$04$};
			
	\draw[->] (b1.south) -- (v11.north);
	\draw[->] (b2.south) -- (v22.north);
	\draw[->] (b3.south) -- (v33.north);
	\draw[->] (b4.south) -- (v44.north);
			
	\node[circle, fill=KITred, opacity=0.2, text opacity=1.0 ] (b11) at (0.9 + 8,5.5) {$1$};
	\node[circle, fill=KITred, opacity=0.2, text opacity=1.0 ] (b22) at (1.8 + 8,5.5) {$2$};

	\draw[->] (v11.south) -- (b11.north);
	\draw[->] (v33.south west) -- (b11.north east);
	\draw[->] (v22.south) -- (b22.north);
	\draw[->] (v44.south west) -- (b22.north east);
	
	\node[rectangle, minimum width=0.4cm, minimum height=0.6cm, fill=KITcyanblue,
		opacity=0.2, text opacity=1.0] (v111) at (0.9 + 8,4) {$05$};
	\node[rectangle, minimum width=0.4cm, minimum height=0.6cm, fill=KITcyanblue,
		opacity=0.2, text opacity=1.0] (v222) at (1.8 + 8,4) {$02$};
		
	\draw[->] (b11.south) -- (v111.north);
	\draw[->] (b22.south) -- (v222.north);
		
	\node[circle, fill=KITred, opacity=0.2, text opacity=1.0 ] (b111) at (0.9 + 8,2.5) {$1$};
	
	\draw[->] (v111.south)      -- (b111.north);
	\draw[->] (v222.south) -- (b111.north east);

	\node[rectangle, minimum width=0.4cm, minimum height=0.6cm, fill=KITcyanblue,
		opacity=0.2, text opacity=1.0] (v1111) at (0.9 + 8,1) {$05$};
	\draw[->] (b111.south) -- (v1111.north);
	
	\draw [decorate,decoration={brace,amplitude=8pt,raise=6pt}]
		(t1.north) -- (t8.north) node [black,midway,yshift=0.9cm]     {Aggregated costs};
	\draw [decorate,decoration={brace,amplitude=8pt,raise=6pt}]
		(v1.north) -- (v8.north) node [black,midway,yshift=0.9cm]     {Disparities};

\end{tikzpicture}%
	\caption { %
		Illustration of the map reduce method to find the \gls*{WTA} disparity with the minimum aggregated cost. %
		In each iteration, the number of aggregated costs that are being processed and the corresponding disparities are halved. %
	} %
	\label{fig:cuda_min_reduc} %
\end{figure}

The map reduce method allows to efficiently find the minimum of a given dataset. 
In this, each active thread performs a comparison between two elements and stores the smaller element in a designated memory space. %
The number of active threads, as well as the elements that are to be processed get halved by each iteration as illustrated by \Cref{fig:cuda_min_reduc}. %
Thus, the search for the minimum cost requires $\log_2(d_\mathrm{max})$ iterations. %
Ideally, the dimensions of the CUDA blocks are chosen in such a way, that there are 32 active threads at the beginning. %
This allows that the map reduce algorithm can be processed in only one warp. %

\section{Vectorized SIMD processing on ARM CPUs}
\label{sec:appendix_cpu_processing}
\glsreset{SIMD}

While conventional \gls*{SISD} processing on the CPU requires one instruction for each data transfer between the memory and the registers of the CPU, as well as for each processing of the data stored inside these registers, the additional use of vector-processors allows to perform one instruction on a set of data simultaneously, the so-called \gls*{SIMD} processing. %
Each vector-processor is divided into multiple vector-lanes, which in turn hold one datum each. %
The memory unit as well as the arithmetic and logical unit of the vector-processor allow to simultaneously transfer multiple data into and from the lanes of the vector-registers, as well as to simultaneously combine the lanes of two registers and store the results into the lanes of a third register.
The vector-processors of the ARMv8 architecture contain 32 vector-registers with a size of 128\,bits each \citep{Arm2017}. %
The number of vector-lanes inside each register differs, depending on the data type which is to be stored inside the register.
Thus, each register can have 16 lanes when data with a size of 8\,bits each is to be stored, or only two lanes, if each lane holds a datum with the size of 64\,bits. %
In particular, image processing is well-suited for the \gls*{SIMD} parallelization on vector-processors, since all pixels in an image are processed in the same manner, only with different data. %

For an efficient use of vectorized SIMD processing, the developer is compelled to consider some design guidelines \citep{Arm2013neon}:
\begin{itemize} %
	\item Reduction of the dependencies between conventional CPU and vectorized \gls*{SIMD} processing, in order to minimize the latency induced by copying data between the \gls*{SISD} and \gls*{SIMD} pipeline. %
	\item Exploitation of cache coherence, to speed up the data transfer between the memory and the vector-registers. %
	\item Dependencies in the data of vector-instructions trigger pipeline-stalls, in which the \gls*{SIMD} pipeline is stopped until the dependencies are resolved slowing down the processing. %
	\item Minimal use of conditional branching, since if the \gls*{BPU} of the CPU predicts the wrong branch, the pipeline has to be recursively cleared until the point of branching and restarted. %
\end{itemize} %

\section{Sorting networks}
\label{sec:appendix_sortingnetworks}

\begin{figure}[ht!] %
	\centering %
	\resizebox{0.6\columnwidth}{!}{\begin{tikzpicture}
	\coordinate (startA) at (1,9);
	\coordinate (endA)   at (14,9);
	\coordinate (startB) at (1,8);
	\coordinate (endB)   at (14,8);
	\coordinate (startC) at (1,7);
	\coordinate (endC)   at (14,7);
	\coordinate (startD) at (1,6);
	\coordinate (endD)   at (14,6);
	\coordinate (startE) at (1,5);
	\coordinate (endE)   at (14,5);
	\coordinate (startF) at (1,4);
	\coordinate (endF)   at (14,4);
	\coordinate (startG) at (1,3);
	\coordinate (endG)   at (14,3);
	\coordinate (startH) at (1,2);
	\coordinate (endH)   at (14,2);
	\coordinate (startI) at (1,1);
	\coordinate (endI)   at (14,1);
	
	\draw[thick] (startA) -- (endA);
	\draw[thick] (startB) -- (endB);
	\draw[thick] (startC) -- (endC);
	\draw[thick] (startD) -- (endD);
	\draw[thick] (startE) -- (endE);
	\draw[thick] (startF) -- (endF);
	\draw[thick] (startG) -- (endG);
	\draw[thick] (startH) -- (endH);
	\draw[thick] (startI) -- (endI);
	
	\node[anchor=south east] (n1) at (startA) {$8$};
	\node[anchor=south east] (n2) at (startB) {$5$};
	\node[anchor=south east] (n3) at (startC) {$6$};
	\node[anchor=south east] (n4) at (startD) {$3$};
	\node[anchor=south east] (n5) at (startE) {$0$};
	\node[anchor=south east] (n6) at (startF) {$1$};
	\node[anchor=south east] (n7) at (startG) {$4$};
	\node[anchor=south east] (n8) at (startH) {$8$};
	\node[anchor=south east] (n9) at (startI) {$3$};
	
	\fill (1,9) circle[radius=2pt];
	\fill (1,8) circle[radius=2pt];
	\draw (1,9) -- (1,8);
	\node[anchor=south] (m1) at (1.2,9) {$5$};
	\node[anchor=south] (m2) at (1.2,8) {$8$};
	
	\fill (1.4,8) circle[radius=2pt];
	\fill (1.4,7) circle[radius=2pt];
	\draw (1.4,8) -- (1.4,7);
	\node[anchor=south] (m3) at (1.6,8) {$6$};
	\node[anchor=south] (m4) at (1.6,7) {$8$};
	
	\fill (1.8,7) circle[radius=2pt];
	\fill (1.8,6) circle[radius=2pt];
	\draw (1.8,7) -- (1.8,6);
	\node[anchor=south] (m5) at (2,7) {$3$};
	\node[anchor=south] (m6) at (2,6) {$8$};
	
	\fill (2.2,6) circle[radius=2pt];
	\fill (2.2,5) circle[radius=2pt];
	\draw (2.2,6) -- (2.2,5);
	\node[anchor=south] (m7) at (2.4,6) {$0$};
	\node[anchor=south] (m8) at (2.4,5) {$8$};

	\fill (2.6,5) circle[radius=2pt];
	\fill (2.6,4) circle[radius=2pt];
	\draw (2.6,5) -- (2.6,4);
	\node[anchor=south] (m9)  at (2.8,5) {$1$};
	\node[anchor=south] (m10) at (2.8,4) {$8$};
	
	\fill (3,4) circle[radius=2pt];
	\fill (3,3) circle[radius=2pt];
	\draw (3,4) -- (3,3);
	\node[anchor=south] (m11) at (3.2,4) {$4$};
	\node[anchor=south] (m12) at (3.2,3) {$8$};
	
	\fill (3.4,3) circle[radius=2pt];
	\fill (3.4,2) circle[radius=2pt];
	\draw (3.4,3) -- (3.4,2);
	\node[anchor=south] (m13) at (3.6,3) {$8$};
	\node[anchor=south] (m14) at (3.6,2) {$8$};
	
	\fill (3.8,2) circle[radius=2pt];
	\fill (3.8,1) circle[radius=2pt];
	\draw (3.8,2) -- (3.8,1);
	\node[anchor=south] (m15) at (4,2) {$3$};
	\node[anchor=south] (m16) at (4,1) {$8$};
	
	\fill (3.8,9) circle[radius=2pt];
	\fill (3.8,8) circle[radius=2pt];
	\draw (3.8,9) -- (3.8,8);
	\node[anchor=south] (m17) at (4,9) {$5$};
	\node[anchor=south] (m18) at (4,8) {$6$};
	
	\fill (4.2,8) circle[radius=2pt];
	\fill (4.2,7) circle[radius=2pt];
	\draw (4.2,8) -- (4.2,7);
	\node[anchor=south] (m19) at (4.4,8) {$3$};
	\node[anchor=south] (m20) at (4.4,7) {$6$};
	
	\fill (4.6,7) circle[radius=2pt];
	\fill (4.6,6) circle[radius=2pt];
	\draw (4.6,7) -- (4.6,6);
	\node[anchor=south] (m21) at (4.8,7) {$0$};
	\node[anchor=south] (m22) at (4.8,6) {$6$};
	
	\fill (5.0,6) circle[radius=2pt];
	\fill (5.0,5) circle[radius=2pt];
	\draw (5.0,6) -- (5.0,5);
	\node[anchor=south] (m23) at (5.2,6) {$1$};
	\node[anchor=south] (m24) at (5.2,5) {$6$};
	
	\fill (5.4,5) circle[radius=2pt];
	\fill (5.4,4) circle[radius=2pt];
	\draw (5.4,5) -- (5.4,4);
	\node[anchor=south] (m25) at (5.6,5) {$4$};
	\node[anchor=south] (m26) at (5.6,4) {$6$};
	
	\fill (5.8,4) circle[radius=2pt];
	\fill (5.8,3) circle[radius=2pt];
	\draw (5.8,4) -- (5.8,3);
	\node[anchor=south] (m27) at (6.0,4) {$6$};
	\node[anchor=south] (m28) at (6.0,3) {$8$};
	
	\fill (6.2,3) circle[radius=2pt];
	\fill (6.2,2) circle[radius=2pt];
	\draw (6.2,3) -- (6.2,2);
	\node[anchor=south] (m29) at (6.4,3) {$3$};
	\node[anchor=south] (m30) at (6.4,2) {$8$};
	
	\fill (6.2,9) circle[radius=2pt];
	\fill (6.2,8) circle[radius=2pt];
	\draw (6.2,9) -- (6.2,8);
	\node[anchor=south] (m31) at (6.4,9) {$3$};
	\node[anchor=south] (m32) at (6.4,8) {$5$};
	
	\fill (6.6,8) circle[radius=2pt];
	\fill (6.6,7) circle[radius=2pt];
	\draw (6.6,8) -- (6.6,7);
	\node[anchor=south] (m33) at (6.8,8) {$0$};
	\node[anchor=south] (m34) at (6.8,7) {$5$};
	
	\fill (7.0,7) circle[radius=2pt];
	\fill (7.0,6) circle[radius=2pt];
	\draw (7.0,7) -- (7.0,6);
	\node[anchor=south] (m35) at (7.2,7) {$1$};
	\node[anchor=south] (m36) at (7.2,6) {$5$};
	
	\fill (7.4,6) circle[radius=2pt];
	\fill (7.4,5) circle[radius=2pt];
	\draw (7.4,6) -- (7.4,5);
	\node[anchor=south] (m37) at (7.6,6) {$4$};
	\node[anchor=south] (m38) at (7.6,5) {$5$};
	
	\fill (7.8,5) circle[radius=2pt];
	\fill (7.8,4) circle[radius=2pt];
	\draw (7.8,5) -- (7.8,4);
	\node[anchor=south] (m39) at (8.0,5) {$5$};
	\node[anchor=south] (m40) at (8.0,4) {$6$};
	
	\fill (8.2,4) circle[radius=2pt];
	\fill (8.2,3) circle[radius=2pt];
	\draw (8.2,4) -- (8.2,3);
	\node[anchor=south] (m41) at (8.4,4) {$3$};
	\node[anchor=south] (m42) at (8.4,3) {$6$};
	
	\fill (8.2,9) circle[radius=2pt];
	\fill (8.2,8) circle[radius=2pt];
	\draw (8.2,9) -- (8.2,8);
	\node[anchor=south] (m43) at (8.4,9) {$0$};
	\node[anchor=south] (m44) at (8.4,8) {$3$};
	
	\fill (8.6,8) circle[radius=2pt];
	\fill (8.6,7) circle[radius=2pt];
	\draw (8.6,8) -- (8.6,7);
	\node[anchor=south] (m45) at (8.8,8) {$1$};
	\node[anchor=south] (m46) at (8.8,7) {$3$};
	
	\fill (9.0,7) circle[radius=2pt];
	\fill (9.0,6) circle[radius=2pt];
	\draw (9.0,7) -- (9.0,6);
	\node[anchor=south] (m47) at (9.2,7) {$3$};
	\node[anchor=south] (m48) at (9.2,6) {$4$};
	
	\fill (9.4,6) circle[radius=2pt];
	\fill (9.4,5) circle[radius=2pt];
	\draw (9.4,6) -- (9.4,5);
	\node[anchor=south] (m49) at (9.6,6) {$4$};
	\node[anchor=south] (m50) at (9.6,5) {$5$};
	
	\fill (9.8,5) circle[radius=2pt];
	\fill (9.8,4) circle[radius=2pt];
	\draw (9.8,5) -- (9.8,4);
	\node[anchor=south] (m51) at (10.0,5) {$3$};
	\node[anchor=south] (m52) at (10.0,4) {$5$};
	
	\fill (10.2,9) circle[radius=2pt];
	\fill (10.2,8) circle[radius=2pt];
	\draw (10.2,9) -- (10.2,8);
	\node[anchor=south] (m53) at (10.4,9) {$0$};
	\node[anchor=south] (m54) at (10.4,8) {$1$};
	
	\fill (10.6,8) circle[radius=2pt];
	\fill (10.6,7) circle[radius=2pt];
	\draw (10.6,8) -- (10.6,7);
	\node[anchor=south] (m55) at (10.8,8) {$1$};
	\node[anchor=south] (m56) at (10.8,7) {$3$};
	
	\fill (11.0,7) circle[radius=2pt];
	\fill (11.0,6) circle[radius=2pt];
	\draw (11.0,7) -- (11.0,6);
	\node[anchor=south] (m57) at (11.2,7) {$3$};
	\node[anchor=south] (m58) at (11.2,6) {$4$};
	
	\fill (11.4,6) circle[radius=2pt];
	\fill (11.4,5) circle[radius=2pt];
	\draw (11.4,6) -- (11.4,5);
	\node[anchor=south] (m59) at (11.6,6) {$3$};
	\node[anchor=south] (m60) at (11.6,5) {$4$};
	
	\fill (11.4,9) circle[radius=2pt];
	\fill (11.4,8) circle[radius=2pt];
	\draw (11.4,9) -- (11.4,8);
	\node[anchor=south] (m61) at (11.6,9) {$0$};
	\node[anchor=south] (m62) at (11.6,8) {$1$};
	
	\fill (11.8,8) circle[radius=2pt];
	\fill (11.8,7) circle[radius=2pt];
	\draw (11.8,8) -- (11.8,7);
	\node[anchor=south] (m63) at (12.0,8) {$1$};
	\node[anchor=south] (m64) at (12.0,7) {$3$};
	
	\fill (12.2,7) circle[radius=2pt];
	\fill (12.2,6) circle[radius=2pt];
	\draw (12.2,7) -- (12.2,6);
	\node[anchor=south] (m65) at (12.4,7) {$3$};
	\node[anchor=south] (m66) at (12.4,6) {$3$};
	
	\fill (12.6,9) circle[radius=2pt];
	\fill (12.6,8) circle[radius=2pt];
	\draw (12.6,9) -- (12.6,8);
	\node[anchor=south] (m69) at (12.8,9) {$0$};
	\node[anchor=south] (m70) at (12.8,8) {$1$};
	
	\fill (13.0,8) circle[radius=2pt];
	\fill (13.0,7) circle[radius=2pt];
	\draw (13.0,8) -- (13.0,7);
	\node[anchor=south] (m71) at (13.2,8) {$1$};
	\node[anchor=south] (m72) at (13.2,7) {$3$};
	
	\fill (13.4,9) circle[radius=2pt];
	\fill (13.4,8) circle[radius=2pt];
	\draw (13.4,9) -- (13.4,8);
	\node[anchor=south] (m73) at (13.6,9) {$0$};
	\node[anchor=south] (m74) at (13.6,8) {$1$};
	
	\node[anchor=south west] (q1) at (endA) {$0$};
	\node[anchor=south west] (q2) at (endB) {$1$};
	\node[anchor=south west] (q3) at (endC) {$3$};
	\node[anchor=south west] (q4) at (endD) {$3$};
	\node[anchor=south west] (q5) at (endE) {$4$};
	\node[anchor=south west] (q6) at (endF) {$5$};
	\node[anchor=south west] (q7) at (endG) {$6$};
	\node[anchor=south west] (q8) at (endH) {$8$};
	\node[anchor=south west] (q9) at (endI) {$8$};

\end{tikzpicture}} %
	\caption{ %
		Exemplary sorting network, implementing the BubbleSort algorithm for nine input values. %
	} %
	\label{fig:sorting_network_bubble} %
\end{figure}
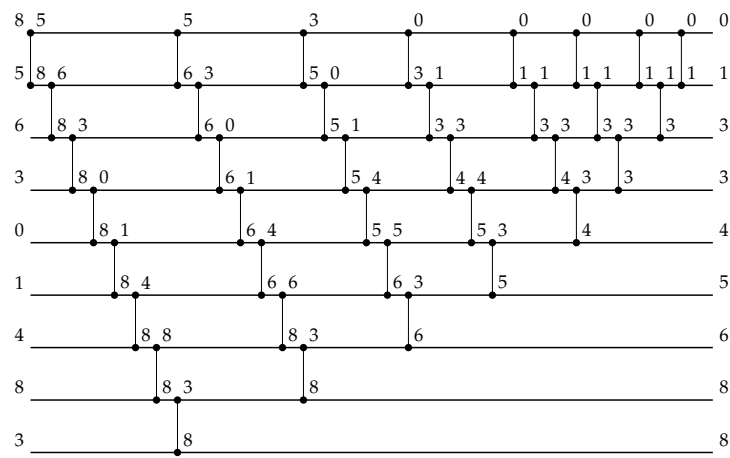

In general, sorting algorithms like BubbleSort, MergeSort or QuickSort are able to sort an arbitrary number of input values and, in turn, require indefinite number of comparative and swapping operations, making a naive implementation inappropriate for vectorized processing. %
In the case of the fixed-size median filter, however, the number of considered input values are fixed and known in advance. %
This allows to use sorting networks \citep{Knuth1998sorting}, which sort an input vector of known size with a fixed number of comparators and swapping operations. %
A sorting network is comprised of two basic building blocks, namely: %
\begin{itemize}
\item wires, which hold and transport one value of the input vector each, and
\item comparators, which are responsible for comparing the values of the connected wires and swap these if necessary.
\end{itemize} 
The comparators always connect two wires with each other and assign the smaller value to the upper wire. %
\Cref{fig:sorting_network_bubble} illustrates the BubbleSort algorithm implemented using a sorting network. %
In this, the horizontal lines represent the wires, while the comparators are illustrated by the vertical connections. %
The values of the input vector move along the wires from left to right and get rearranged by the comparators, resulting in a sorted output vector on the right with the smallest value at the top. %

\reftitle{References}


\externalbibliography{yes}
\bibliography{embedded-stereo_biblio}





\end{document}